\documentclass[twocolumn]{article}

\usepackage[utf8]{inputenc} % allow utf-8 input
\usepackage[T1]{fontenc}    % use 8-bit T1 fonts
\usepackage{hyperref}       % hyperlinks
\usepackage{url}            % simple URL typesetting
\usepackage{booktabs}       % professional-quality tables
\usepackage{amsfonts}       % blackboard math symbols
\usepackage{nicefrac}       % compact symbols for 1/2, etc.
\usepackage{microtype}      % microtypography
\usepackage{lipsum}
\usepackage{graphicx}
\graphicspath{ {./images/} }

\usepackage{graphicx}
\usepackage{amsmath}
\usepackage{amsthm}
\usepackage{booktabs}
\usepackage{color}
\usepackage{bm}
\usepackage{amssymb}
\usepackage{float}
\usepackage{subfig}
\usepackage{enumitem}
\usepackage{wrapfig} 
\usepackage{multirow}

\newtheorem{assumption}{Assumption}

\newtheorem{theorem}{Theorem}
\newtheorem{definition}{Definition}
\newtheorem{remark}{Remark}

\usepackage{setspace}

\title{Structured Directional Pruning}

\author{%
  Yinchuan Li$^{2*}$,  Xiaofeng Liu$^{1,2}$\thanks{Equal Contribution. This work was completed while Xiaofeng Liu was a member of the Huawei Noah's Ark Lab for Advanced Study.}, Yunfeng Shao$^{2}$, Qing Wang$^{1}\thanks{Corresponding author.}$, Yanhui Geng$^{2}$\\
  $^1$Tianjin University, $^2$Huawei Noah's Ark Lab\\
  \{xiaofengliull, wangq\}@tju.edu.cn\\
  \{liyinchuan, shaoyunfeng, geng.yanhui\}@huawei.com
}
\date{}
\begin{document}
\maketitle
\begin{abstract}
Structured pruning is an effective compression technique to reduce the computation of neural networks, which is usually achieved by adding perturbations to reduce network parameters at the cost of slightly  increasing training loss. A more reasonable approach is to find a sparse minimizer along the flat minimum valley found by optimizers, i.e. stochastic gradient descent, which keeps the training loss constant. To achieve this goal, we propose the structured directional pruning based on orthogonal projecting the perturbations onto the flat minimum valley.  We also propose a fast solver {\texttt{AltSDP}} and further prove that it achieves directional pruning asymptotically after sufficient training. Experiments using VGG-Net and ResNet on CIFAR-10 and CIFAR-100 datasets show that our method obtains the state-of-the-art pruned accuracy (i.e. 93.97\% on VGG16, CIFAR-10 task) without retraining. Experiments using DNN, VGG-Net and WRN28$\times$10 on MNIST, CIFAR-10 and CIFAR-100 datasets demonstrate our method performs structured directional pruning, reaching the same minimum valley as the optimizer.
\end{abstract}

\section{Introduction}

Deep Neural Network (DNN) has developed rapidly in recent years \textcolor{black}{owing to} its state-of-the-art performance in various domains~\cite{krizhevsky2012imagenet,he2016deep,deng2014deep}. \textcolor{black}{The development of DNN involves some heuristics, such as the use of deeper and more extensive models, which is also a development trend in recent years~\cite{brock2018large,simonyan2014very}. These heuristics enhance the expressive ability of neural networks by overparameterization~\cite{belkin2019reconciling}, however, in turn, restricting their usage on resource-limited devices, such as mobile phones, autonomous cars and augmented reality devices. This has prompted technological developments in shrinking DNN while maintaining accuracy.}

\textcolor{black}{Sparse DNN is a representative algorithm for shrinking DNN, which is popular since it requires less memory and storage capacity and reduces inference time~\cite{NEURIPS2020_a09e75c5}.
Here, sparse neural networks refer to neural networks with most parameters of zero. Magnitude pruning is an effective way to obtain sparse DNNs~\cite{NEURIPS2020_a09e75c5,liu2018rethinking,lemaire2019structured,frankle2018lottery,lin2019towards}.        Magnitude pruning is divided into {\em unstructured} pruning (fine-grained pruning)~\cite{NEURIPS2020_a09e75c5,liu2018rethinking,NIPS1992_303ed4c6} and {\em structured} pruning (coarse-grained pruning)~\cite{lemaire2019structured,frankle2018lottery,lin2019towards} according to whether the structure of neural networks is used. Unstructured pruning directly prunes weights independently in each layer to achieve higher sparsity. However, it usually requires dedicated hardware or software accelerators to accelerate access to irregular memory, which affects the efficiency of online reasoning~\cite{lin2019towards}.  
In contrast, structured pruning does not require dedicated hardware/software packages, as it only removes structured weights (including 2D kernels, filters, or layers) and does not yield irregular memory accesses.  }

Unfortunately, structured pruning still suffers some open issues. After removing the entire structure of the network, retraining or fine tuning is needed for better performance~\cite{lemaire2019structured}, which requires extra effort and more intensive computing~\cite{frankle2018lottery}. Moreover, these structured pruning methods are typically tailored  to specific network structures, such as filters or kernels, and cannot be flexibly applied to heterogeneous structures~\cite{lin2019towards}. In this paper, we propose a general structured  directional pruning (SDP) scheme based on perturbation orthogonal projection to solve the above problems, which does not require fine tuning or retraining. Group lasso regularization, which has shown excellent performance in areas such as  compressed sensing, online learning and tiny AI~\cite{donoho2006compressed, yang2010online, ochiai2017automatic}, is adopted to explore structural sparsity in neural networks. Subsequently, the perturbations caused by sparse regularization are orthogonally projected onto a plane with constant loss function values. Using the projected perturbation to update the network eliminates the need for fine tuning and retraining, since the accuracy of the network is not compromised. Moreover, the technique can be flexibly applied to heterogeneous structures as it can prune different structures simultaneously.

\subsection{Contributions}

In this paper, we propose a general structured pruning scheme for directional pruning of neural networks, which reaches the flat minimum valley found by optimizers, such as stochastic gradient descent~(SGD), when pruning. In particular, we orthogonally project the sparse perturbations onto a constant loss value plane and update the network accordingly. Hence, our structured directional pruning suppresses only the unimportant parameters and encourages the important ones simultaneously, while traditional structured pruning methods tend to suppress all parameters, resulting in performance losses.

In addition, a fast implementation solver, named alternating structured directional pruning ({\texttt{AltSDP}}) algorithm, based on regularized dual averaging is proposed, which can quickly adjust the weights on each structural unit to achieve orthogonal projection. Moreover, we further theoretically prove that {\texttt{AltSDP}} achieves the effect of the structured directional pruning after sufficient training.

We optimize the implementation of the proposed algorithm so that it can be combined with many optimizers and algorithms (for example stochastic gradient descent (SGD) and SGD with momentum algorithm) in the deep learning framework, e.g. Tensorflow or PyTorch. This allows our algorithm to achieve optimal pruning performance on a wide range of datasets and networks. Experiments using VGG-Net and ResNet on CIFAR-10 and CIFAR-100 datasets show that our method obtains the state-of-the-art pruned accuracy (e.g. 93.97\% on VGG16, CIFAR-10 task) without retraining. Experiments using DNN, VGG-Net and WRN28$\times$10 on MNIST, CIFAR-10 and CIFAR-100 datasets demonstrate our method performs structured directional pruning, reaching the same minimum valley as the optimizer.

\subsection{Related Works}
%\subsubsection{Structured Pruning}
\textbf{Structured pruning:} In \cite{liu2017learning}, a network slimming method based on the channel-level sparsity was proposed to automatically identify and prune insignificant channels. In~\cite{he2017channel}, a channel pruning method was proposed via a LASSO regression based channel selection and least square reconstruction. AutoML for model compression was proposed in~\cite{he2018amc}, which utilizes reinforcement learning to improve the model compression quality. Discrimination-aware channel pruning was proposed in~\cite{NEURIPS2018_55a7cf9c} to choose channels that significantly contribute to discriminative power.
In~ \cite{he2018soft}, the soft filter pruning was proposed to inference procedure of deep convolutional neural networks, which has larger model capacity and less dependence on the pre-trained model. Filter pruning via geometric median was proposed in~\cite{he2019filter}, which improves pruning performance in the cases where ``smaller-norm-less-important'' criterion does not hold.  In addition, collaborative channel pruning was proposed in~\cite{peng2019collaborative} to reduce the computational overhead of deep networks.
Polarization regularizer was proposed in~\cite{NEURIPS2020_703957b6} to suppress only unimportant neurons while keeping important neurons intact. Moreover, correlation-based pruning was proposed in~\cite{ijcai2019-525}, which utilizes parameter-quantity and computational-cost regularization terms to enable the users to customize the compression according to their preference. Unfortunately, the above methods still suffer from a loss of accuracy when pruning. Retraining and fine-tuning are hard to avoid, which requires extra effort and more intensive computing.

\textbf{Directional pruning:} Directional pruning is first proposed in~\cite{NEURIPS2020_a09e75c5}, which searches for a sparse minimizer in or close to the flat minimum valley in training loss obtained by the stochastic gradient descent.  Retraining or the expert knowledge on the sparsity level  is no longer needed.
This work motivates us to propose structured directional pruning. However, extending directional pruning to structured directional pruning is not straightforward.  Since the algorithm and theoretic analysis have major differences when the sparse $\ell_1$-norm regularization  is replaced by the group LASSO regularization.

%\cite{liu2017learning,he2017channel,he2018amc,NEURIPS2018_55a7cf9c,he2018soft,he2019filter,peng2019collaborative,NEURIPS2020_703957b6,ijcai2019-525}

\section{Structured Directional Pruning}
\subsection{Structured Pruning}
Considering a deep neural network with overparameterization ${\bm w}\in\mathbb{R}^d$, the structured pruning aims to eliminate redundant parameters in ${\bm w}$ structurally, which can be formulated as
\begin{align}
    \arg \min_{{\bm w}} \|{\bm w}^{\star}-{\bm w}\|_{2}^{2} + \mathcal{S}({\bm w},\mathcal{G}),\label{eq_structured_pruning}
\end{align}
where ${\bm w}^{\star}$ denotes a minimizer of the model parameters satisfying  $\nabla \ell({\bm w}^{\star})=0$ with $\ell: \mathbb{R}^{d}\rightarrow\mathbb{R}$ being the loss function; $\mathcal{G}$ is a structured partition of $\{1,2,\dots,d\}$ that used to divide/structure ${\bm w}$ into $|\mathcal{G}|$ groups or vectors, e.g., $\mathcal{G}=\{\{1,2,3\}, \cdots , \{d-1,d\}\}$; ${\mathcal{S}}$ is a sparse regularization, e.g., $\ell_0$ norm, to utilize the structured sparsity of parameters according to $\mathcal{G}$. Taking the group lasso regularization as an example, \eqref{eq_structured_pruning} reduces to
\begin{align}
    \arg \min_{{\bm w}} \|{\bm w}^{\star}-{\bm w}\|_{2}^{2} + \sum_{i=1}^{|\mathcal{G}|} \lambda\|{\bm w}_{i}\|_2,\label{eq_group_lasso_pruning}
\end{align}
where $\lambda > 0$ is a weight factor; ${\bm w}_{i}$ is the $i$-th group coefficients of ${\bm w}$ for $i\in[1,\cdots,|\mathcal{G}|]$~\footnote{An example: ${\bm w} = ({w}_{1},\cdots,{w}_{5}), \mathcal{G} = \{\{1,2\},\{3,4,5\}\}$, then ${\bm w}_1=(w_1,w_2), {\bm w}_2=(w_3,w_4,w_5)$.}. We can change $\mathcal{G}$ to achieve different sparse structure, e.g., filter-level sparsity, kernel-level sparsity and vector-level sparsity. And if $|\mathcal{G}|=d$ with $\mathcal{G}=\{\{1\}, \{2\}, \cdots , \{d\}\}$, the structured pruning reduces to non-structured pruning or fine-grained
pruning, which prunes weights irregularly.

Note that, the sparse regularization in \eqref{eq_group_lasso_pruning} penalizes all ${\bm w}_{i}$ to realize structured pruning, which may increase training loss while pruning. Figure~\ref{fig_structured_directional_pruning} (left) demonstrates this limitation intuitively, in which the dark blue region contains all possible case for traditional structured pruning. To solve this problem, we propose the structured directional pruning in the next subsection.

\subsection{SDP: Problem Formulation}

\begin{figure}[t]
	\centering
	\subfloat{
	\includegraphics[height=1.1in]{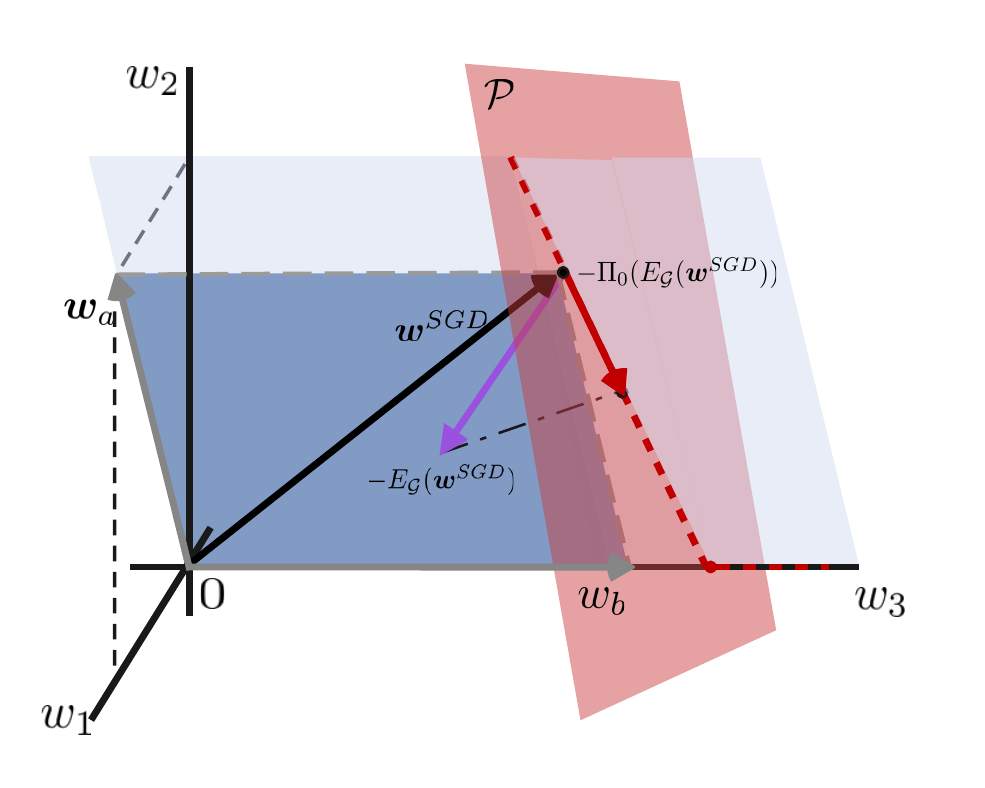}
	}
	\subfloat{
	\includegraphics[height=1.0in]{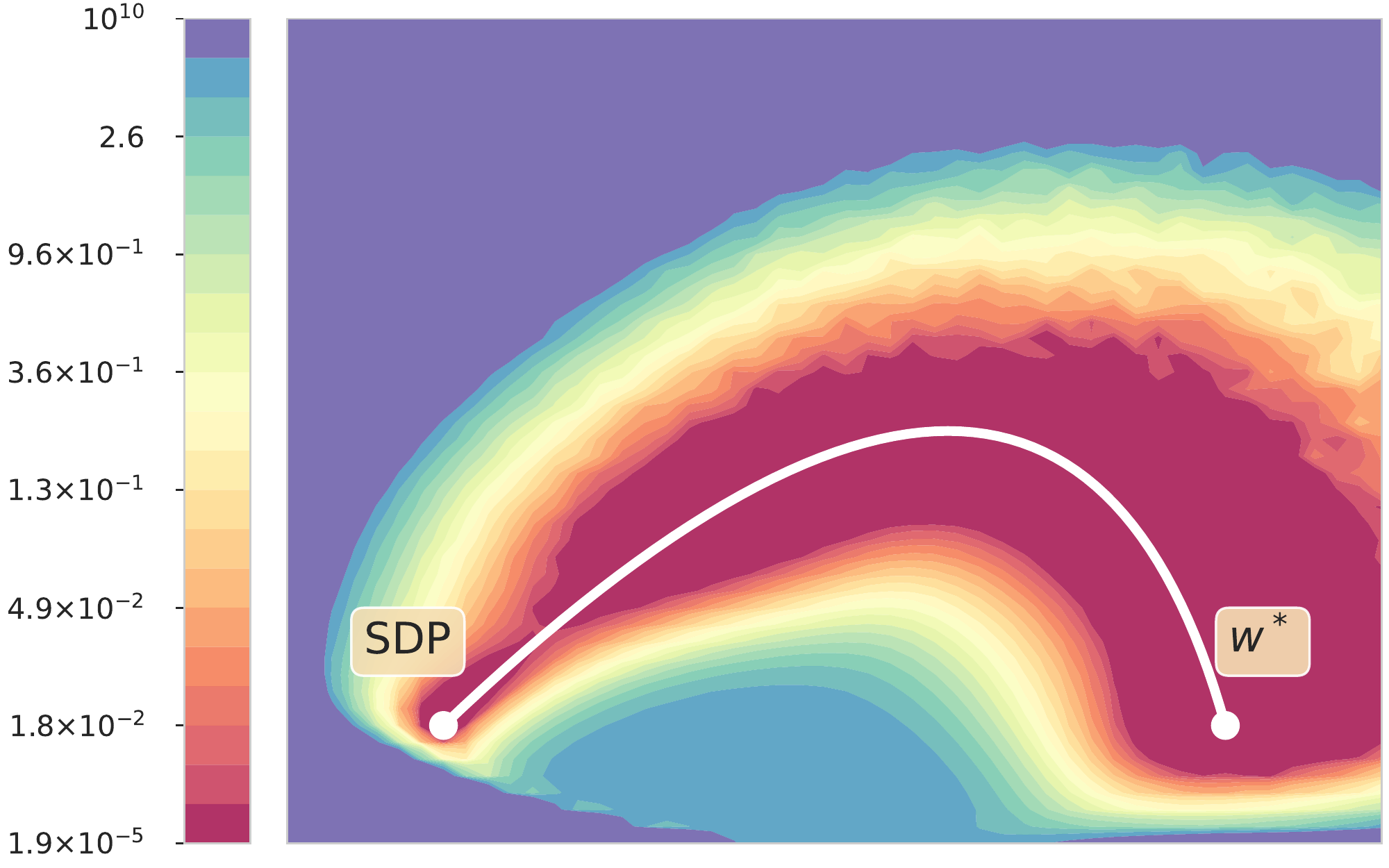}
	}
	\caption{\textbf{Left:} a 3D graphical illustration of the structured pruning, where ${\bm w}^{\star}=(w_1, w_2, w_3)$ with $\mathcal{G}=\{\{1,2\},\{3\}\}$. The dark blue region contains all possible directions of traditional structured pruning, while the red dashed line contains directions of structured directional pruning  with different $\lambda$, which is the orthogonal projection of traditional directions onto the constant loss value plane ${\mathcal{P}}$ (red). \textbf{Right:} the contour of the training loss on CIFAR10, VGG16 task around the solutions found by {\texttt{AltSDP}} (the algorithm we propose to achieve SDP) and SGD ($\bm w^{*}$). The white curve can be understood as the red dashed line in the left figure, while the dark red region can be understood as the plane ${\mathcal{P}}$.
	}
	\label{fig_structured_directional_pruning} 
\end{figure}

	\textcolor{black}{Structured directional pruning tends to prune the neural network along the direction that does not change the training loss. The idea behind is  first to  find a subspace, called $\mathcal {P}$ (red subspace in Figure.~\ref{fig_structured_directional_pruning}), where the training loss is fixed, and then project the sparse perturbation onto it.  The network is updated with the perturbation after projection to keep the training loss constant.}
	
To find ${\mathcal{P}}$, we first analysis the local geometry of the loss function through its Hessian matrix. Since $\nabla f({\bm w}^{\star})\approx 0$, the Hessian $\nabla^{2}\ell({\bm w}^{\star})$ has multiple nearly zero eigenvalues~\cite{ghorbani2019investigation, papyan2019measurements}. According to the second Taylor expansion of $\ell({\bm w}^{\star})$, the training loss will be almost constant when pruning in directions related to these eigenvalues. This means that the subspace $\mathcal{P}$ can be generated based on these directions. Note that, traditional structured pruning (the purple vector in Figure~\ref{fig_structured_directional_pruning}) is difficult to prune networks along $\mathcal{P}$, since it is nearly orthogonal to ${\bm w}^{\star}$~\cite{ghorbani2019investigation}, which may reveal why traditional structured pruning requires fine tuning or retraining.

To prune ${\bm w}^{\star}$ along $\mathcal{P}$, inspired by the directional pruning~\cite{NEURIPS2020_a09e75c5}, we first introduce direction factors $s_{i}, i = 1,\cdots,|\mathcal{G}|$, which reflects the angle between ${\bm w}_{i}$ and ${\Pi}_{i}({\bm w}^{\star})$, where  ${\Pi}(\cdot)$ represents an operator of projecting the input vector onto the subspace $\mathcal{P}$, and ${\Pi}_i(\cdot)$ denotes its $i$-th group that is separated w.r.t. $\mathcal{G}$. Different from \eqref{eq_group_lasso_pruning}, structured directional pruning, defined in Definition~\ref{definition_1}, decrease the magnitude of ${\bm w}_{i}^{\star}$ with $s_{i}>0$ (acute angle) and simultaneously increase the magnitude of ${\bm w}_{i}^{\star}$ with $s_{i}<0$ (obtuse angle).

\begin{definition}
\label{definition_1}
(Structured directional pruning). Suppose that ${\bm w}^{\star}\in\mathbb{R}^{d}$ is a minimizer satisfies $\nabla \ell({\bm w}^{\star})=0$ with $\ell(\cdot)$ being the loss function. Assume that none of the coefficients in ${\bm w}^{\star}$ is zero. The structured directional pruning is given by
\begin{align}
    \arg\min_{{\bm w}} \frac{1}{2}\| {\bm w}^{\star} - {\bm w} \|_{2}^{2} + \lambda \sum_{i=1}^{|\mathcal{G}|} s_{i}\| {\bm w}_{i} \|_2,    \label{eq_def_sdp}
\end{align}
where $\lambda > 0$ is a weight factor, $\mathcal{G}$ is the structured partition, and $s_{i}$ is the direction factor
\begin{align}
    s_{i} := \left \langle E({\bm w}_{i}^{\star})   ,  {\Pi}_{i}(E_{\mathcal{G}}({\bm w}^{\star}))  \right \rangle> 0   \label{eq_define_sa}
\end{align}
with $\langle \cdot \rangle$ being the inner product, $E(\cdot)$ being the normalization operator, i.e., $E({\bm w})={\bm w}/\|{\bm w}\|_{2}$, and $E_{\mathcal{G}}(\cdot)$ being the normalization operator w.r.t. $\mathcal{G}$, i.e., $E_{\mathcal{G}}({\bm w}) = [ E({\bm w}_{1})^T, E({\bm w}_{2})^T,...,E({\bm w}_{|\mathcal{G}|})^T ]^T$.
\end{definition}

Our structured directional pruning can also be understood as being based on the orthogonal projection of perturbations. That is, structured pruning $\bm w^{\star}$ in \eqref{eq_structured_pruning} can be rewritten as a perturbation of $\bm w^{\star}$, i.e., 
\begin{align}
\bm w_i^{\star} - \xi_i E( \bm w_i^{\star}), ~i = 1,...,|\mathcal{G}|,
\end{align}
where $0 \leq \xi_i \leq \| \bm w_i^{\star} \|_2$ and the $i$-th group of $\bm w^{\star}$ is pruned if $\xi_i = \| \bm w_i^{\star} \|_2$. Since usually $[ \xi_1 E({\bm w}_{1}^{\star})^T,  \xi_2  E({\bm w}_{2}^{\star})^T,...,   \xi_{|\mathcal{G}|} E({\bm w}_{|\mathcal{G}|}^{\star})^T ]^T  \notin \mathcal{P}$, retraining is need for traditional structured pruning. By comparision, our SDP can be viewed as pruning  $\bm w^{\star}$ by setting $\xi_i = \lambda s_{i}$, then we have $\bm w_i^{\star} - \lambda   {\Pi}_{i}(E_{\mathcal{G}}({\bm w}^{\star})), ~i = 1,...,|\mathcal{G}|$ by noting that $s_{i} E( \bm w_i^{\star}) =  {\Pi}_{i}(E_{\mathcal{G}}({\bm w}^{\star}))  $, i.e., pruning along $\mathcal{P}$ by using the orthogonal projection of perturbations.

\subsection{Difference from Directional Pruning}

For comparison, we briefly present directional pruning~\cite{NEURIPS2020_a09e75c5}, which prunes weights one by one as follows
\begin{align}
    \arg\min_{{\bm w}} \frac{1}{2}\| {\bm w}^{\star} - {\bm w} \|_{2}^{2} + \lambda \sum_{j=1}^{d} s_{j}^{\rm d} | {w}_{j} |,    \label{eq_def_dp}
\end{align}
where ${w}_{j}$ denotes the $j$-th element in $\bm w$ and 
\begin{align}
    s_{j}^{\rm d} := {\rm sign}({w}_{j}^{\star})   \cdot  ({\Pi}^{\rm d}\{ {\rm sign}({\bm w}^{\star}) \}  )_j> 0,   \label{eq_define_sj}
\end{align}
where  ${\Pi}^{\rm d}(\cdot)$ represents the operator of projecting the input vector onto the subspace $\mathcal{P}^{\rm d}$, where the training loss in \eqref{eq_def_dp} is fixed.
%Obviously, our SDP and directional pruning have the following differences: 1) SDP uses regularization based on group LASSO, while directional pruning uses regularization based on $\ell_1$-norm; 2) In \eqref{eq_define_sj}, we only need to calculate one projection ${\Pi}_{0}(\cdot)$. In contrast, SDP introduces different grouping structures, and each group has its own different projection operator ${\Pi}_{i}(\cdot)$. Calculating them simultaneously makes the problem more complicated; 3) The vector $ {\rm sign}({\bm w}^{\star})$ is on the vertices of the unit hypercube, while $E_{\mathcal{G}}({\bm w}^{\star})$ is on the unit hypersphere. It is more difficult to project a vector on the unit hypersphere onto ${\mathcal{P}}$, since the number of vectors on it is far more than that on the vertices of the unit hypercube; 4) We obtain $s_j$ via performing magnitude correction on the sign of $w_j$ according to \eqref{eq_define_sj}, while we obtain $s_i$ by calculating the inner product between two normalized vectors according to \eqref{eq_define_sa}.
Obviously, our SDP and directional pruning have the following differences: 
\begin{itemize}
\item SDP uses regularization based on group LASSO, while directional pruning uses that based on $\ell_1$-norm; 

\item In \eqref{eq_define_sj}, we only need to calculate one projection ${\Pi}^{\rm d}(\cdot)$. In contrast, SDP introduces different grouping structures, and each group has its own different projection operator ${\Pi}_{i}(\cdot)$. Calculating them simultaneously makes the problem more complicated; 

\item The vector $ {\rm sign}({\bm w}^{\star})$ is on the vertices of the unit hypercube, while $E_{\mathcal{G}}({\bm w}^{\star})$ is on the unit hypersphere. It is more difficult to project a vector on the unit hypersphere onto ${\mathcal{P}}$, since the number of vectors on it is far more than that on the vertices of the unit hypercube; 

%\item We obtain $s_j^{\rm d}$ in \eqref{eq_define_sj} by performing magnitude correction on the sign of $w_j$, while we obtain $s_i$ in \eqref{eq_define_sa} by calculating the inner product between two normalized vectors.

\item We obtain $s_j^{\rm d}$ in \eqref{eq_define_sj} by performing magnitude correction on the sign of $w_j$, while we obtain $s_i$ in \eqref{eq_define_sa} by calculating the inner product between two normalized vectors and limiting the angle between them to be an acute angle, i.e., the inner product is greater than 0.
\end{itemize}

The above differences make the solutions of SDP and directional pruning different, and also make the corresponding solvers different. We will propose the SDP solution and the corresponding fast solver in the next section.

\section{Algorithm \& Theoretical Analysis}
In this section, we first present the optimal solution of the structured directional pruning in \eqref{eq_def_sdp}. Since it is computationally unfriendly to neural networks, we then propose its fast solver, and further prove that the proposed solver can asymptotically achieve the effect of structured directional pruning  under some reasonable assumptions.

%\subsection{Preparations}
\subsection{Algorithm}\label{sub_set_algorithm}
To start with, since the objective function in \eqref{eq_def_sdp} is separable for each group/structure, we propose the following Theorem~\ref{theorem_solution} to demonstrate that each subproblem has an explicit solution.
\begin{theorem}\label{theorem_solution}
Consider the optimization problem
\begin{align}
    \label{eq_prop_01}
    \arg\min_{ {\bm w}_{i} } \left\{ \frac{1}{2} \|{\bm w}_{i}^{\star}-{\bm w}_{i}\|_{2}^{2} + \lambda {s}_{i}   \|{\bm w}_{i}\|_{2} \right\}.
\end{align}
For ${\bm w}_{i}^{\star} \in \mathbb{R}^{d}\backslash\{\bm 0\}, {s}_{i} \in \mathbb{R}, \lambda > 0$, \eqref{eq_prop_01} has an explicit solution:
\begin{align}
    \label{eq_prop_02}
    \hat{\bm w}_{i} = \left( 1-\frac{\lambda {s}_{i}}{\|{\bm w}_{i}^{\star}\|_{2}}\right)_{+}{\bm w}_{i}^{\star}.
\end{align}
\end{theorem}
The above theorem gives the solution to the subproblem of \eqref{eq_def_sdp}, which is proved in Appendix. Since the original problem is a superposition of subproblems, i.e., the above problem indirectly elucidates the solution of directional pruning. However, determining $s_i$ in Theorem~\ref{theorem_solution} requires computing the Hessian $\nabla^{2}\ell({\bm w}^{\star})$ to find $\mathcal{P}_{0}$, which is computationally cumbersome. We hence propose a fast solver by alternating minimization, named {\texttt{AltSDP}}, to asymptotically obtain the structural sparse model in the following. The idea behind is to separate the progress of finding the optimal ${\bm w}^{\star}$ and its sparse structure through an alternative manner.

Considering an overparameterized DNN with training data $Z_{i}=\left\{(X_i,Y_i)\right\}_{i=1}^N$ and parameters ${\bm w}\in\mathbb{R}^d$. Assume that $h(x;\bm w)$ is the network output, denote $f(\bm w; Z) := \mathcal{L}(h(X;\bm w);Y)$ with $\mathcal{L}(h;y)$ being a loss function, e.g. the cross-entropy loss or the mean squared error (MSE) loss. And let $\nabla f({\bm w},Z)$ be the gradient of $f({\bm w},Z)$ w.r.t. ${\bm w}$. To this end, our {\texttt{AltSDP}} is given by
\begin{align}
    {\bm v}_{n+1} &= {\bm v}_{n}-\gamma \nabla f({\bm w}_{n};{Z}_{n+1})   \tag{{\texttt{AltSDP}}-(a)} \label{eq_SDP_SMD_1}\\
    {\bm w}_{n+1} &= \arg\min_{{\bm w}\in\mathbb{R}^{d}} \Big \{\frac{1}{2}\|{\bm w}\|_{2}^{2} - {\bm w}^{T}{\bm v}_{n+1} \notag \\
    &~~~~~~~~~~~~~~~~~~~+ g(n,\gamma)\sum_{i=1}^{|\mathcal{G}|}\|{\bm w}_{i}\|_{2} \Big\} ,     \tag{{\texttt{AltSDP}}-(b)}\label{eq_SDP_SMD_2}
\end{align}
where $n = 0, 1, \cdots$ is the iteration number;  $Z_{n+1}\in Z$ is the $n$-th training data; $g(n,\gamma) = c\sqrt{\gamma}(n\gamma)^{\mu}$ is the tuning function movitated by \cite{chao2019generalization} with $c, \mu >0$ being two hyperparameters that control the strength of penalization. The iteration of {\texttt{AltSDP}} can be easily started with a random initialization ${\bm w}_{0}$. Note that, following the proof of Theorem~\ref{theorem_solution}, we can have the solution of \eqref{eq_SDP_SMD_2}, i.e., for each $i \in [ 1,..., |\mathcal{G}|]$, we have
\begin{align}
    \label{sDprun-2-solution}
    {\bm w}_{n+1,i} = \left( 1-\frac{    g(n,\gamma)   }{\|   {\bm v}_{n+1, i}   \|_{2}}\right)_{+}   {\bm v}_{n+1, i}  .
\end{align}

\begin{remark}
Following the analysis in \cite{chao2019generalization, orabona2015generalized}, it shows that $g(n,\gamma)$ is the most important part to achieve the structured directional pruning, where 
$(n\gamma)^{\mu}$ is used to match the growing magnitude of \eqref{eq_SDP_SMD_1}.
If $g(n,\gamma)=n\gamma$ and ${\bm w}_{0}=0$, our {\texttt{AltSDP}} reduces to the group lasso based regularized dual averaging~(RDA) algorithm~\cite{xiao2009dual}, and no longer has structured directional pruning ability.
\end{remark}

\subsection{Theoretical Analysis}

 In this subsection, we show {\texttt{AltSDP}} achieves the structured directional pruning asymptotically based on the stochastic gradient descent, i.e., under the condition that $\bm w^{\star} = \bm w^{SGD}$.
 Denote $G({\bm w}): = \nabla\ell({\bm w}):=\mathbb{E}_{\mathcal{Z}}[\nabla f(\bm w;Z)]$, where $\mathbb{E}_{\mathcal{Z}}[\nabla f(\bm w,Z)] = 1/N \sum_{i=1}^N f(\bm w;Z_i)$. Define $\Sigma({\bm w}):=\mathbb{E}_{\mathcal{Z}}\left[ (\nabla f({\bm w};Z)-   \nabla\ell({\bm w})  ) (\nabla f({\bm w};Z)- \nabla\ell({\bm w})  )^{T} \right].$   
 Define the gradient flow ${\bm w}(t)$ to be the solution of the ordinary differential equation (ODE)
 \begin{align}
     \label{eq_gradient_flow}
     \dot{{\bm w}} = -G({\bm w}), {\bm w}(0)={\bm w}_{0},
 \end{align}
 where ${\bm w}_{0}$ is a random initializer. It is known that $\bm w(t)$ can find a good global minimizer under various conditions~\cite{arora2018optimization}. Hence,  we assume the solution of \eqref{eq_gradient_flow} is unique.

 Let $H(\cdot):=\mathbb{E}_{\mathcal{Z}}[\nabla^{2}f(\cdot;Z)]$ be the Hessian matrix, and $\Phi(t,s)$ be the principal matrix solution of the matrix ODE system~\cite{teschl2012ordinary}:
 \begin{align}
     \label{eq_ode_system}
     \frac{d\Phi(t,s)}{dt}=-H({\bm w}(t))\Phi(t,s),\quad \Phi(s,s) = I_{d}.
 \end{align}

Define ${\bm w}_{\gamma}(t):={\bm w}_{\left\lfloor t/\gamma \right\rfloor}$, where $\left\lfloor x \right \rfloor$ denotes the greatest integer not greater than $x$.
	 Then, we make the following reasonable assumptions.

\begin{assumption}
\label{assumption_1}
$\nabla\ell({\bm w}) : \mathbb{R}^d \rightarrow \mathbb{R}^d$  is continuous on $\mathbb{R}^d$.  
\end{assumption}

\begin{assumption}
\label{assumption_2}
$\Sigma({\bm w}) : \mathbb{R}^d \rightarrow \mathbb{R}^{d \times d} $ is continuous, and $\mathbb{E}_{\mathcal{Z}} [\sup_{\|{\bm w}\|\leq K}\|\nabla f({\bm w}, Z)]<\infty$ a.s. for any $K$.
\end{assumption}

\begin{assumption}
\label{assumption_3}
The Hessian matrix $H({\bm w}) : \mathbb{R}^d \rightarrow \mathbb{R}^{d \times d}  $ is continuous.
\end{assumption}

\begin{assumption}
\label{assumption_3_b}
There exists a non-negative definite matrix $\bar{H}$ such that $\int_{0}^{\infty}\| H({\bm w}(s))-\bar{H} \|ds\leq\infty$ with $\|\cdot\|$ being the spectral norm, and the eigenspace of $\bar{H}$ associated with zero eigenvalues matches the subspace $\mathcal{P}$.
\end{assumption}

\begin{assumption}
\label{assumption_4}
${c\int_{0}^{t}\Phi(t,s)\frac{\partial E_{\mathcal{G}}({\bm w}(s))s^{\mu}}{\partial s}ds = {o}(t^{\mu})}$, where $E_{\mathcal{G}}({\bm w}(t))$ is defined in \eqref{eq_define_sa}.
\end{assumption}

\begin{assumption}
\label{assumption_5}
There exists $\bar{T}>0$, such that $E_{\mathcal{G}}({\bm w}(t)) = E_{\mathcal{G}}({\bm w}^{SGD}(t)) = E_{\mathcal{G}}({\bm w}(\bar{T}))$ a.s. for $t>\bar{T}$.
\end{assumption}

\begin{figure}[ht]
	\centering
	\includegraphics[height=1.5in]{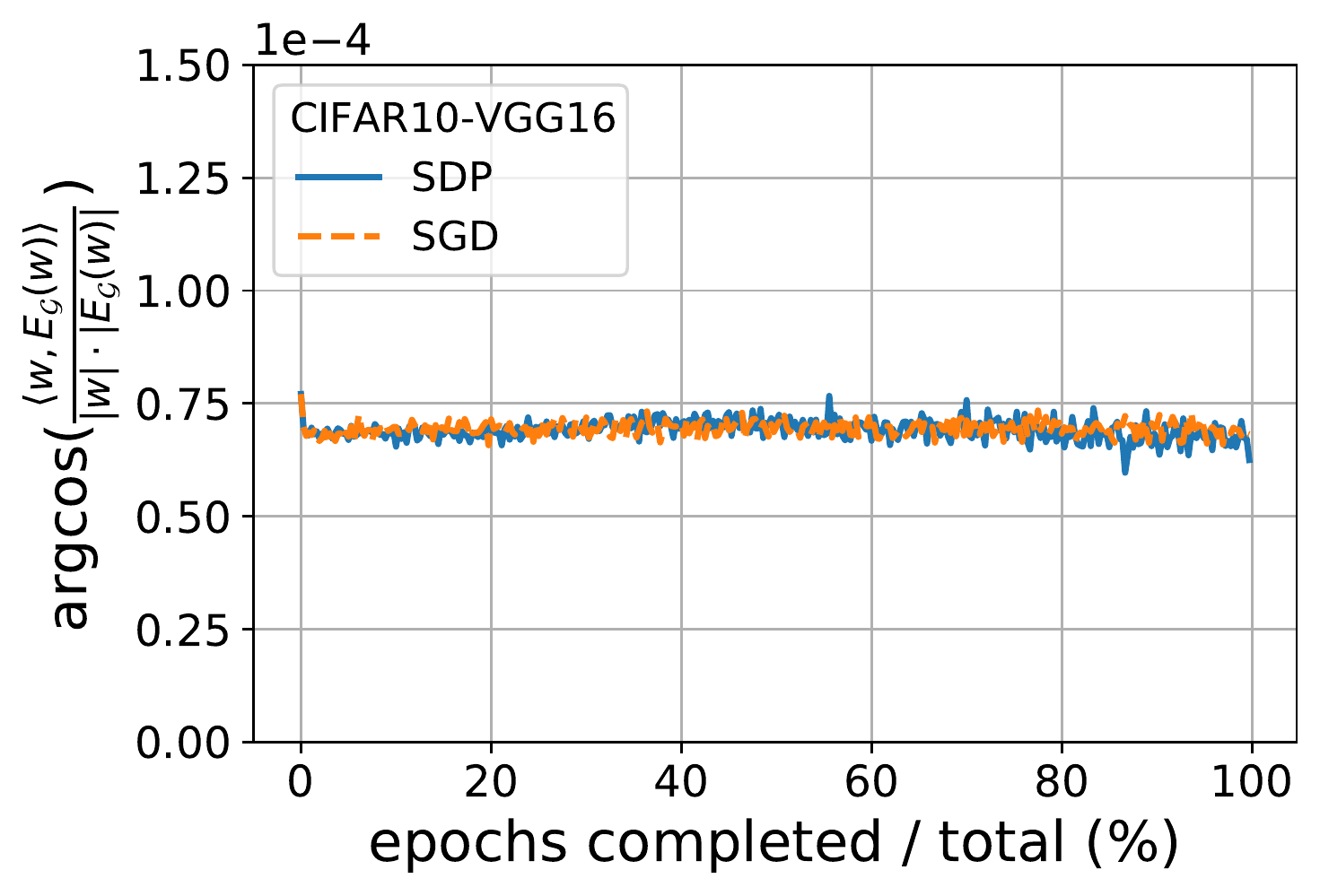}
	\caption{The angle between $\bm w(t)$ and ${ E_{\mathcal{G}}({\bm w}(t))}$.}
	\label{angleFig}
\end{figure}

Note that the above assumptions hold empirically and strictly under some conditions. We explain in detail in the remarks below. Next, we propose Theorem~\ref{theorem_1} to show that {\texttt{AltSDP}} achieves the structured directional pruning asymptotically, which is proved in Appendix.

\begin{remark}
\label{remark-A4}
(Condition (A4)). This condition can be verified on some simple networks under the MSE loss. It is known that once $\bm w(t)$ and SGD converge to the same flat valley of minima, the subspace $\mathcal{P}$ matches the eigenspace of $\bar{H}$ associated with the zero eigenvalues~\cite{NEURIPS2020_a09e75c5}. In addition, \cite{du2018convolutional,zhang2019learning} showed that $\bm w(t) \rightarrow \bm w^{\star}$ for one hidden layer networks under the teacher-student framework and MSE loss, then the limit $\bar{H} = H(\bm w^{\star})$ and the condition holds. 
\end{remark}

\begin{remark}
(Condition (A5)). This assumption can be understood as that ${ E_{\mathcal{G}}({\bm w}(t))}$ is assumed to be mainly restricted in the eigenspace of $H(\bm w(t))$ associated with positive eigenvalues as $t \rightarrow \infty$. Since~\cite{gur2018gradient,ghorbani2019investigation} proved that $\bm w(t)$ lies mainly in the subspace of $H(\bm w(t))$ associated with positive eigenvalues, and Figure~\ref{angleFig} shows that the angle between $\bm w(t)$ and ${ E_{\mathcal{G}}({\bm w}(t))}$ is very small, we can conclude that this assumption holds empirically.
\end{remark}

\begin{remark}
(Condition (A6)). For the MSE loss, we have $\bm w(t) \rightarrow \bm w^{\star}$ under some conditions according to Remark~\ref{remark-A4}, hence $E_{\mathcal{G}}({\bm w}(t)) = E_{\mathcal{G}}({\bm w}(\bar{T}))$ holds. 
For the cross-entropy loss,  \cite{xiao2009dual, gunasekar2018characterizing} proved that $\bm w(t)/ \| \bm w(t) \|_2$ converges to a unique direction when $\| \bm w(t) \|_2 \rightarrow \infty$, which shows that $E_{\mathcal{G}}({\bm w}(t))$ stabilizes after a finite time and hence we have $E_{\mathcal{G}}({\bm w}(t)) = E_{\mathcal{G}}({\bm w}(\bar{T}))$ holds. In addition, if $\gamma\rightarrow 0$, the deviation between the gradient flow and the SGD is small, and hence $E_{\mathcal{G}}({\bm w}(t)) = E_{\mathcal{G}}({\bm w}^{SGD}(t))$ holds.
\end{remark}

\begin{table*}[ht] \small
  \centering
  \scalebox{1}{
  \begin{tabular}{ccccccc}
    \toprule
    \multirow{3}{*}{Dataset} &\multirow{3}{*}{Model} &\multirow{3}{*}{Method} & Baseline  & \multicolumn{3}{c}{Pruned} \\
    \cmidrule(r){4-4}
    \cmidrule(r){5-7}
        &   &   & \multirow{2}{*}{Acc. (\%)} & \multirow{2}{*}{Acc. (\%)} & Acc. & FLOPs  \\
        &   &   &   &   & Drop (\%) & Reduction \\
    \midrule
    \multirow{15}{*}{CIFAR-10} & \multirow{11}{*}{ResNet-56} & NS~\cite{liu2017learning} (New)   & 93.80 & 93.27 & 0.53 & 48\%\\
       &   & CP~\cite{he2017channel}     & 92.80 & 91.80 & 1.00 & 50\%\\
       &   & AMC~\cite{he2018amc}    & 92.80 & 91.90 & 0.90 & 50\%\\
       &   & DCP~\cite{NEURIPS2018_55a7cf9c}    & 93.80 & 93.49 & 0.31 & 50\%\\
       &   & DCP-adapt~\cite{NEURIPS2018_55a7cf9c}     & 93.80 & 93.81 & -0.01 & 47\%\\
       &   & SFP~\cite{he2018soft}    & 93.59 & 93.35 & 0.24 & 51\%\\
       &   & FPGM~\cite{he2019filter}   & 93.59 & 93.49 & 0.10 & 53\%\\
       &   & CCP~\cite{peng2019collaborative}    & 93.50 & 93.46 & 0.04 & 47\%\\
       &   & DeepHoyer~\cite{peng2019collaborative}     & 93.80 & 93.54 & 0.26 & 48\%\\
       &   & PR~\cite{NEURIPS2020_703957b6}     & 93.80 & 93.83 & -0.03 & 47\%\\
       &   & Ours     & 93.80 & \textbf{93.90} & \textbf{-0.10} & \textbf{55\%}\\
    \cmidrule(r){2-7}
        & \multirow{4}{*}{VGG-16} & NS~\cite{liu2017learning} (New)   & 93.88 & 93.62 & 0.26 & 51\%\\
        &   & FPGM~\cite{he2019filter}   & 93.58 & 93.54 & 0.04 & 34\%\\
        &   & PR~\cite{NEURIPS2020_703957b6}   & 93.88 & 93.92 & -0.04 & 54\%\\
        &   & Ours   & 93.88 & \textbf{93.97} & \textbf{-0.09} & \textbf{55\%}\\
    \midrule
    \multirow{7}{*}{CIFAR-100} & \multirow{3}{*}{ResNet-56} & NS~\cite{liu2017learning} (New)   & 72.49 & 71.40 & 1.09 & 24\%\\
        &   & PR~\cite{NEURIPS2020_703957b6}     & 72.49 & 72.46 & 0.06 & \textbf{25\%}\\
        &   & Ours     & 72.49 & \textbf{72.55} & \textbf{-0.06} & 24\%\\
        \cmidrule(r){2-7}
        & \multirow{4}{*}{VGG-16} & NS~\cite{liu2017learning} (New)   & 73.83 & 74.20 & -0.37 & 38\%\\
        &   & COP~\cite{ijcai2019-525}   & 72.59 & 71.77 & 0.82 & 43\%\\
        &   & PR~\cite{NEURIPS2020_703957b6}   & 73.83 & 74.25 & -0.42 & 43\%\\
        &   & Ours   & 73.83 & \textbf{74.29} & \textbf{-0.46} & \textbf{43\%}\\
    \bottomrule
  \end{tabular}}
  \caption{Results on CIFAR-10 and CIFAR-100. Best results are bolded.}
  \label{dif-alg-table}
\end{table*}

\begin{table*}[ht] \small
  \centering
  \scalebox{1}{
  \begin{tabular}{ccccccc}
    \toprule
    \multirow{3}{*}{Dataset} &\multirow{3}{*}{Model} &\multirow{3}{*}{Method} & Baseline  & \multicolumn{3}{c}{Pruned} \\
    \cmidrule(r){4-4}
    \cmidrule(r){5-7}
        &   &   & \multirow{2}{*}{Acc. (\%)} & \multirow{2}{*}{Acc. (\%)} & Acc. & FLOPs  \\
        &   &   &   &   & Drop (\%) & Reduction \\
    \midrule
    \multirow{5}{*}{CIFAR-10} & \multirow{5}{*}{ResNet-56} & NS~\cite{liu2017learning} (New)   & 93.80 & 91.20 & 2.60 & 68\%\\
       &   & DeepHoyer~\cite{peng2019collaborative}     & 93.80 & 91.26 & 2.54 & 71\%\\
       &   & UCS~\cite{NEURIPS2020_703957b6}     & 93.80 & 92.25 & 1.55 & 70\%\\
       &   & PR~\cite{NEURIPS2020_703957b6}     & 93.80 & 92.63 & 1.17 & 71\%\\
       &   & Ours     & 93.80 & \textbf{92.94} & \textbf{0.86} & \textbf{72\%}\\
    \bottomrule
  \end{tabular}}
  \caption{Results on large FLOPs reduction. Best results are bolded.}
  \label{larg-flops-table}
\end{table*}

\begin{theorem}\label{theorem_1}
Under Assumptions~\ref{assumption_1}-\ref{assumption_5}, suppose $\mu\in(0.5,1)$ and $c>0$,  when $\gamma\rightarrow 0$,  {\texttt{AltSDP}} achieves structured directional pruning based on ${\bm w}^{SGD}(t)$ asymptotically, i.e., we have for $t>\bar{T}$
\begin{align}
    {\bm w}_{\gamma}(t) &\overset{d}{\approx} \arg\min_{{\bm w} \in \mathbb{R}^d} \frac{1}{2}\| {\bm w}^{SGD}(t) - {\bm w} \|_{2}^{2}  \notag \\
    &~~~~~~~~~~~~~~~~~~~~~~~~~~~~~~~~~ + \lambda_{\gamma,t} \sum_{i=1}^{|\mathcal{G}|} \bar s_{i}\| {\bm w}_{i} \|_2 \label{eq_sdp_direction_pruning}
\end{align}
and 
\begin{align}
    {\bm w}_{\gamma, i}(t) \overset{d}{\approx}  \left( 1-\frac{\lambda_{\gamma,t} \bar {s}_{i}}{\|{\bm w}_{i}^{SGD}\|_{2}}\right)_{+}{\bm w}_{i}^{SGD},~i = 1,...,|\mathcal{G}|, \notag
\end{align}
where $\lambda_{\gamma,t}=c\sqrt{\gamma}t^{\mu}$;  $\overset{d}{\approx}$ represents ``asymptotic in distribution'' under the empirical probability measure of gradients; and $\bar s_i$ satisfies $\lim_{t\rightarrow \infty} | \bar s_i  - s_i| = 0$ for all $i$.
\end{theorem}

Theorem~\ref{theorem_1} shows that {\texttt{AltSDP}} achieves directional pruning asymptotically after enough training ($t>T$) with learning rate $\gamma\rightarrow 0$. This conclusion is crucial for fitting directional structured pruning into neural network training, which avoids computing the Hession matrix and makes {\texttt{AltSDP}} work as fast as the basic SGD. The left side in \eqref{eq_sdp_direction_pruning} denotes the finally solution found by {\texttt{AltSDP}}, while the right side in \eqref{eq_sdp_direction_pruning} denotes the optimal solution of structured directional pruning according to Definition~\ref{definition_1} and Theorem~\ref{theorem_solution}.

\section{Experiments}
In this section, we carry out extensive experiments to evaluate our {\texttt{AltSDP}} algorithm, and present the evidence that {\texttt{AltSDP}} achieves the structured directional pruning asymptotically. We compare different structured pruning algorithms in Section~\ref{sub_set_performance}. In Section~\ref{sub_set_connectivity}, we analyze the effect of hyperparameters in {\texttt{AltSDP}} and show that it performs the structured directional pruning by checking whether the {\texttt{AltSDP}} algorithm reaches the same valley as the SGD algorithm.

\subsection{Experimental Setup}\label{sub_set_setup}
We use {\texttt{AltSDP}} algorithm to simultaneously train and prune two widely-used deep CNN structures (the VGG-Net~\cite{simonyan2014very}, and ResNet~\cite{he2016deep} ) on both a small dataset (MNIST~\cite{lecun1998gradient})  and large datasets (CIFAR 10/100~\cite{krizhevsky2009learning}). Specifically, our method doesn't need any post-processes like retraining. All experiments were conducted on a NVIDIA Quadro RTX 6000 environment, and our code implementation is based on Pytorch~\cite{NEURIPS2019_bdbca288}.

\begin{table*}[ht] \small
  \centering
  \scalebox{1}{
  \begin{tabular}{cccccccc}
    \toprule
        &SGD   &\multicolumn{6}{c}{Structured directional pruning}\\
    \cmidrule(r){2-2}\cmidrule(r){3-8}
        & no other   & $c=$ 5e-7 & $c=$ 5e-7 & $c=$ 5e-7 & $c=$ 8e-7 &  $c=$ 8e-7 & $c=$ 8e-7\\
        & parameters & $u=0.40$  & $u=0.51$  & $u=0.55$  & $u=0.40$  & $u=0.51$  & $u=0.55$\\
    \midrule
    Train loss  & 0.0001  & 0.0002  & 0.0012  & 0.0026  & 0.0006  & 0.0023  & 0.0027 \\
    Test Acc.   & 0.9089  & 0.9080  & 0.9091  & 0.9090  & 0.9077  & 0.9110  & 0.9127 \\
    Sparsity    & 0.0000  & 0.0000  & 0.0090  & 0.1242  & 0.0084  & 0.4391  & 0.6767 \\
    \bottomrule
  \end{tabular}}
  \caption{The effect of hyper-parameters}
  \label{hyper-table}
\end{table*}

\begin{figure*}[ht]
	\centering
	\includegraphics[height=1.6in]{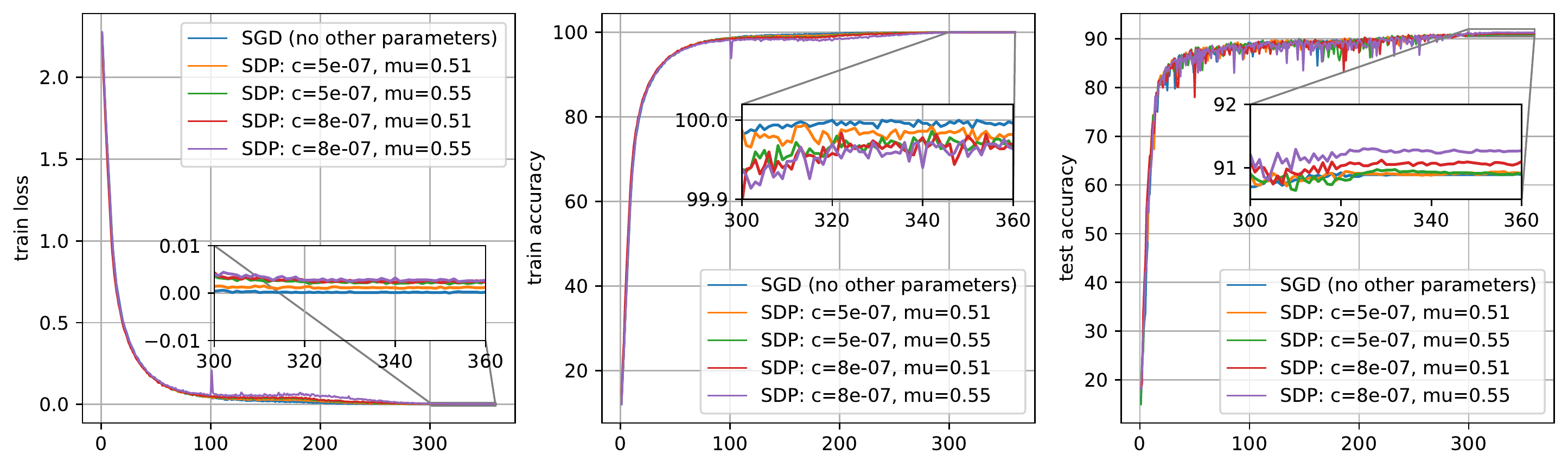}
	\caption{Performance comparison results of {\texttt{AltSDP}} (SDP) and SGD under different hyperparameters on VGG-16, CIFAR-10 task. \textbf{Left:} training loss. \textbf{Center:} training accuracy. \textbf{Right:} testing accuracy. }
	\label{fig_SDP_SGD_cifar10_vgg16}
\end{figure*}

We compare {\texttt{AltSDP}} with different methods that have published results in terms of the test accuracy and Floating-point Operations (FLOPs) reduction. Some methods are reproduced by~\cite{NEURIPS2020_703957b6} and obtain better performance than the originally published ones, then we use the better results in our comparisons with appending a label ``(New)''.  The base ResNet model is implemented following~\cite{he2016deep, NEURIPS2020_703957b6} and the base VGG model is implemented following~\cite{liu2017learning, NEURIPS2020_703957b6}. The detailed parameters for training are list in Appendix. For each method, we present its baseline model accuracy, pruned model accuracy, the accuracy drop between baseline and pruned model, and FLOPs reduction after pruning. A negative accuracy drop indicates that the pruned model performances better than its unpruned baseline model. Specifically, the pruned model accuracy reported for our {\texttt{AltSDP}} is without fine-tune or retraining. More experiments on the WRN28$\times$10 network and MNIST datasets can be found in the Appendix.

\subsection{Performance Comparison Results}\label{sub_set_performance}

Table~\ref{dif-alg-table} shows the performance of different methods on CIFAR datasets, which is the most widely used dataset for pruning task. On CIFAR-10, ResNet-56 task, our method obtains the smallest accuracy drop (-0.10\%) and the best pruned accuracy (93.90\%) with highest FLOPs reduction (55\%). On CIFAR-10, VGG-16 task, our method also obtains the smallest accuracy drop (-0.09\%) and the best pruned accuracy (93.97\%) with highest FLOPs reduction (55\%). Since few structured pruning results on CIFAR-100 dataset are reported in previous works, we only compared with three different algorithms. And as shown in Table~\ref{dif-alg-table}, our method still achieves the smallest accuracy drop (-0.06 for ResNet-56 and -0.46 for VGG-16) and the best pruned accuracy (72.55 for ResNet-56 and 74.29 for VGG-16) under similar FLOPs reduction.

Table~\ref{larg-flops-table} shows the performance of different methods on large FLOPs reduction. Since there exist little related works on structured pruning with large FLOPs reduction, the comparison results are mainly from \cite{NEURIPS2020_a09e75c5}. Specifically, 400 total epochs are used to training and fine-tune/retraining for other methods reproduced in \cite{NEURIPS2020_a09e75c5}, while we only train 200+ epochs for {\texttt{AltSDP}} without retraining. Our method achieves the smallest accuracy drop (0.86\%) and the best pruned accuracy (92.94\%) with highest FLOPs reduction (72\%).

\subsection{Analysis}\label{sub_set_connectivity}
In this Section, we first empirically study the effect of hyper-parameters in {\texttt{AltSDP}}. Then, we train a basic DNN on MNIST, VGG-Net on CIFAR-10 and WRAN20$\times$10 on CIFAR-100 to checking whether {\texttt{AltSDP}} performs structured directional pruning and reaches the same  flat minimum valley obtained by SGD. Similar analysis strategy has been done by \cite{NEURIPS2020_a09e75c5, NEURIPS2018_be3087e7, nguyen2019connected, draxler2018essentially, NEURIPS2019_01d8bae2} and the base VGG model and method for visualizing are implemented following \cite{NEURIPS2018_be3087e7, NEURIPS2020_a09e75c5}.

% \textbf{The Effect of Hyper-parameters} 
We then displays the performance of SGD and {\texttt{AltSDP}} with different hyper-parameter $c$ and $\mu$ on VGG-16, CIFAR-10 task. As shown in Figure~\ref{fig_SDP_SGD_cifar10_vgg16}, the training loss of {\texttt{AltSDP}} is almost the same with SGD (diff. less than 0.003 in Table~\ref{hyper-table}) when pruning,  which implies that {\texttt{AltSDP}} reaches the same flat minimum valley found by SGD. And the test accuracy of {\texttt{AltSDP}} is similar with SGD. Table~\ref{hyper-table} shows more details of Figure~\ref{fig_SDP_SGD_cifar10_vgg16}, where sparsity denotes the non-zero parameter ratio after training. We find that {\texttt{AltSDP}} performs worse than SGD when $\mu=0.40$, but performs better than SGD under other parameter settings. This is reasonable according to Theorem~\ref{theorem_1}, which suggests that $\mu$ should be slightly greater than 0.5. Moreover, as hyperparameters $c$ and $\mu$ become larger, {\texttt{AltSDP}} pushes more parameters to zero and the sparsity becomes larger.

Finally, we check whether {\texttt{AltSDP}} reaches the same valley found by SGD. We train VGG16 on CIFAR-10 until nearly zero training loss using both SGD and {\texttt{AltSDP}}. We use the method of~\cite{garipov2018loss} to search for a quadratic B{\'e}zier curve of minimal training loss connecting the minima found by optimizers.
We can see that {\texttt{AltSDP}} performs the structured directional
pruning since the learned parameters of both SGD and {\texttt{AltSDP}} lie in the same flat minimum valley on the training loss landscape if $\mu$ and $c$ is properly tuned, namely $\mu = 0.51$ and $c = 5\times 10^{-7}$.

\begin{figure}[t]
	\centering
	\subfloat{
	\includegraphics[height=1.2in]{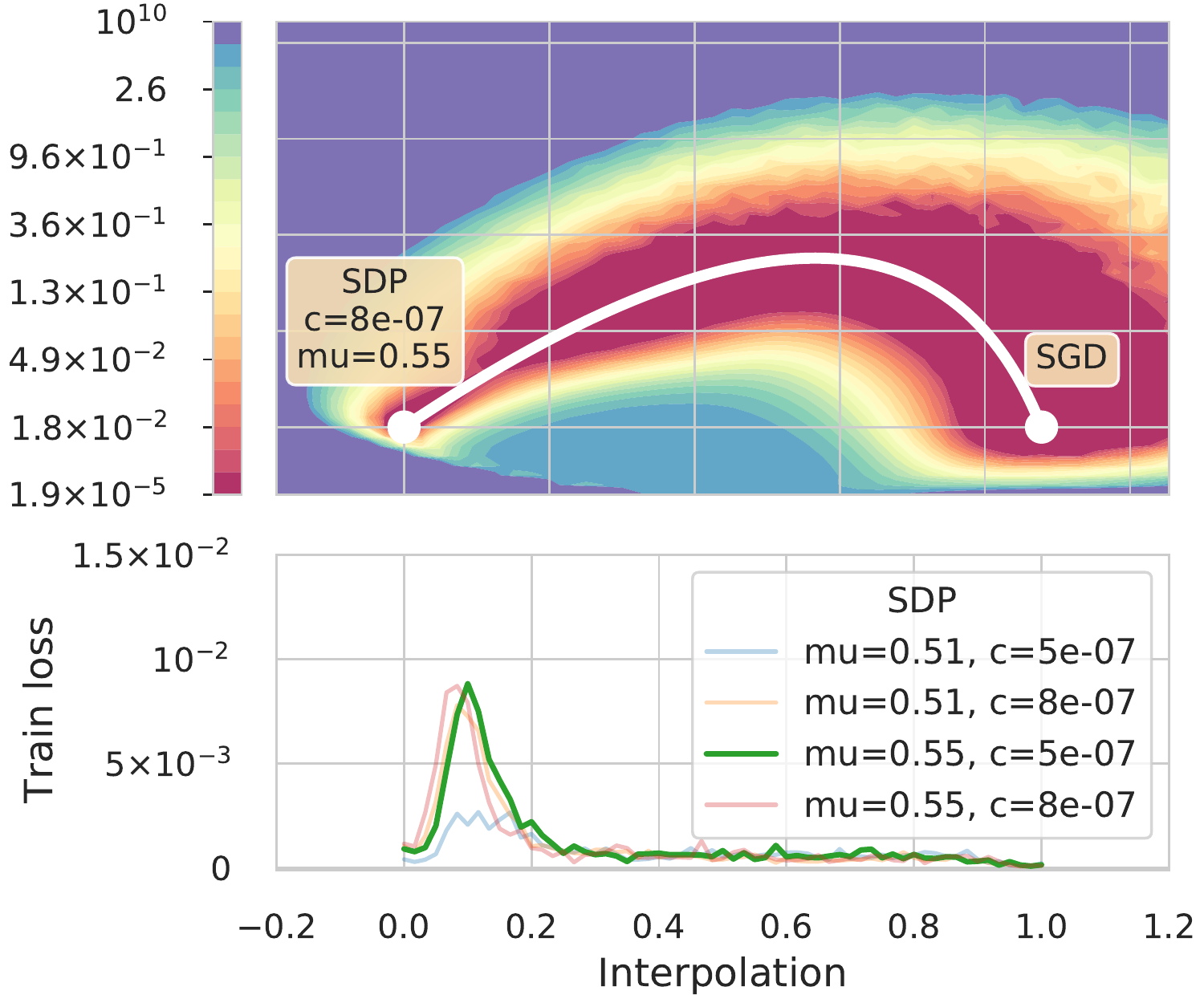}
	}
	\subfloat{
	\includegraphics[height=1.2in]{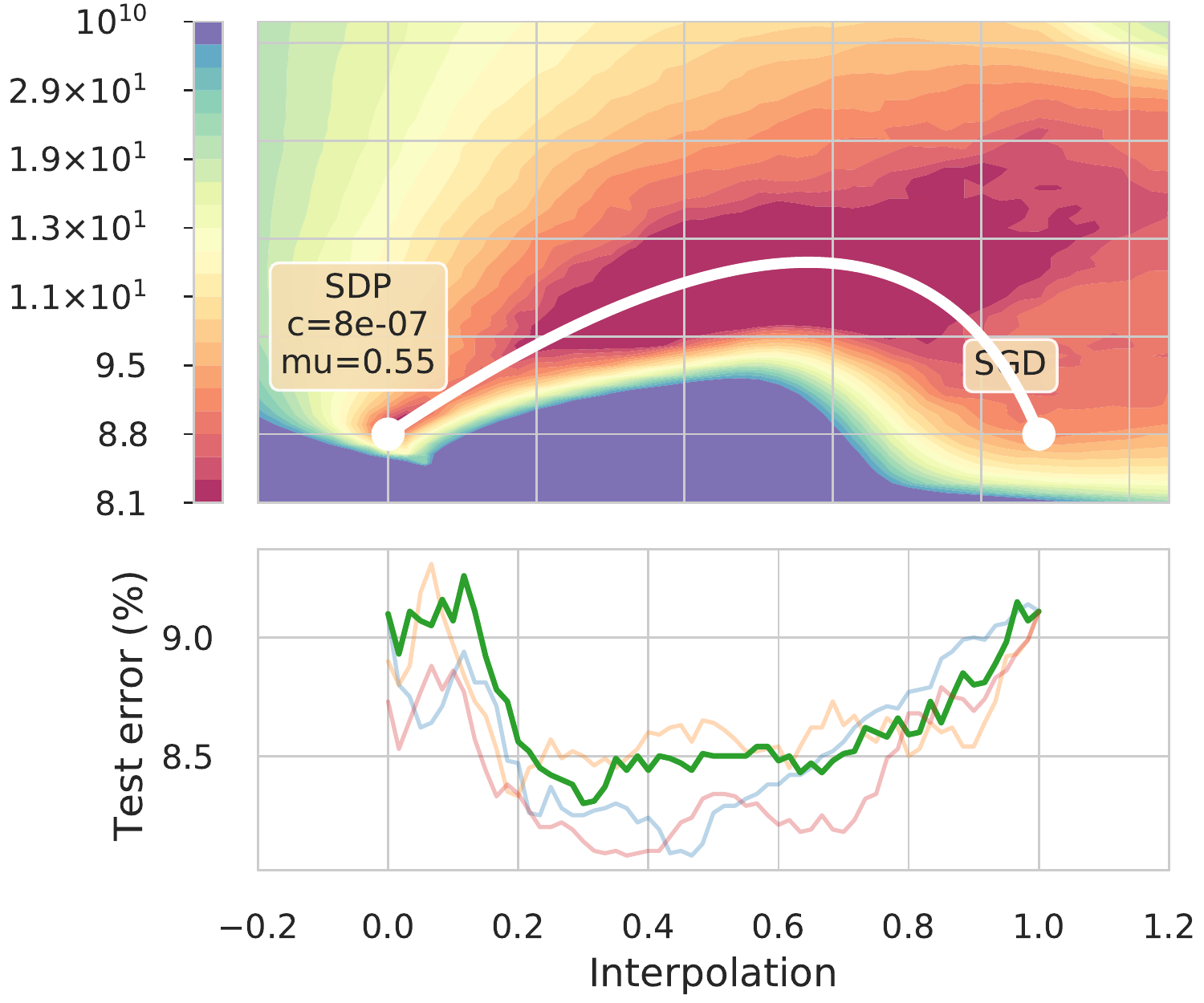}
	}
	\caption{The upper figures show the contour of training loss and testing error on the hyperplane. \textbf{Left:} Train loss on CIFAR-10 and VGG16 task. \textbf{Right:} Test error on CIFAR-10 and VGG16 task. The lower figures show the corresponding white curve (B{\'e}zier curve), which contains the interpolating minimizers of SGD and {\texttt{AltSDP}} under different $\mu$.}
	\label{fig_conn_cifar10_vgg16}
\end{figure}

\section{Conclusions}
In this paper we propose the structured directional pruning method to compress deep neural networks while preserving accuracy, which is based on orthogonal projecting the sparse perturbations onto the flat minimum valley found by optimizers. A fast solver {\texttt{AltSDP}} is also proposed to achieve structured directional pruning. Theoretically, we prove that {\texttt{AltSDP}} achieves directional pruning after sufficient training. Experimentally, we demonstrate the benefits of structured directional pruning and show that it achieves the state-of-the-art result. Experiments using VGG-Net and ResNet on CIFAR-10 and CIFAR-100 datasets show that our method obtains the best pruned accuracy (i.e. 93.97\% on VGG16, CIFAR-10 task) without retraining. Moreover, experiments using DNN, VGG-Net and WRN28$\times$10 on MNIST, CIFAR-10 and CIFAR-100 datasets demonstrate our method performs directional pruning, reaching the same minimal valley as the optimizer.

\bibliographystyle{unsrt} 
\bibliography{references.bib}

\clearpage
\onecolumn
\begin{spacing}{1.0}

\appendix

\section{Proof of Main Theorems}
\subsection{Proof of Theorem~\ref{theorem_solution}}\label{section_auxiliary_theorem}
\textbf{Theorem~\ref{theorem_solution}.}
{\it Consider the optimization problem
\begin{align}
    \label{eq_prop_1}
    \arg\min_{ {\bm w}_{i} } \left\{ \frac{1}{2} \|{\bm w}_{i}^{\star}-{\bm w}_{i}\|_{2}^{2} + \lambda {s}_{i}   \|{\bm w}_{i}\|_{2} \right\}.
\end{align}
For ${\bm w}_{i}^{\star} \in \mathbb{R}^{d}\backslash\{\bm 0\}, {s}_{i} \in \mathbb{R}, \lambda > 0$, \eqref{eq_prop_1} has an explicit solution:
\begin{align}
    \label{eq_prop_2}
    \hat{\bm w}_{i} = \left( 1-\frac{\lambda {s}_{i}}{\|{\bm w}_{i}^{\star}\|_{2}}\right)_{+}{\bm w}_{i}^{\star}.
\end{align}}
\begin{proof}
Let's $\hat{\bm w}_{i}$ denotes the solution for \eqref{eq_prop_1}, we prove $\hat{\bm w}_{i}$ follows the formulation in \eqref{eq_prop_2} from three perspectives: ${s}_{i}=0, {s}_{i}>0$ and ${s}_{i}<0$. Set $f({\bm w}_{i}) := \frac{1}{2} \|{\bm w}_{i}^{\star}-{\bm w}_{i}\|_{2}^{2} + \lambda {s}_{i}   \|{\bm w}_{i}\|_{2}$. First, when ${s}_{i}=0$, the solution $\hat{\bm w}_{i}={\bm w}_{i}^{\star}$. Then, on the one hand, when ${s}_{i}>0$, the objective function is convex, therefore $\nabla f(\hat{\bm w}_{i}) = 0$. We have
    \begin{align}
        \label{eq_prop_3}
        \nabla f(\hat{\bm w}_{i}) = \hat{\bm w}_{i}-{\bm w}_{i}^{\star}+\frac{\lambda {s}_{i}\hat{\bm w}_{i}}{\|\hat{\bm w}_{i}\|_{2}}=0,
    \end{align}
    which yields ${\bm w}_{i}^{\star}=\left(1+\lambda {s}_{i}/\|\hat{\bm w}_{i}\|_{2}\right)\hat{\bm w}_{i}$. Since  $1+\lambda {s}_{i}/\|\hat{\bm w}_{i}\|_{2}>0$ is a scalar, we have ${\bm w}_{i}^{\star}$ and $\hat{\bm w}_{i}$ in the same direction, hence
    \begin{align}
        \label{eq_prop_4}
        \frac{\hat{\bm w}_{i}}{\|\hat{\bm w}_{i}\|_{2}}=\frac{{\bm w}_{i}^{\star}}{\|{\bm w}_{i}^{\star}\|_{2}}.
    \end{align}
    By substituting \eqref{eq_prop_4} into \eqref{eq_prop_3}, we finish the proof for ${s}_{i}>0$.
    
  On the other hand,  when ${s}_{i}<0$, the objective function is not convex, therefore we need to check the value of $f({\bm w}_{i})$ at stationary points. If ${\bm w}_{i}^{\star} = 0$, then $\hat{\bm w}_{i}=0$ is the solution for \eqref{eq_prop_1}.  If ${\bm w}_{i}^{\star} \neq 0$, we have
    \begin{itemize}\item On $\|{\bm w}_{i}\|_{2}=-\lambda s_{i}$, $\nabla f({\bm w}_{i})=-{\bm w}_{i}^{\star} \neq 0$, there is no stationary point.
    
    \item On $\|{\bm w}_{i}\|_{2} > -\lambda s_{i}$, we have $(1+\lambda s_{i}/\|{\bm w}_{i}\|_{2}) > 0$. The stationary point ${\bm w}_{sp1}$ is $(1-\lambda s_{i}/\|{\bm w}_{i}^{\star}\|_{2}) {\bm w}_{i}^{\star}$ with objective function value $f({\bm w}_{sp1})$
    \begin{align}
        f({\bm w}_{sp1})= \frac{1}{2}\left\|\frac{\lambda s_{i}{\bm w}_{i}^{\star}}{\|{\bm w}_{i}^{\star}\|_{2}}\right\|_{2}^{2} + \lambda s_{i} \left\|(1-\frac{\lambda s_{i}}{\|{\bm w}_{i}^{\star}\|_{2}}){\bm w}_{i}^{\star}\right\|_{2} . \notag
    \end{align}
    
    \item On $\|{\bm w}_{i}\|_{2} < -\lambda s_{i}$, we have $(1+\lambda s_{i}/\|{\bm w}_{i}\|_{2}) < 0$. The stationary point ${\bm w}_{sp1}$ is $(1+\lambda s_{i}/\|{\bm w}_{i}^{\star}\|_{2}) {\bm w}_{i}^{\star}$ with objective function value $f({\bm w}_{sp2})$
    \begin{align}
        f({\bm w}_{sp2})&= \frac{1}{2}\left\|\frac{\lambda s_{i}{\bm w}_{i}^{\star}}{\|{\bm w}_{i}^{\star}\|_{2}}\right\|_{2}^{2} + \lambda s_{i} \left\|(1+\frac{\lambda s_{i}}{\|{\bm w}_{i}^{\star}\|_{2}}){\bm w}_{i}^{\star}\right\|_{2}. \notag
    \end{align}
    Since $\lambda s_{i} < 0$ and ${\bm w}_{i}^{\star} \neq 0$, we have
    \begin{align}
        \left\|(1-\frac{\lambda s_{i}}{\|{\bm w}_{i}^{\star}\|_{2}}){\bm w}_{i}^{\star}\right\|_{2} > \left\|(1+\frac{\lambda s_{i}}{\|{\bm w}_{i}^{\star}\|_{2}}){\bm w}_{i}^{\star}\right\|_{2}. \notag
    \end{align}
    Then $ f({\bm w}_{sp1})< f({\bm w}_{sp2})$, which means the global minimizer of $f({\bm w}_{i})$ is the stationary point $(1-\lambda s_{i}) {\bm w}_{i}^{\star}/\|{\bm w}_{i}^{\star}\|_{2}$ on $\|{\bm w}_{i}\|_{2} > -\lambda s_{i}$. We finish the proof for ${s}_{i}>0$.
\end{itemize}
Then we complete the proof of Theorem~\ref{theorem_solution}.
\end{proof}

\subsection{Proof of Theorem~\ref{theorem_1}}

\textbf{Theorem~\ref{theorem_1}.}
{\it
Under Assumptions~\ref{assumption_1}-\ref{assumption_5}, suppose $\mu\in(0.5,1)$ and $c>0$,  when $\gamma\rightarrow 0$,  {\texttt{AltSDP}} achieves structured directional pruning based on ${\bm w}^{SGD}(t)$ asymptotically, i.e., we have for $t>\bar{T}$
\begin{align}\label{eq_sdp_direction_pruning-A1}
    {\bm w}_{\gamma}(t) \overset{d}{\approx}  \arg\min_{{\bm w} \in \mathbb{R}^d} \frac{1}{2}\| {\bm w}^{SGD}(t) - {\bm w} \|_{2}^{2} + \lambda_{\gamma,t} \sum_{i=1}^{|\mathcal{G}|} \bar s_{i}\| {\bm w}_{i} \|_2,
\end{align}
and 
\begin{align}\label{eq_sdp_direction_pruning-A2}
    {\bm w}_{\gamma, i}(t) \overset{d}{\approx}  \left( 1-\frac{\lambda_{\gamma,t} \bar {s}_{i}}{\|{\bm w}_{i}^{SGD}(t)\|_{2}}\right)_{+}{\bm w}_{i}^{SGD}(t),~i = 1,...,|\mathcal{G}|,
\end{align}
where $\lambda_{\gamma,t}=c\sqrt{\gamma}t^{\mu}$;  $\overset{d}{\approx}$ represents ``asymptotic in distribution'' under the empirical probability measure of gradients; and $\bar s_i$ satisfies $\lim_{t\rightarrow \infty} | \bar s_i  - s_i| = 0$ for all $i$.
}

To proof Theorem~\ref{theorem_1}, we first present the following useful Theorem~\ref{theorem_2}, which is proved in Appendix A.3.
\begin{theorem}\label{theorem_2}
Suppose (A\ref{assumption_1}), (A\ref{assumption_2}) and (A\ref{assumption_3}) hold, and assume that the root of the coordinates in ${\bm w}(t)$ occur at time $\{T_{k}\}_{k=1}^{\infty} \subset [0,\infty)$. Let ${\bm w}_{0}$ with ${ w}_{0,j} \neq 0$ (e.g. from a normal distribution) and $T_{0}=0$. Then, as $\gamma$ is small, for $t \in (T_{K},T_{K+1})$,
\begin{align}
    {\bm v}_{\gamma}(t) &\overset{d}{\approx} {\bm w}(t) + \sqrt{\gamma}ct^{\mu}E_{\mathcal{G}}({\bm w}(t)) - \sqrt{\gamma} c\sum_{k=1}^{K}\Big\{\Phi(t,T_{k}) \left\{E_{\mathcal{G}}({\bm w}(T_{k}^{+}))-E_{\mathcal{G}}({\bm w}(T_{k}^{-})) \right\} T_{k}^{\mu} \Big\} \notag\\
    &~~~~~- \sqrt{\gamma}{ c\int_{0}^{t}\Phi(t,s)\frac{\partial E_{\mathcal{G}}({\bm w}(s))s^{\mu}}{\partial s}ds} + \sqrt{\gamma}\int_{0}^{t}\Phi(t,s){\Sigma}^{1/2}({\bm w}(s))d{\bm B}(s)
    \label{theorem3-1}
\end{align}
where $\overset{d}{\approx}$ denotes approximately in distribution, ${\bm B}(s)$ is a $d$-dimensional standard Brownian motion, $\Phi(t,s) \in \mathbb{R}^{d \times d}$ is the principal matrix solution of the matrix ODE system,
\begin{align}
    \label{eq_phi_ode}
    d {\bm x}(t) = -H({\bm w}(t)){\bm x}(t) d t, \quad {\bm x}(t_0) = {\bm x}_{0},
\end{align}
and $\Sigma({\bm w})$ is defined as
\begin{align}
    \Sigma({\bm w}):=\mathbb{E}_{Z}\left[ (\nabla f({\bm w};Z)-G({\bm w}) (\nabla f({\bm w};Z)-G({\bm w})^{T} \right]. \notag
\end{align}
\end{theorem}

Theorem~\ref{theorem_2} presents the distribution dynamics of $\bm v_{\gamma}(t)$ in  \eqref{eq_SDP_SMD_1} with ${\bm v}_{\gamma}(t):={\bm v}_{\left\lfloor t/\gamma \right\rfloor}$.
Next, we start prove Theorem~\ref{theorem_1}, which equals to prove the distribution dynamics of $\bm w_{\gamma,i}(t)$ in  \eqref{eq_SDP_SMD_2} approximately in distribution with $\left( 1-\frac{\lambda_{\gamma,t} \bar {s}_{i}}{\|{\bm w}_{i}^{SGD}\|_{2}} \right){\bm w}_{i}^{SGD}$.

\begin{proof}

To start with, recall that
\begin{align}
    {\bm v}_{n+1} &= {\bm v}_{n}-\gamma \nabla f({\bm w}_{n};{Z}_{n+1})   \tag{{\texttt{AltSDP}}-(a)} \label{eq_SDP_SMD_1_recall}\\
    {\bm w}_{n+1} &= \arg\min_{{\bm w}\in\mathbb{R}^{d}}\{\frac{1}{2}\|{\bm w}\|_{2}^{2} - {\bm w}^{T}{\bm v}_{n+1} + g(n,\gamma)\sum_{i=1}^{|\mathcal{G}|}\|{\bm w}_{i}\|_{2}\} .     \tag{{\texttt{AltSDP}}-(b)}\label{eq_SDP_SMD_2_recall}
\end{align}
To analysis the distribution dynamics of $\bm w_{\gamma,i}(t)$, we first define
\begin{align}
\label{zeta-1}
     \zeta_{\gamma,i}({\bm w}_{\gamma,i}(t)) := \frac{1}{2}\|{\bm w}_{\gamma,i}(t)\|_{2}^{2} + \sqrt{\gamma}ct^{\mu}\|{\bm w}_{\gamma,i}(t)\|_{2}.
\end{align}
Then we have its Fenchel conjugate is given by~\cite{chao2019generalization}
\begin{align}
     \zeta_{\gamma,i}^{*}({\bm v}_{\gamma,i}(t)) =  \max_{{\bm w}_{\gamma,i}(t)}\left\{  {\bm w}_{\gamma,i}(t)^{T}{\bm v}_{\gamma,i}(t)    
     -  \frac{1}{2}\| {\bm w}_{\gamma,i}(t)\|_{2}^{2} - \sqrt{\gamma}ct^{\mu}\|{\bm w}_{\gamma,i}(t)\|_{2}   \right\}, \notag
\end{align}
and the derivative of its Fenchel conjugate is given by~\cite{chao2019generalization}
\begin{align}
\label{d-FC}
     \nabla \zeta_{\gamma,i}^{*}({\bm v}_{\gamma,i}(t)) := \arg \min_{{\bm w}_{\gamma,i}(t)}\left\{\frac{1}{2}\|{\bm w}_{\gamma,i}(t)\|_{2}^{2} + \sqrt{\gamma}ct^{\mu}\|{\bm w}_{\gamma,i}(t)\|_{2} - {\bm w}_{\gamma,i}(t)^{T}{\bm v}_{\gamma,i}(t)\right\}.
\end{align}
Hence, by noting that \eqref{d-FC} is a convex function, let the gradient equals to zero we have 
\begin{align}
\label{d-FC-0grad}
{\bm w}_{\gamma,i}(t)  +  \sqrt{\gamma}ct^{\mu} {\bm w}_{\gamma,i}(t)/ \|{\bm w}_{\gamma,i}(t)\|_{2}   -  {\bm v}_{\gamma,i}(t)  =  0, 
\end{align}
which yields ${\bm v}_{\gamma,i}(t)  =\left(1+\sqrt{\gamma}ct^{\mu}/\|{\bm w}_{\gamma,i}(t)\|_{2}\right){\bm w}_{\gamma,i}(t)$. Since  $(1+\sqrt{\gamma}ct^{\mu}/\|{\bm w}_{\gamma,i}(t)\|_{2})>0$ is a scalar, we have ${\bm v}_{\gamma,i}(t)$ and ${\bm w}_{\gamma,i}(t)$ in the same direction, hence
    \begin{align}
        \label{same-d-2}
        \frac{{\bm v}_{\gamma,i}(t)}{\|{\bm v}_{\gamma,i}(t)\|_{2}}=\frac{{\bm w}_{\gamma,i}(t)}{\|{\bm w}_{\gamma,i}(t)\|_{2}}.
    \end{align}
    By substituting \eqref{same-d-2} into \eqref{d-FC-0grad}, we have
\begin{align}
\label{w-v-relation}
    {\bm w}_{\gamma,i}(t) = \nabla \zeta_{\gamma,i}^{*}({\bm v}_{\gamma,i}(t)) = {\bm v}_{\gamma,i}(t)-\sqrt{\gamma}ct^{\mu}E({\bm v}_{\gamma,i}(t)).
\end{align}
Now, we obtain the relationshop between  ${\bm w}_{\gamma,i}(t)$ and ${\bm v}_{\gamma,i}(t)$ in \eqref{w-v-relation}. Next, we prove \eqref{w-v-relation}  approximately in distribution with $\left( 1-\frac{\lambda_{\gamma,t} \bar {s}_{i}}{\|{\bm w}_{i}^{SGD}\|_{2}} \right){\bm w}_{i}^{SGD}$ based on Theorem~\ref{theorem_2}. Follows by \eqref{theorem3-1} in Theorem~\ref{theorem_2} we have
\begin{align}
    {\bm v}_{\gamma}(t) &\overset{d}{\approx} {\bm w}(t) + \sqrt{\gamma}ct^{\mu}E_{\mathcal{G}}({\bm w}(t)) - \sqrt{\gamma} c\sum_{k=1}^{K}\Big\{\Phi(t,T_{k}) \left\{E_{\mathcal{G}}({\bm w}(T_{k}^{+}))-E_{\mathcal{G}}({\bm w}(T_{k}^{-})) \right\} T_{k}^{\mu} \Big\} \notag\\
    &~~~~~- \sqrt{\gamma} { c\int_{0}^{t}\Phi(t,s)\frac{\partial E_{\mathcal{G}}({\bm w}(s))s^{\mu}}{\partial s}ds}   + \sqrt{\gamma}\int_{0}^{t}\Phi(t,s){\Sigma}^{1/2}({\bm w}(s))d{\bm B}(s)\notag\\
    &=  {\bm w}(t) + \sqrt{\gamma}ct^{\mu}E_{\mathcal{G}}({\bm w}(t)) - \sqrt{\gamma}{\bm \delta(t)} + \sqrt{\gamma}{\bm U}(t), 
    \label{v-distribution-1}
\end{align}
where ${\bm \delta}(t) = {\bm \delta}_{1}(t) + {\bm \delta}_{2}(t)$ and ${\bm U}(t)$ are given by
\begin{align}
    &{\bm \delta}_{1}(t) := c\sum_{k=1}^{K}\Big\{\Phi(t,T_{k}) \left\{E_{\mathcal{G}}({\bm w}(T_{k}^{+}))-E_{\mathcal{G}}({\bm w}(T_{k}^{-})) \right\} T_{k}^{\mu} \Big\}, \notag \\
    &{\bm \delta}_{2}(t) :=  c \int_{0}^{t}\Phi(t,s)\frac{\partial E_{\mathcal{G}}({\bm w}(s))s^{\mu}}{\partial s}ds, \notag \\
    &{\bm U}(t) := \int_{0}^{t}\Phi(t,s){\Sigma}^{1/2}({\bm w}(s))d{\bm B}(s). \notag
\end{align}
By substituting \eqref{v-distribution-1} into \eqref{w-v-relation}, for each $i$ we have
\begin{align}
\label{w-v-relation-2}
    {\bm w}_{\gamma,i}(t) & \overset{d}{\approx}  {\bm w}_i(t) + \sqrt{\gamma}{\bm U}_i(t) -\sqrt{\gamma}{\bm \delta}_{i}(t)+\sqrt{\gamma}ct^{\mu}\left( E({\bm w}_{\gamma,i}(t))-E({\bm v}_{\gamma,i}(t)) \right), \notag\\
 &= {\bm w}_i(t) + \sqrt{\gamma}{\bm U}_i(t) -\sqrt{\gamma}{\bm \delta}_{i}(t),
\end{align}
where the  equality follows by \eqref{same-d-2}. Following the analysis in~\cite{belkin2019reconciling,benveniste2012adaptive},  the piecewise constant process of SGD follows
\begin{align}
%    \label{eq_piecewise_process_0}
    w_{j}^{SGD}(t) \overset{d}{\approx} w_{j}(t) +\sqrt{\gamma}{\bm U}_{j}(t),  ~ j =  {1,2,\cdots,d}, \notag
\end{align}
yields
\begin{align}
%    \label{eq_piecewise_process_1}
    {\bm w}_{i}^{SGD} \overset{d}{\approx} {\bm w}_{i}(t) + \sqrt{\gamma}{\bm U}_{i}(t), ~ i = 1,2,..., |\mathcal{G}|. \notag
\end{align}
Then we have
\begin{align}
\label{w-wSGD-sigma}
    {\bm w}_{\gamma,i}(t)  \overset{d}{\approx} {\bm w}_{i}^{SGD}(t)-\sqrt{\gamma}{\bm \delta}_{i}(t), ~ i = 1,2,..., |\mathcal{G}|.
\end{align}

We next prove for $t \rightarrow \infty$, 
\begin{align}
    {\bm \delta}(t) = ct^{\mu}{\Pi}E_{\mathcal{G}}({\bm w}(t)) + o(t^{\mu}) + O(t^{\mu-1}). \notag
\end{align}
To obtain this, we need to find the principal matrix solution $\Phi(t,s)$ in ${\bm \delta}_{i}(t)$. Recall \eqref{eq_ode_system} that 
 \begin{align}
     \frac{d\Phi(t,s)}{dt}=-H({\bm w}(t))\Phi(t,s),\quad \Phi(s,s) = I_{d}. \notag
 \end{align}
Following the Levinson theorem~\cite{eastham1985asymptotic},  when $a_{t} \rightarrow 0$, there exists a real symmetric matrix ${\bar H} = P\Lambda P^{T} $ satisfying
\begin{align}
    \int_{t}^{\infty} \|H({\bm w}(s)) - {\bar H} \|ds = O(a_{t}), \notag
\end{align}
where $\Lambda = {\rm diag}(\lambda_1,\cdots,\lambda_d)$ is a diagonal matrix with non-negative values and $P$ is an orhonormal matrix with its column vectors are eigenvectors $\bm u_j$. 
Following the proof in \cite{NEURIPS2020_a09e75c5} and Levinson theorem in \cite{eastham1985asymptotic}, we get the principal matrix solution $\Phi(t,s)$ in \eqref{eq_phi_ode} satisfies
\begin{align}
    \label{eq_phi_levinson}
    \Phi(\tau,s) = P(I_{d}+O(a_{\tau}))e^{-\Lambda (\tau - s)}P^{T} = P_{0}P_{0}^{T} + O(e^{-\underline{\lambda}(\tau -s)}) + O(a_{\tau}),
\end{align}
where $\underline{\lambda}$ is the least positive eigenvalue of ${\bar H}$, the column vectors of $P_0$ are eigenvectors associated with the zero eigenvalue, i.e., $P_{0}P_{0}^{T} = \sum_{j:\bar H \bm u_j = 0} \bm u_j \bm u_j^T$. Then we have
\begin{align}
    \label{eq_delta1_0}
    {\bm \delta}_{1}(t) &= c\sum_{k=1}^{K}\Big\{\Phi(t,T_{k}) \left\{E_{\mathcal{G}}({\bm w}(T_{k}^{+}))-E_{\mathcal{G}}({\bm w}(T_{k}^{-})) \right\} T_{k}^{\mu} \Big\}\notag\\
    &= cP_{0}P_{0}^{T}\sum_{k=1}^{K}\Big\{\left\{E_{\mathcal{G}}({\bm w}(T_{k}^{+}))-E_{\mathcal{G}}({\bm w}(T_{k}^{-})) \right\} T_{k}^{\mu} \Big\}+ O(e^{-\underline{\lambda}(t -T_{K})}T_{k}^{\mu}) + O(a_{t}T_{k}^{\mu}),
\end{align}
where the first term of $\delta_{1}(t)$ can be rewritten as
\begin{align}
    \label{eq_delta1_1}
    &~~~~~cP_{0}P_{0}^{T}\sum_{k=1}^{K}\Big\{\left\{E_{\mathcal{G}}({\bm w}(T_{k}^{+}))-E_{\mathcal{G}}({\bm w}(T_{k}^{-})) \right\} T_{k}^{\mu} \Big\}\notag\\
    &= cP_{0}P_{0}^{T}\left\{E_{\mathcal{G}}({\bm w}(T_{K})^{+})T_{K}^{\mu} - \sum_{k=1}^{K}\left\{E_{\mathcal{G}}({\bm w}(T_{k}^{-}))T_{k}^{\mu} - E_{\mathcal{G}}({\bm w}(T_{k-1}^{+}))T_{k-1}^{\mu} \right\} \right\}\notag\\
    &= cP_{0}P_{0}^{T}\left\{E_{\mathcal{G}}({\bm w}(T_{K}^{+}))T_{K}^{\mu} -E_{\mathcal{G}}({\bm w}(t))t^{\mu}\big|_{T_{K-1}}^{T_{K}} - \cdots -E_{\mathcal{G}}({\bm w}(t))t^{\mu}\big|_{T_{0}}^{T_{1}} \right\} \notag\\
    &= cP_{0}P_{0}^{T}\left\{E_{\mathcal{G}}({\bm w}(T_{K}^{+}))T_{K}^{\mu} - 
    \int_{T_{K-1}}^{T_{K}} \frac{\partial        E_{\mathcal{G}}({\bm w}(s))s^{\mu}         }{   \partial s  }   ds - \cdots -
    \int_{0}^{T_{1}} \frac{\partial        E_{\mathcal{G}}({\bm w}(s))s^{\mu}         }{   \partial s  }   ds \right\} \notag\\
    &= cP_{0}P_{0}^{T}\left\{ E_{\mathcal{G}}({\bm w}(T_{K}^{+}))T_{K}^{\mu} - 
    \int_{0}^{T_{K}} \frac{\partial        E_{\mathcal{G}}({\bm w}(s))s^{\mu}         }{   \partial s  }     ds \right\}.
\end{align}

Combining \eqref{eq_delta1_1} and \eqref{eq_delta1_0}, we get
\begin{align}
    {\bm \delta}_{1}(t) = cP_{0}P_{0}^{T}\left\{ E_{\mathcal{G}}({\bm w}(T_{K}^{+}))T_{K}^{\mu} - 
    \int_{0}^{T_{K}} \frac{\partial        E_{\mathcal{G}}({\bm w}(s))s^{\mu}         }{   \partial s  }     ds \right\} + 
    O(e^{-\underline{\lambda}(t -T_{K})}T_{k}^{\mu}) + O(a_{t}T_{k}^{\mu}). \notag
\end{align}

Then by substituting \eqref{eq_phi_levinson} into $\bm \delta_{2}(t)$, we next obtain
\begin{align}
    \label{eq_delta2_0}
    {\bm \delta}_{2}(t) &= c \int_{0}^{t}\Phi(t,s)\frac{\partial E_{\mathcal{G}}({\bm w}(s))s^{\mu}}{\partial s}ds \notag\\
    &= c  P_{0}P_{0}^{T}
    \int_{0}^{t}   \frac{\partial E_{\mathcal{G}}({\bm w}(s))s^{\mu}}{\partial s}ds
    + O\left( c \int_{0}^{t}      e^{-\underline{\lambda}(t-s)}        \frac{\partial E_{\mathcal{G}}({\bm w}(s))s^{\mu}}{\partial s}      ds \right) \notag\\
    &~~~~~~~~~~~~~~~~~~~~~~~~~~~~~~~~~~~~~~~~~~~~~~~~~~~~~~~~+O\left( c   \int_{0}^{t}     a_{t}    \frac{\partial E_{\mathcal{G}}({\bm w}(s))s^{\mu}}{\partial s}       ds \right)
\end{align}
where $\left\| E_{\mathcal{G}}({\bm w}(s)) \right\| = |\mathcal{G}|^{1/2}$. Then we have
\begin{align}
    O\left(  c \int_{0}^{t}      e^{-\underline{\lambda}(t-s)}        \frac{\partial E_{\mathcal{G}}({\bm w}(s))s^{\mu}}{\partial s}      ds \right)
    &  \lesssim  
    c \int_{0}^{t}      e^{-\underline{\lambda}(t-s)}        \frac{\left\|  \partial E_{\mathcal{G}}({\bm w}(s)) \right\|   s^{\mu}}{\partial s}      ds    \notag \\
    &  = 
     c   |\mathcal{G}|^{1/2} \int_{0}^{t}     e^{-\underline{\lambda}(t-s)}        \frac{\partial s^{\mu}}{\partial s} ds  \notag \\
 & \leq    c  \mu d^{1/2}\int_{0}^{t}     s^{\mu-1}e^{-\underline{\lambda}(t-s)}ds \overset{(a)}{=}  O(t^{\mu-1}) \label{eq_delta2_1} 
 \end{align}
 and
\begin{align}
    & O\left( c   \int_{0}^{t}     a_{t}    \frac{\partial E_{\mathcal{G}}({\bm w}(s))s^{\mu}}{\partial s}       ds \right)    \lesssim     c\mu |\mathcal{G}|^{1/2}\int_{0}^{t}s^{\mu-1}a_{t}ds = O(a_{t}t^{\mu}) \label{eq_delta2_2}
\end{align}
where $(a)$ follows by using the similar arguments as the proof of Theorem 4.2 in \cite{chao2019generalization} with $\mu \in (0.5,1)$. Combining  \eqref{eq_delta2_0}, \eqref{eq_delta2_1} and \eqref{eq_delta2_2} we get
\begin{align}
    {\bm \delta}_{2}(t) =  c  P_{0}P_{0}^{T}
    \int_{0}^{t}   \frac{\partial E_{\mathcal{G}}({\bm w}(s))s^{\mu}}{\partial s}ds + O(t^{\mu-1}) + O(a_{t}t^{\mu}). \notag
\end{align}

Note that $a_{t}t^{\mu} > a_{t}T_{K}^{\mu}$ for $\mu > 0, t > T_{K}$ and $e^{-\underline{\lambda}(t-T_{K})} \rightarrow 0$. Then
\begin{align}
%    \label{eq_delta_final}
    {\bm \delta}(t) &= {\bm \delta}_{1}(t) + {\bm \delta}_{2}(t)\notag\\
    &=  cP_{0}P_{0}^{T}\left\{ E_{\mathcal{G}}({\bm w}(T_{K}^{+}))T_{K}^{\mu} - 
    \int_{0}^{T_{K}} \frac{\partial        E_{\mathcal{G}}({\bm w}(s))s^{\mu}         }{   \partial s  }     ds
    + \int_{0}^{t}   \frac{\partial E_{\mathcal{G}}({\bm w}(s))s^{\mu}}{\partial s}ds  \right\}      \notag\\
    &~~~~~~~~+ O(t^{\mu-1}) + O(a_{t}t^{\mu}) \notag\\
    &=cP_{0}P_{0}^{T}\left\{ E_{\mathcal{G}}({\bm w}(T_{K}^{+}))T_{K}^{\mu} + 
    \int_{T_{K}}^{t}   \frac{\partial          E_{\mathcal{G}}({\bm w}(s))s^{\mu}          }{   \partial s  }        ds 
    \right\} + O(t^{\mu-1}) + O(a_{t}t^{\mu}) \notag\\
    &=cP_{0}P_{0}^{T}\left\{ E_{\mathcal{G}}({\bm w}(T_{K}^{+}))T_{K}^{\mu} + 
        E_{\mathcal{G}}({\bm w}(s))s^{\mu}        \big|_{T_{K}}^{t}
    \right\} + O(t^{\mu-1}) + O(a_{t}t^{\mu}) \notag\\
    &= cP_{0}P_{0}^{T}E_{\mathcal{G}}({\bm w}(t))t^{\mu} + O(t^{\mu-1}) + O(a_{t}t^{\mu}) \notag\\
    &\overset{(a)}{=} c{\Pi}E_{\mathcal{G}}({\bm w}(t))t^{\mu} + O(t^{\mu-1}) + o(t^{\mu}), \notag
\end{align}
where ${\Pi} = P_{0}P_{0}^{T}$ and $(a)$ is due to
\begin{align}
    \lim_{t \rightarrow 0} \frac{a_{t}t^{\mu}}{t^{\mu}} = \lim_{t \rightarrow 0} a_{t} = 0. \notag
\end{align}

Then set ${s_{i}} = E({\bm w}(t))\cdot\left\{{\Pi}E_{\mathcal{G}}({\bm w}(t))\right\}_{i}$, we get
\begin{align}
    \label{eq_delta_a_final}
    {\bm \delta}_{i}(t) = ct^{\mu}s_{i}E({\bm w}_i(t)) + o(t^{\mu}) + O(t^{\mu-1}). 
\end{align}

To this end, by substituting \eqref{eq_delta_a_final} into \eqref{w-wSGD-sigma}, we have
\begin{align}
%\label{w-wSGD-sigma-2}
    {\bm w}_{\gamma,i}(t)  &\overset{d}{\approx} {\bm w}_{i}^{SGD}(t)-\sqrt{\gamma}  \left( 
    ct^{\mu}s_{i}E({\bm w}_i(t)) + o(t^{\mu}) + O(t^{\mu-1})
    \right), ~ i = 1,2,..., |\mathcal{G}|  \notag\\
    &= {\bm w}_{i}^{SGD}(t)-\sqrt{\gamma}  
    ct^{\mu}\bar s_{i}E({\bm w}_i(t)) 
     , ~ i = 1,2,..., |\mathcal{G}|, \notag
\end{align}
where $\bar s_{i} =   s_{i}+ \frac{o(t^{\mu})}{ c \sqrt{\gamma} t^{\mu} } + \frac{O(t^{\mu-1})}{ c \sqrt{\gamma} t^{\mu} } $ and hence $\lim_{t\rightarrow \infty} | \bar s_i  - s_i| = 0$. Based on Assumption~6 we can further have
\begin{align}
%\label{w-wSGD-sigma-2}
    {\bm w}_{\gamma,i}(t)  \overset{d}{\approx} 
     {\bm w}_{i}^{SGD}(t)-  
    c \sqrt{\gamma}  t^{\mu}\bar s_{i}E({\bm w}_i^{SGD}(t)) 
     , ~ i = 1,2,..., |\mathcal{G}|, \notag
\end{align}
hence we obtain \eqref{eq_sdp_direction_pruning-A2}. Based on Theorem~\ref{theorem_solution}, we further obtain \eqref{eq_sdp_direction_pruning-A1}. Then we complete the proof.
\end{proof}

\subsection{Proof of Theorem~\ref{theorem_2}}

\begin{proof}

To prove
\begin{align}
    {\bm v}_{\gamma}(t) &\overset{d}{\approx} {\bm w}(t) + \sqrt{\gamma}ct^{\mu}E_{\mathcal{G}}({\bm w}(t)) - \sqrt{\gamma} c\sum_{k=1}^{K}\Big\{\Phi(t,T_{k}) \left\{E_{\mathcal{G}}({\bm w}(T_{k}^{+}))-E_{\mathcal{G}}({\bm w}(T_{k}^{-})) \right\} T_{k}^{\mu} \Big\} \notag\\
    &~~~~~- \sqrt{\gamma}{ c\int_{0}^{t}\Phi(t,s)\frac{\partial E_{\mathcal{G}}({\bm w}(s))s^{\mu}}{\partial s}ds} + \sqrt{\gamma}\int_{0}^{t}\Phi(t,s){\Sigma}^{1/2}({\bm w}(s))d{\bm B}(s), \notag
\end{align}
we define the centered and scaled processes
\begin{align}
    \label{eq_centered_scaled}
    {\bm V}_{\gamma}(t) := \frac{{\bm v}_{\gamma}(t) - {\bm w}(t)}{\sqrt{\gamma}},
\end{align}
then we need to prove
\begin{align}
%    \label{eq_v_begin-1}
    {\bm V}_{\gamma}(t)  &\overset{d}{\approx} ct^{\mu}E_{\mathcal{G}}({\bm w}(t)) - c\sum_{k=1}^{K}\Big\{\Phi(t,T_{k}) \left\{E_{\mathcal{G}}({\bm w}(T_{k}^{+}))-E_{\mathcal{G}}({\bm w}(T_{k}^{-})) \right\} T_{k}^{\mu} \Big\} \notag\\
    &~~~~~- c\int_{0}^{t}\Phi(t,s)\frac{\partial E_{\mathcal{G}}({\bm w}(s))s^{\mu}}{\partial s}ds + \int_{0}^{t}\Phi(t,s){\Sigma}^{1/2}({\bm w}(s))d{\bm B}(s). \notag
\end{align}
By Theorem 3.13 in \cite{chao2019generalization}, ${\bm V}_{\gamma} \overset{d}{\approx} {\bm V}$ on $(T_{k},T_{k+1})$ for each $k=0,\dots,K$ as $\gamma$ is small, where $V$ obeys the stochastic differential equation (SDE): 
\begin{align}
    \label{eq_SDE}
    d {\bm V}(t)= -H({\bm w}(t))        \nabla\tilde {\zeta}^{*}(\bm V(t))        dt + {\Sigma}^{1/2}({\bm w}(t))d{\bm B}(t),
\end{align}
where the initial ${\bm V}(T_k)={\bm V}(T_k^{-})$, ${\bm B}(t)$ is the $d$-dimensional standard Brownian motion, and $\tilde {\zeta}^{*}(\bm V(t))$ is the Fenchel conjugate of $\tilde{\zeta}({\bm W}(t))$.  
The function $\tilde{\zeta}(\cdot)$ is defined as $\tilde{\zeta}({\bm w}(t)) := \lim_{\gamma \rightarrow 0} \tilde {\zeta}_{\gamma}(\bm w(t))$ with $\tilde {\zeta}_{\gamma}(\bm w(t))$ being the local Bregman divergence of $ {\zeta}_{\gamma}(\bm w(t))$  in \eqref{zeta-1}  at $(\bm v(t), \bm w(t))$. In particular, we have
\begin{align}
    \tilde {\zeta}_{\gamma}(\bm u(t)) &:= {\gamma}^{-1}\bigg( {\zeta}_{\gamma}({\bm w}(t)+\sqrt{\gamma}{\bm u}(t))-{\zeta}_{\gamma}({\bm w}(t))-\left\langle \sqrt{\gamma}{\bm u}(t), {\bm w}(t) \right\rangle\bigg)\notag\\
    &= {\gamma}^{-1} \bigg( \frac{1}{2}\|{\bm w}(t)+\sqrt{\gamma}{\bm u}(t)\|^{2}_2 -\frac{1}{2}\|{\bm w}(t)\|^{2}_2 - \left\langle \sqrt{\gamma}{\bm u}(t), {\bm w}(t) \right\rangle \notag\\
    &~~~~~~~~~~~~~~~~~~~~~~~~~~+{g(\left\lfloor t/\gamma \right\rfloor,\gamma)}\sum_{i \in \mathcal{G}}   \left[        \|{\bm w}_{i}(t)+\sqrt{\gamma}{\bm u}_{i}(t)\|_{2} - \|{\bm w}_{i}(t)\|_{2} \right]        \bigg)\notag\\
    &= \frac{1}{2}\|{\bm u}(t)\|_{2}^{2} + {\gamma}^{-1}\bigg(       {g(\left\lfloor t/\gamma \right\rfloor,\gamma)}\sum_{i \in \mathcal{G}}   \left[        \|{\bm w}_{i}(t)+\sqrt{\gamma}{\bm u}_{i}(t)\|_{2} - \|{\bm w}_{i}(t)\|_{2} \right]           \bigg),  \notag
\end{align}
where $g(\left\lfloor t/\gamma \right\rfloor,\gamma) := c{\gamma}^{1/2}(n\gamma)^{\mu} $ with $c>0$ and $n = \left\lfloor t/\gamma \right\rfloor$. Then we have
\begin{align}
    \tilde{ \zeta }({\bm w}(t)) &:= \lim_{\gamma \rightarrow 0} \tilde { \zeta }_{\gamma}(\bm w(t))\notag\\
    &= \lim_{\gamma \rightarrow 0}         \left\{            
         \frac{1}{2}\|{\bm w}(t)\|_{2}^{2} + {\gamma}^{-1}\bigg( {g(\left\lfloor t/\gamma \right\rfloor,\gamma)}\sum_{i \in \mathcal{G}}(\|(1+\sqrt{\gamma}){\bm w}_{i}(t)\|_{2} - \|{\bm w}_{i}(t)\|_{2})\bigg)                  
          \right\}       \notag\\
    &= \lim_{\gamma \rightarrow 0}                    \left\{      
    \frac{1}{2}\|{\bm w}(t)\|_{2}^{2} + c(n\gamma)^{\mu} \sum_{i \in \mathcal{G}}\|{\bm w}_{i}(t)\|_{2}
    \right\}              \notag\\
    &= \frac{1}{2}\|{\bm w}(t)\|_{2}^{2} + ct^{\mu} \sum_{i \in \mathcal{G}}\|{\bm w}_{i}(t)\|_{2}.  \notag
\end{align}
Hence, the derivative of $\tilde{ \zeta }^*({\bm v}(t))$ satisfies
\begin{align}
    \nabla\tilde{ \zeta }^*({\bm v}(t)) &= \arg \min_{{\bm w(t)} \in \mathbb{R}^d}\left\{ \tilde { \zeta }(\bm w(t)) - {\bm w}(t)^{T}{\bm v(t)}\right\} \notag\\
    &=\arg \min_{{\bm w(t)} \in \mathbb{R}^d}\left\{ \frac{1}{2}\|{\bm w(t)}\|_{2}^{2} + ct^{\mu}\sum_{i \in \mathcal{G}}\|{\bm w}_{i}(t)\|_{2} - {\bm w(t)}^{T}{\bm v}(t)\right\}, \notag
\end{align}
where $\nabla\tilde{\zeta}^{*}({\bm v}(t)) = \left[            \nabla\tilde{\zeta}_1^{*}({\bm v}(t))^T  ,   \nabla\tilde{\zeta}_2^{*}({\bm v}(t))^T  ,  ... ,   \nabla\tilde{\zeta}_{|\mathcal{G}|}^{*}( {\bm v}(t) ) ^T          \right]^T$ and we have
\begin{align}
\label{nobal-tilde-zeta}
    \nabla\tilde{\zeta}_i^{*}({\bm v}(t)) &= \arg \min_{{\bm w}_{i}}\left\{ \frac{1}{2}\|{\bm w_{i}(t)}\|_{2}^{2} + ct^{\mu}\|{\bm w}_{i}(t)\|_{2} - {\bm w_{i}(t)}^{T}{\bm v}_{i}(t)\right\}\notag\\
    &\overset{(a)}{=} {\bm v}_{i}(t) - ct^{\mu}{E}_{i}({\bm w}_{i}(t)),
\end{align}
where $(a)$ follows by $\{\frac{1}{2}\|{\bm w_{i}(t)}\|_{2}^{2} + ct^{\mu}\|{\bm w}_{i}(t)\|_{2} - {\bm w_{i}(t)}^{T}{\bm v}_i(t)\}$ with $c>0$ is a convex function, the minimizer is at the point when its gradient equals to zero.
Substitute $\nabla\tilde {\zeta}^{*}(\bm V(t))$ according to \eqref{nobal-tilde-zeta} into \eqref{eq_SDE}, we have
\begin{align}
    \label{eq_SDE-2}
    d {\bm V}(t)= -H({\bm w}(t))[{\bm V}(t) - E_{\mathcal{G}}({\bm w}(t))ct^{\mu}]dt + {\Sigma}^{1/2}({\bm w}(t))d{\bm B}(t). 
\end{align}

Next, based on Assumptions 3 and 4, the solution operator $\Phi(t,s)$ of the inhomogeneous ODE system
\begin{align}
    d {\bm x}(t) = -H({\bm w}(t)){\bm x}(t) d t, \quad {\bm x}(t_0) = {\bm x}_{0}   \notag
\end{align}
uniquely exists, and the solution is ${\bm x}(t) = \Phi(t,t_0){\bm x}_{0}$ by Theorem 5.1 of \cite{pazy2012semigroups} and for $0<s<m<t$,
\begin{align}
    (s,t) &\mapsto \Phi(t,s)~\text{is continuous}, \label{eq_phi_contious}\\
    \Phi(t,t) &= I_d, \label{eq_phi_tt}\\
    \frac{\partial}{\partial t}\Phi(t,s) &= -H({\bm w}(t))\Phi(t,s), \label{eq_phi_par_t}\\
    \frac{\partial}{\partial s}\Phi(t,s) &= \Phi(t,s)H({\bm w}(s)), \label{eq_phi_par_s}\\
    \Phi(t,s) &= \Phi(t,m)\Phi(m,s). \label{eq_phi_rec}
\end{align}
Then \eqref{eq_SDE-2} can be verified by \eqref{eq_phi_par_t} and Ito calculus for $t\in(T_k,T_{k+1})$ is given by
\begin{align}
%    \label{eq_ito_v_0}
    {\bm V}(t) 
    &= \Phi(t,t_0){\bm V}_0 + \int_{T_k}^{t}\Phi(t,s)H({\bm w}(s))E_{\mathcal{G}}({\bm w}(s))cs^{\mu}ds + \int_{T_k}^{t}\Phi(t,s){\Sigma}^{1/2}({\bm w}(s))d{\bm B}(s),  \notag
\end{align}
since $\frac{\partial \Phi(t,t_0){\bm V}_0}{ \partial t}  =  -H({\bm w}(t))\Phi(t,t_0) {\bm V}_0  = -H({\bm w}(t))  {\bm V}(t) $ and we assume that the root of the coordinates in ${\bm w}(t)$ occur at time $\{T_{k}\}_{k=1}^{\infty} \subset [0,\infty)$. 
By substituting with initial $t_0 = T_k$ and ${\bm V}(T_k)={\bm V}(T_k^{-})$, we have
\begin{align}
    \label{eq_ito_v}
    {\bm V}(t) = \Phi(t,T_k){\bm V}(T_k^{-}) + \int_{T_k}^{t}\Phi(t,s)H({\bm w}(s))E_{\mathcal{G}}({\bm w}(s))cs^{\mu}ds + \int_{T_k}^{t}\Phi(t,s){\Sigma}^{1/2}({\bm w}(s))d{\bm B}(s)
\end{align}
is the solution of \eqref{eq_SDE-2}. Note that ${\bm V}(T_0)={\bm V}(0)={\bm V}_{\gamma}(0)=0$ almost surely.

Set $d\Delta_{1}(s) = H({\bm w}(s))E_{\mathcal{G}}({\bm w}(s))cs^{\mu}ds$ and $d\Delta_{2}(s) = {\Sigma}^{1/2}({\bm w}(s))d{\bm B}(s)$. If $t>T_K$, we have
\begin{align}
    \label{eq_v_0}
    &~~~~~{\bm V}(t) \notag\\
    &= \Phi(t,T_K){\bm V}(T_K^{-}) + \int_{T_K}^{t}\Phi(t,s)d\Delta_{1}(s) + \int_{T_K}^{t}\Phi(t,s)d\Delta_{2}(s)\notag\\
    &\overset{(a)}{=} \Phi(t,T_K)\left\{\Phi(T_{K},T_{K-1}){\bm V}(T_{K-1}^{-}) + \int_{T_{K-1}}^{T_{K}}\Phi(T_{K},s)d\Delta_{1}(s) + \int_{T_{K-1}}^{T_{K}}\Phi(T_{K},s)d\Delta_{2}(s)\right\} \notag\\
    &~~~~~~~~~~~+\int_{T_K}^{t}\Phi(t,s)d\Delta_{1}(s) + \int_{T_K}^{t}\Phi(t,s)d\Delta(s)\notag\\
    &\overset{(b)}{=}\Phi(t,T_{K-1}){\bm V}(T_{K-1}^{-}) + \int_{T_{K-1}}^{t}\Phi(t,s)d\Delta_{1}(s) + \int_{T_{K-1}}^{t}\Phi(t,s)d\Delta_{2}(s)\notag\\
    &~~~~~~~~~~~\vdots \notag\\
    &=\Phi(t,0){\bm V}(0) + \int_{0}^{t}\Phi(t,s)d\Delta_{1}(s) + \int_{0}^{t}\Phi(t,s)d\Delta_{2}(s)\notag\\\
    &\overset{(c)}{=}\int_{0}^{t}\Phi(t,s)d\Delta_{1}(s) + \int_{0}^{t}\Phi(t,s)d\Delta_{2}(s),
\end{align}
where $(a)$ is by unfolding ${\bm V}(T_K^{-})$ according to \eqref{eq_ito_v} with $k=K-1$, $(b)$ follows by \eqref{eq_phi_rec} and $(c)$ is due to ${\bm V}(T_0)={\bm V}(0)=0$.

We next analysis the first term in \eqref{eq_v_0}, which can be rewritten as
\begin{align}
    \label{eq_delta_1}
    &~~~~~\int_{0}^{t}\Phi(t,s)d\Delta_{1}(s) \notag\\
    &= \int_{0}^{t}\Phi(t,s)H({\bm w}(s))E_{\mathcal{G}}({\bm w}(s))cs^{\mu}ds \notag\\
    &\overset{(a)}{=} \int_{0}^{t}\frac{\partial}{\partial s}\Phi(t,s)E_{\mathcal{G}}({\bm w}(s))cs^{\mu}ds \notag\\
    &= \Phi(t,s)E_{\mathcal{G}}({\bm w}(s))cs^{\mu}\Big|_{0}^{t} - { c\int_{0}^{t}\Phi(t,s)\frac{\partial E_{\mathcal{G}}({\bm w}(s))s^{\mu}}{\partial s}ds}.
\end{align}
where $(a)$ follows by \eqref{eq_phi_par_s}, and the first term in \eqref{eq_delta_1} can be further rewritten as
\begin{align}
    \label{eq_delta_1_p1}
    &~~~~~\Phi(t,s)E_{\mathcal{G}}({\bm w}(s))cs^{\mu}\Big|_{0}^{t}\notag\\
    &= \Phi(t,s)E_{\mathcal{G}}({\bm w}(s))cs^{\mu}\Big|_{T_{K}}^{t} + \Phi(t,s)E_{\mathcal{G}}({\bm w}(s))cs^{\mu}\Big|_{T_{K-1}}^{T_{K}} + \cdots+\Phi(t,s)E_{\mathcal{G}}({\bm w}(s))cs^{\mu}\Big|_{0}^{T_{1}}\notag\\
    &\overset{(a)}{=} \Phi(t,t)E_{\mathcal{G}}({\bm w}(t))ct^{\mu} - \sum_{k=1}^{K}\Big\{\Phi(t,T_{k}) \left\{E_{\mathcal{G}}({\bm w}(T_{k}^{+}))-E_{\mathcal{G}}({\bm w}(T_{k}^{-})) \right\} cT_{k}^{\mu} \Big\}\notag\\
    &\overset{(b)}{=} ct^{\mu}E_{\mathcal{G}}({\bm w}(t)) - c\sum_{k=1}^{K}\Big\{\Phi(t,T_{k}) \left\{E_{\mathcal{G}}({\bm w}(T_{k}^{+}))-E_{\mathcal{G}}({\bm w}(T_{k}^{-})) \right\} T_{k}^{\mu} \Big\},
\end{align}
where $(a)$ is due to \eqref{eq_phi_contious} and $(b)$ is due to \eqref{eq_phi_tt}.

Combining \eqref{eq_v_0}, \eqref{eq_delta_1} and \eqref{eq_delta_1_p1}, we have
\begin{align}
%    \label{eq_v_final}
    {\bm V}(t) &= ct^{\mu}E_{\mathcal{G}}({\bm w}(t)) - c\sum_{k=1}^{K}\Big\{\Phi(t,T_{k}) \left\{E_{\mathcal{G}}({\bm w}(T_{k}^{+}))-E_{\mathcal{G}}({\bm w}(T_{k}^{-})) \right\} T_{k}^{\mu} \Big\} \notag\\
    &~~~~~-  c\int_{0}^{t}\Phi(t,s)\frac{\partial E_{\mathcal{G}}({\bm w}(s))s^{\mu}}{\partial s}ds +   \int_{0}^{t}\Phi(t,s){\Sigma}^{1/2}({\bm w}(s))d{\bm B}(s). \notag
\end{align}

Based on \eqref{eq_centered_scaled}, for $t \in (T_{K},T_{K+1})$ we further have
\begin{align}
%    \label{eq_v_gamma_final}
    {\bm v}_{\gamma}(t) &\overset{d}{\approx} {\bm w}(t) + \sqrt{\gamma}ct^{\mu}E_{\mathcal{G}}({\bm w}(t)) - \sqrt{\gamma} c\sum_{k=1}^{K}\Big\{\Phi(t,T_{k}) \left\{E_{\mathcal{G}}({\bm w}(T_{k}^{+}))-E_{\mathcal{G}}({\bm w}(T_{k}^{-})) \right\} T_{k}^{\mu} \Big\} \notag\\
    &~~~~~-  c\int_{0}^{t}\Phi(t,s)\frac{\partial E_{\mathcal{G}}({\bm w}(s))s^{\mu}}{\partial s}ds +  + \sqrt{\gamma}\int_{0}^{t}\Phi(t,s){\Sigma}^{1/2}({\bm w}(s))d{\bm B}(s). \notag
\end{align}
Then we complete the proof.
\end{proof}

\newpage

\section{Experimental Setup Details}

We did all experiments in this paper using servers with a GPU (NVIDIA Quadro RTX 6000 with 24GB memory), two CPUs (each with 12 cores, Inter Xeon Gold 6136), and 192 GB memory. We use PyTorch~\cite{NEURIPS2019_bdbca288} for all experiments.

\subsection{Training Setup in Section~4.2}

The base ResNet model is implemented following~\cite{he2016deep, NEURIPS2020_703957b6}, and the base VGG model is implemented following~\cite{liu2017learning, NEURIPS2020_703957b6}. For our experiments in Table~\ref{dif-alg-table} and \ref{larg-flops-table}, we mainly follow the codes of \cite{NEURIPS2020_703957b6}. The detail hyperparameters to obtain the best results are summarized in Table~\ref{parameters-best-result-table}, where the learning rate decay scheme ``{[60, 160,...]}@[0.2, 0.2,...]'' means that the learning rate multiplied by 0.2 at 60 epoch and multiplied by 0.2 at 160 epoch, etc. We set  nonzero ratio lower bounds 0.3 and 0.25 respectively for experiments in Table~\ref{dif-alg-table} and \ref{larg-flops-table}  to avoid excessive pruning.

\subsection{Training Setup in Section~4.3}
The base VGG model and method for visualizing are implemented following \cite{NEURIPS2018_be3087e7, NEURIPS2020_a09e75c5}, which does not have batch normalization. For both SDP and {\texttt{AltSDP}} we use the similar learning rate schedule adopted by \cite{NEURIPS2020_a09e75c5}: fix the learning rate equals to $0.1$ at first 50\% epochs, then reduce the learning rate to 0.1\% of the base learning rate between 50\% and 90\% epochs, and keep reducing it to 0.1\% for the last 10\% epochs. The minibatch size is 128 for all experiments in Section~4.3.

\section{Additional Experimental Results}

\subsection{Visualizing Results for VGG-16 and WRN28X10}

\begin{figure}[!htbp]
	\centering
	\subfloat{
	\includegraphics[width=2.5in]{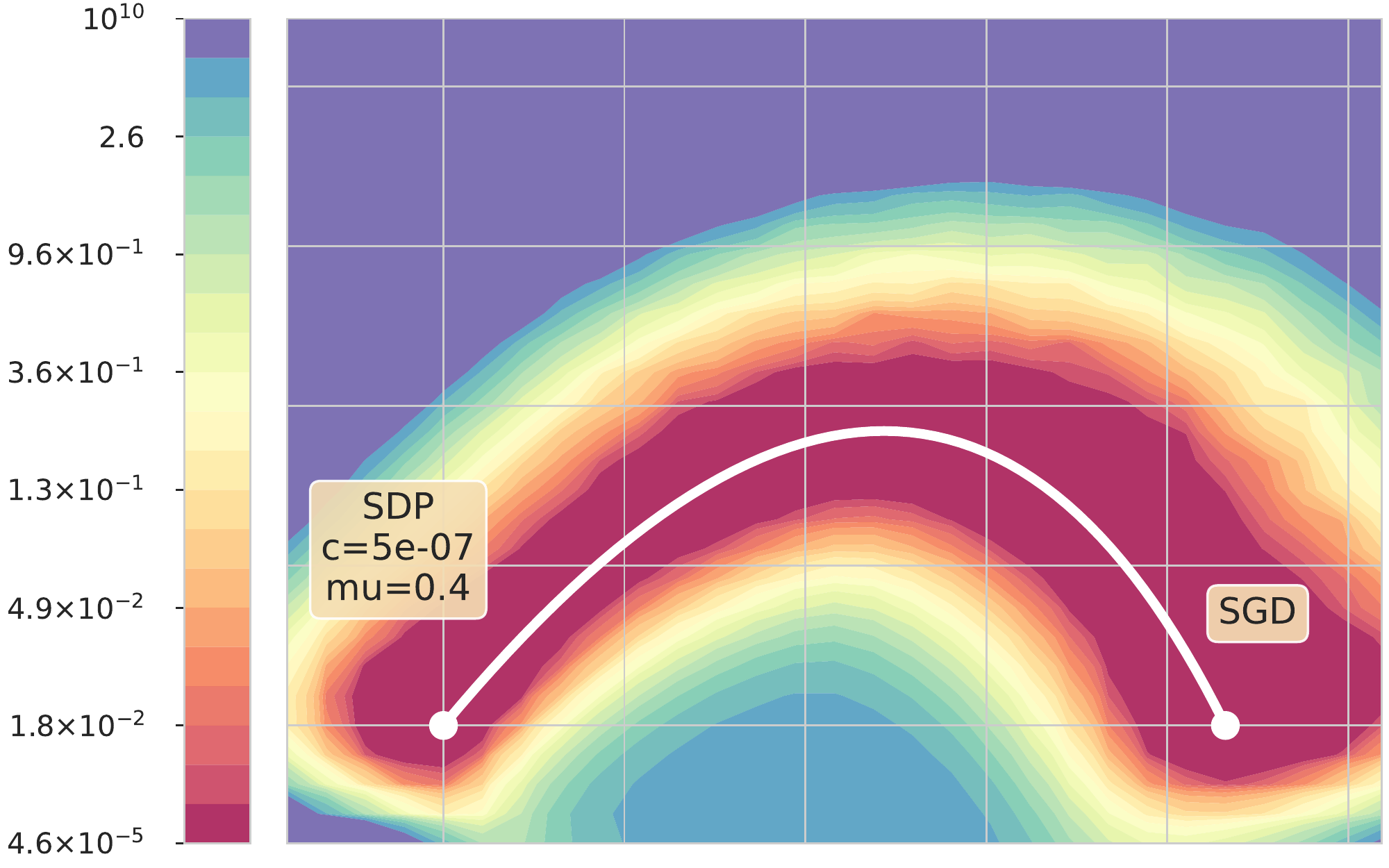}
	}
	\quad
	\subfloat{
	\includegraphics[width=2.5in]{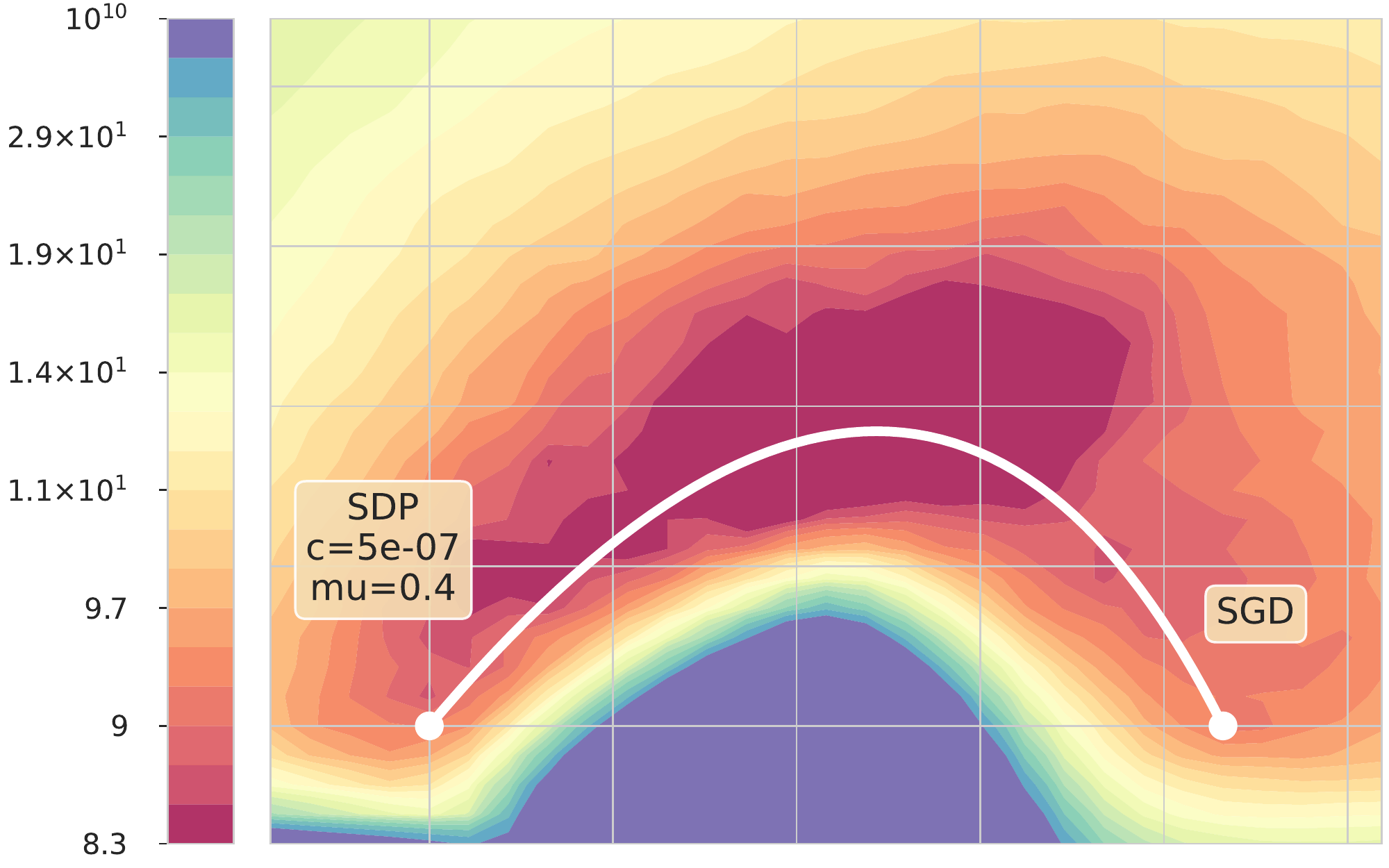}
	}
	\caption{The contour of training loss and testing error on the hyperplane for VGG-16 on CIFAR-10. Sparsity = 0 for both {\texttt{AltSDP}} and SGD, where we set $c=5 \times 10^{-7}$ and $\mu = 0.4$. The test accuracy of {\texttt{AltSDP}} is 0.9080 while that of SGD is 0.9089. When the values of $c$ and $\mu$ are small, our {\texttt{AltSDP} } cannot obtain sparse results and suffer performance loss.}
	\label{AP-contour-1}
\end{figure}

\begin{figure}[htbp]
	\centering
	\subfloat{
	\includegraphics[width=2.5in]{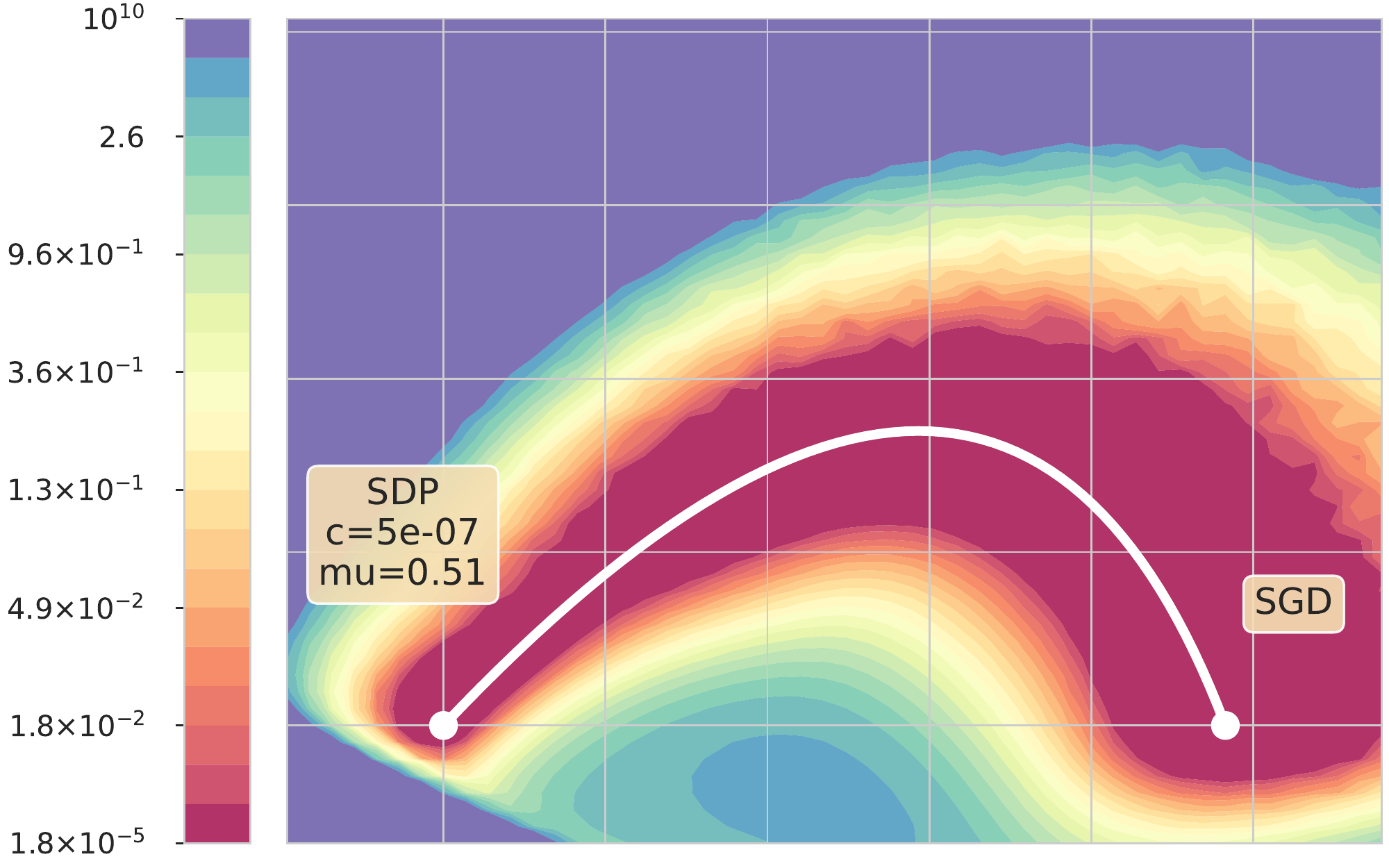}
	}
	\quad
	\subfloat{
	\includegraphics[width=2.5in]{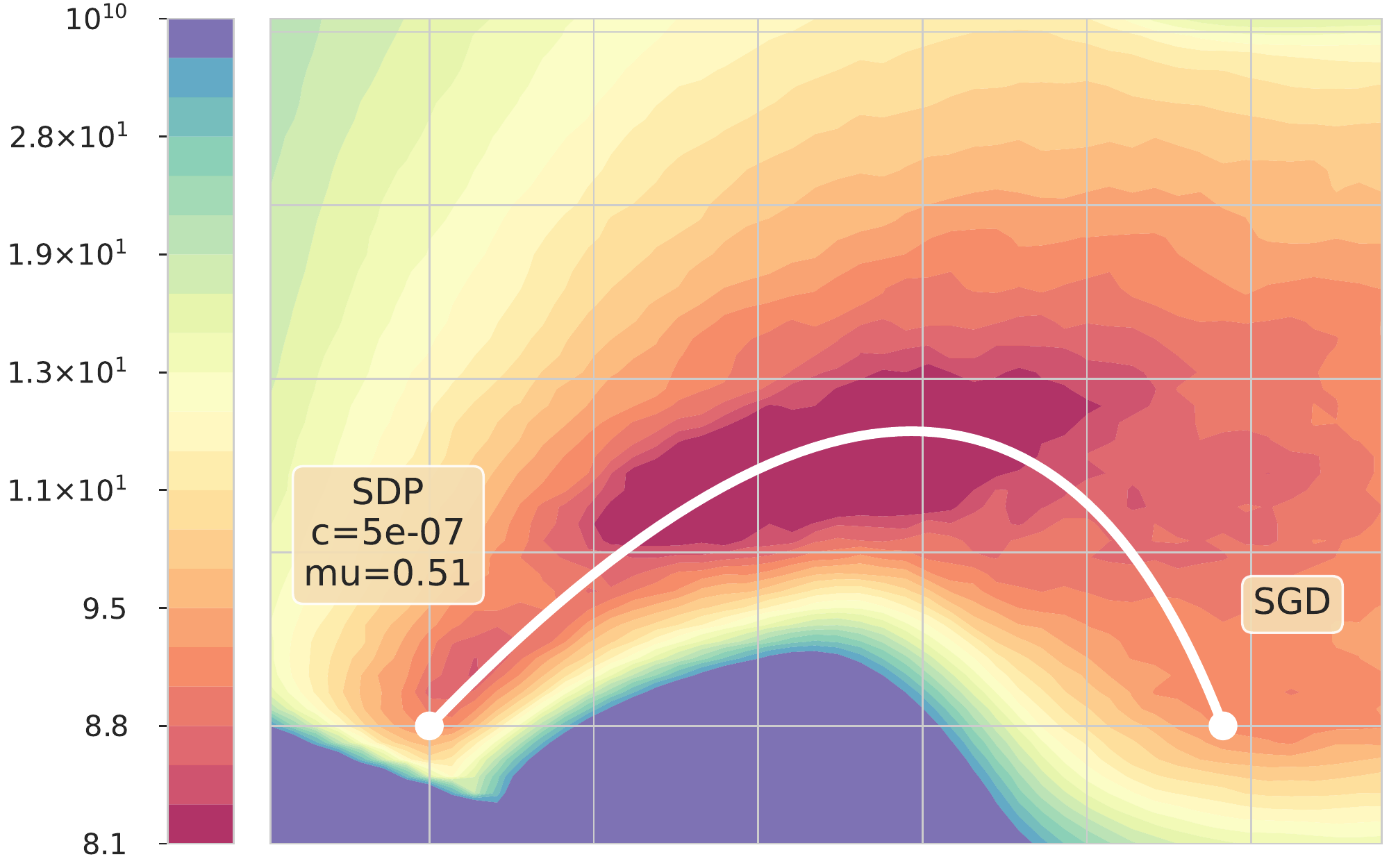}
	}
	\caption{The contour of training loss and testing error on the hyperplane for VGG-16 on CIFAR-10. Sparsity = 0.009 for {\texttt{AltSDP}} and Sparsity = 0 for SGD, where we set $c=5 \times 10^{-7}$ and $\mu = 0.51$.  The test accuracy of {\texttt{AltSDP}} is 0.9091 while that of SGD is 0.9089. When $\mu$ is slightly greater than 0.5, the model can become sparse and has good test accuracy.}
	\label{AP-contour-2}
\end{figure}

\begin{figure}[htbp]
	\centering
	\subfloat{
	\includegraphics[width=2.5in]{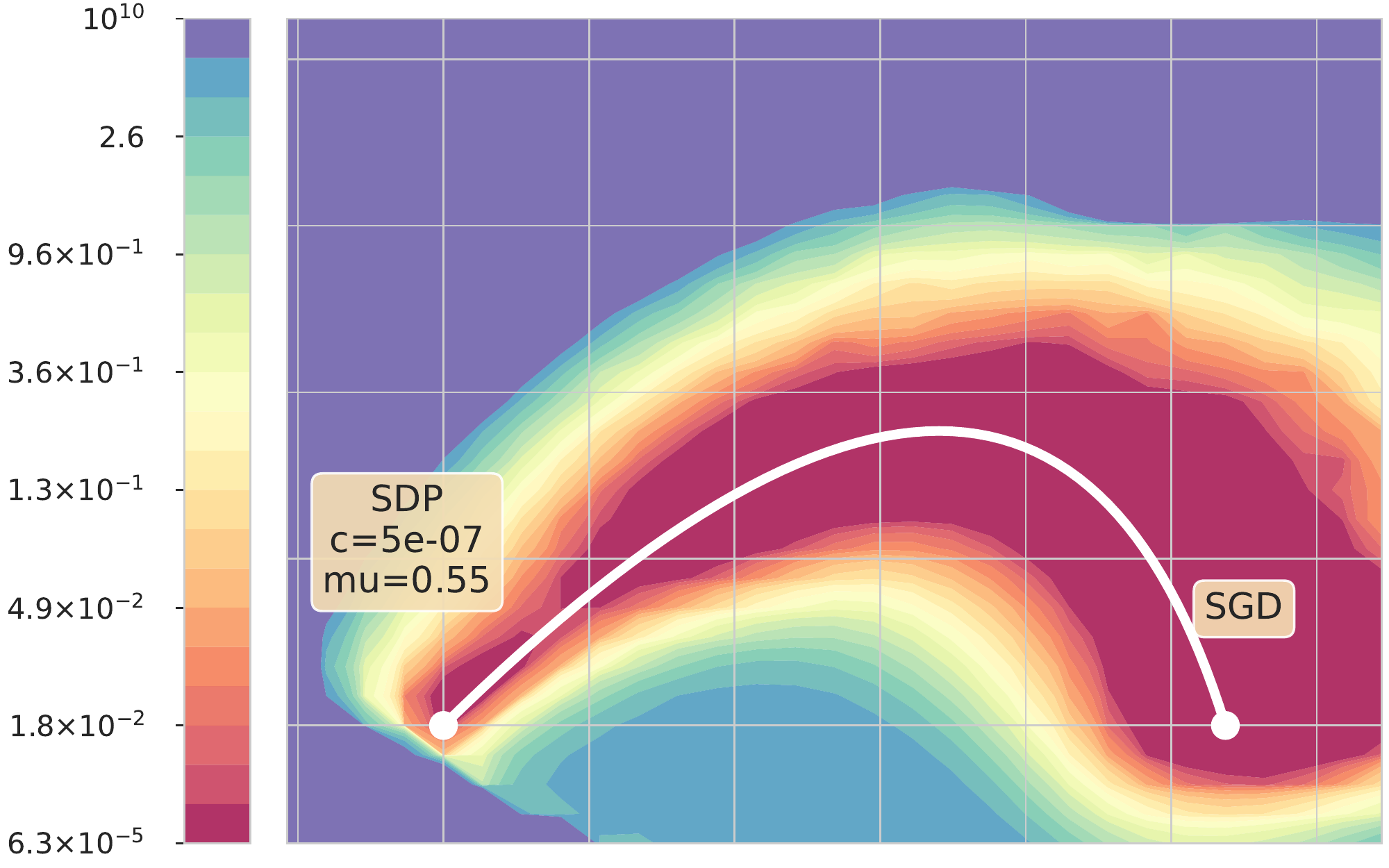}
	}
	\quad
	\subfloat{
	\includegraphics[width=2.5in]{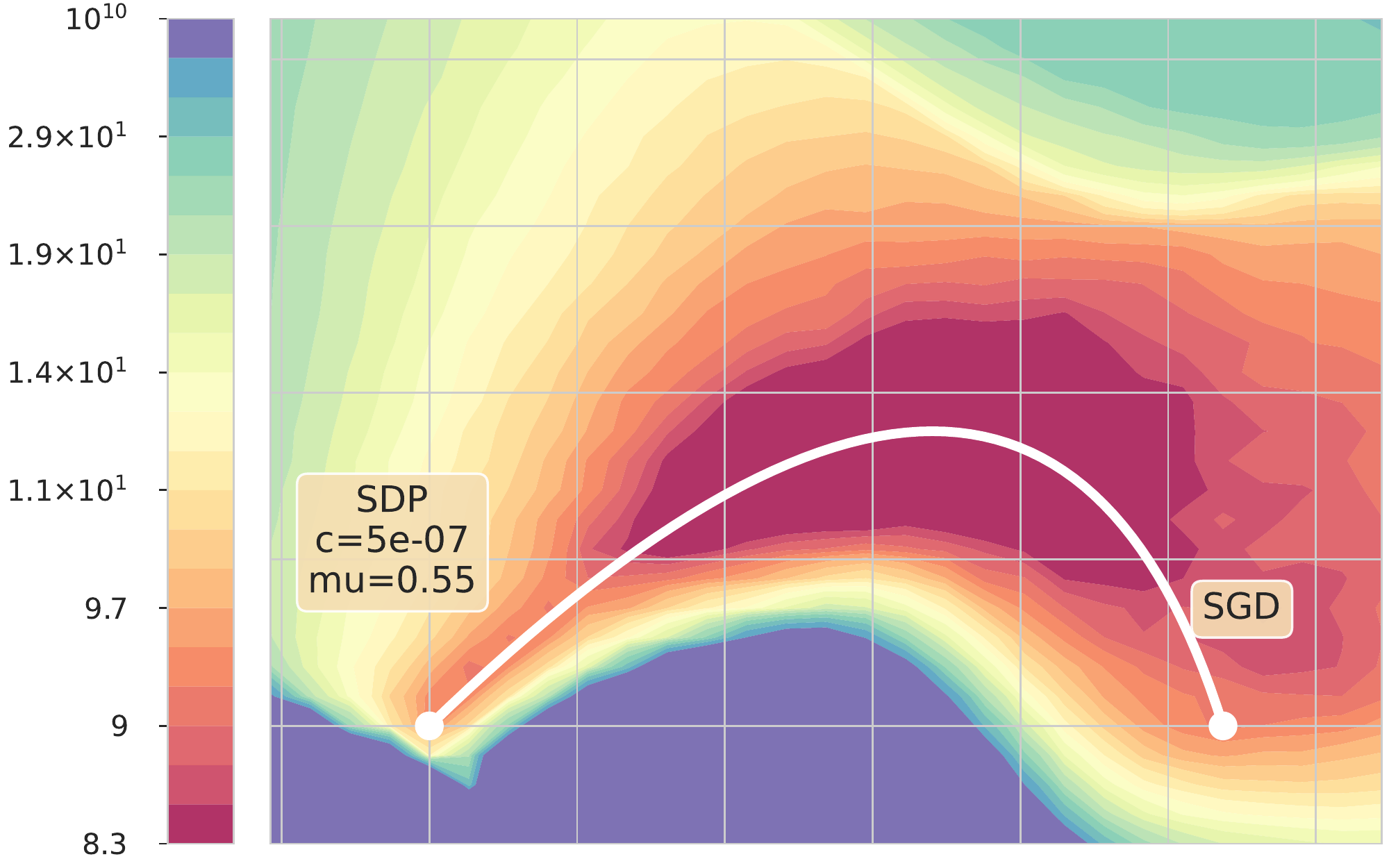}
	}
	\caption{The contour of training loss and testing error on the hyperplane for VGG-16 on CIFAR-10. Sparsity = 0.124 for {\texttt{AltSDP}} and Sparsity = 0 for SGD, where we set $c=5 \times 10^{-7}$ and $\mu = 0.55$.  The test accuracy of {\texttt{AltSDP}} is 0.9090 while that of SGD is 0.9089. When $\mu$ is slightly greater than 0.5, the model can become sparse and has good test accuracy.}
	\label{AP-contour-3}
\end{figure}

\begin{figure}[htbp]
	\centering
	\subfloat{
	\includegraphics[width=2.5in]{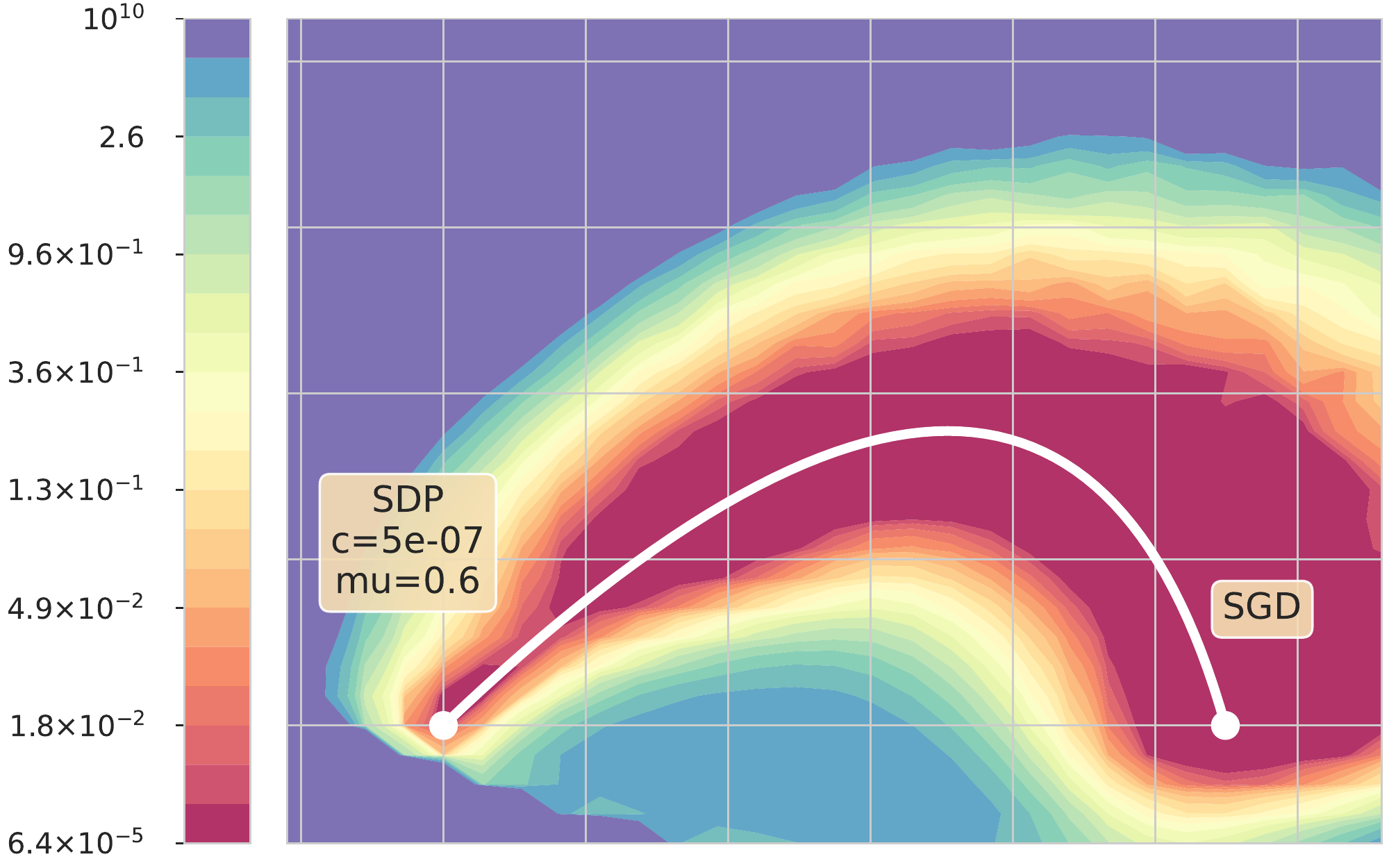}
	}
	\quad
	\subfloat{
	\includegraphics[width=2.5in]{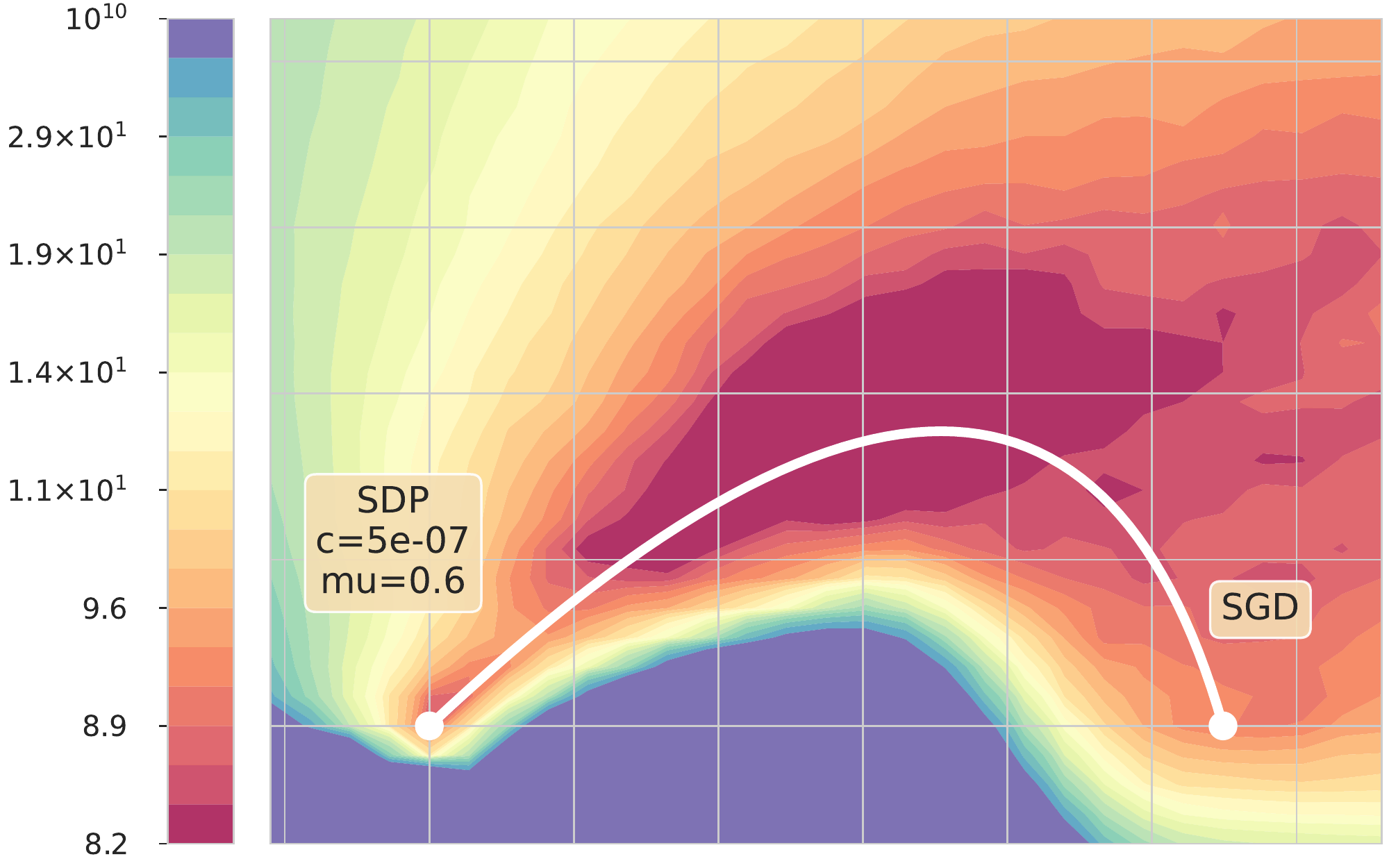}
	}
	\caption{The contour of training loss and testing error on the hyperplane for VGG-16 on CIFAR-10. Sparsity = 0.719 for {\texttt{AltSDP}} and Sparsity = 0 for SGD, where we set $c=5 \times 10^{-7}$ and $\mu = 0.6$.  The test accuracy of {\texttt{AltSDP}} is 0.9153  while that of SGD is 0.9089. As $\mu$ continues to grow larger, the model may become sparser. Here we find that, quite coincidentally, the test accuracy also improves, but as $\mu$ increases further, the model becomes sparser, which may eventually lead to performance degradation.}
	\label{AP-contour-4}
\end{figure}

\begin{figure}[htbp]
	\centering
	\subfloat{
	\includegraphics[width=2.5in]{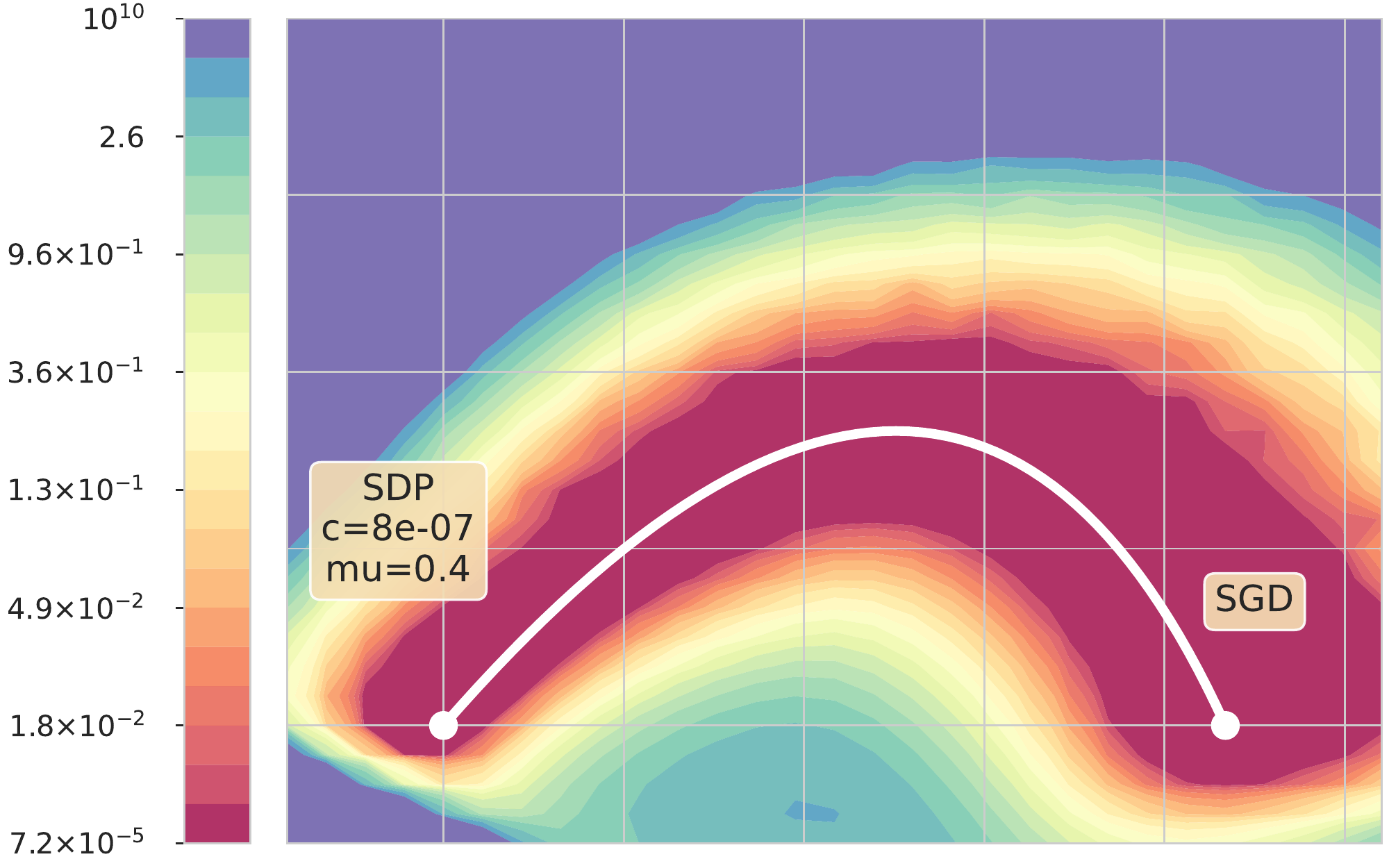}
	}
	\quad
	\subfloat{
	\includegraphics[width=2.5in]{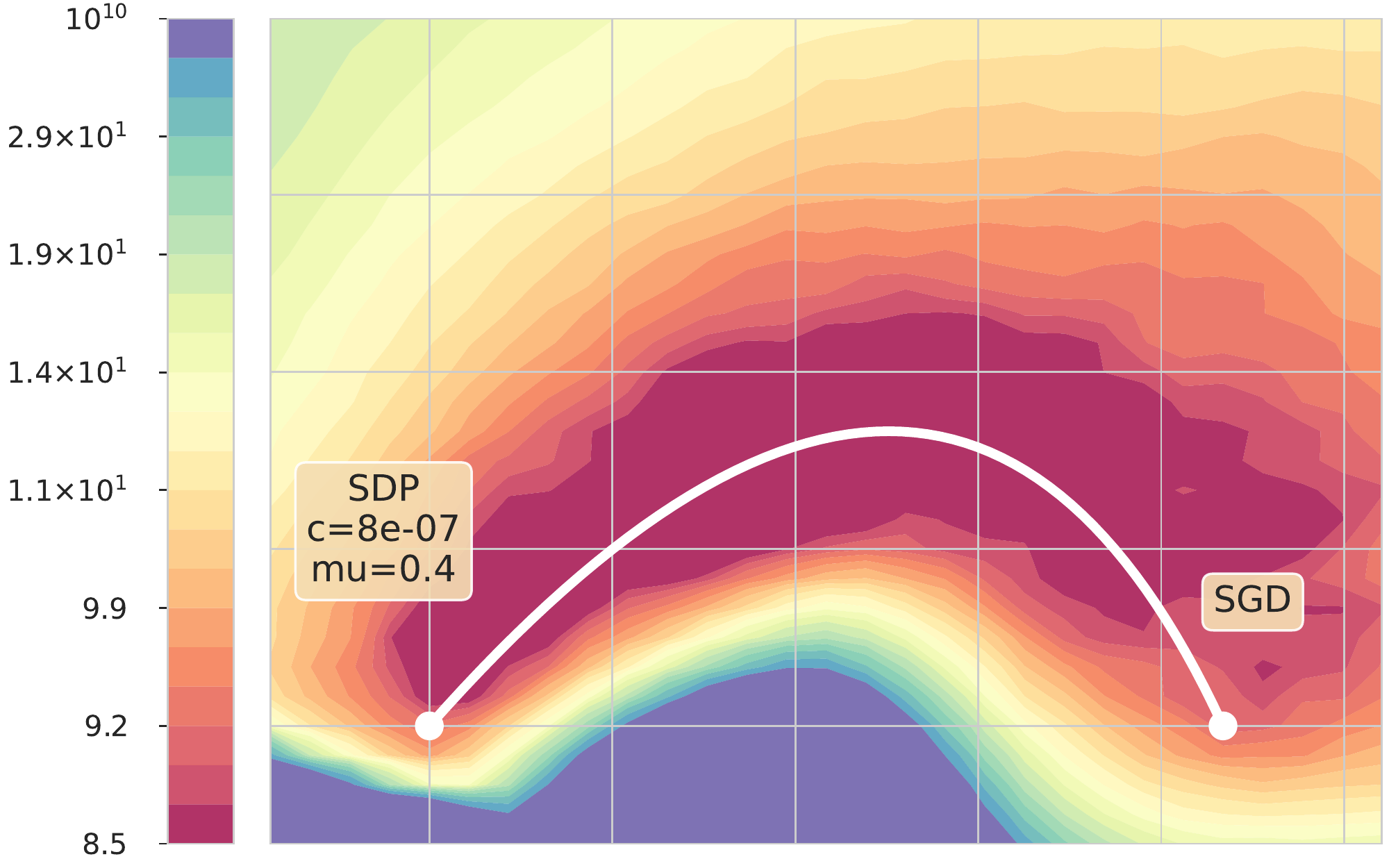}
	}
	\caption{The contour of training loss and testing error on the hyperplane for VGG-16 on CIFAR-10. Sparsity = 0.008 for {\texttt{AltSDP}} and Sparsity = 0 for SGD, where we set $c=8 \times 10^{-7}$ and $\mu = 0.4$.   The test accuracy of {\texttt{AltSDP}} is 0.9077 while that of SGD is 0.9089. When $c$ is increased, the model also becomes sparse, but the test accuracy decreases.}
	\label{AP-contour-5}
\end{figure}

\begin{figure}[htbp]
	\centering
	\subfloat{
	\includegraphics[width=2.5in]{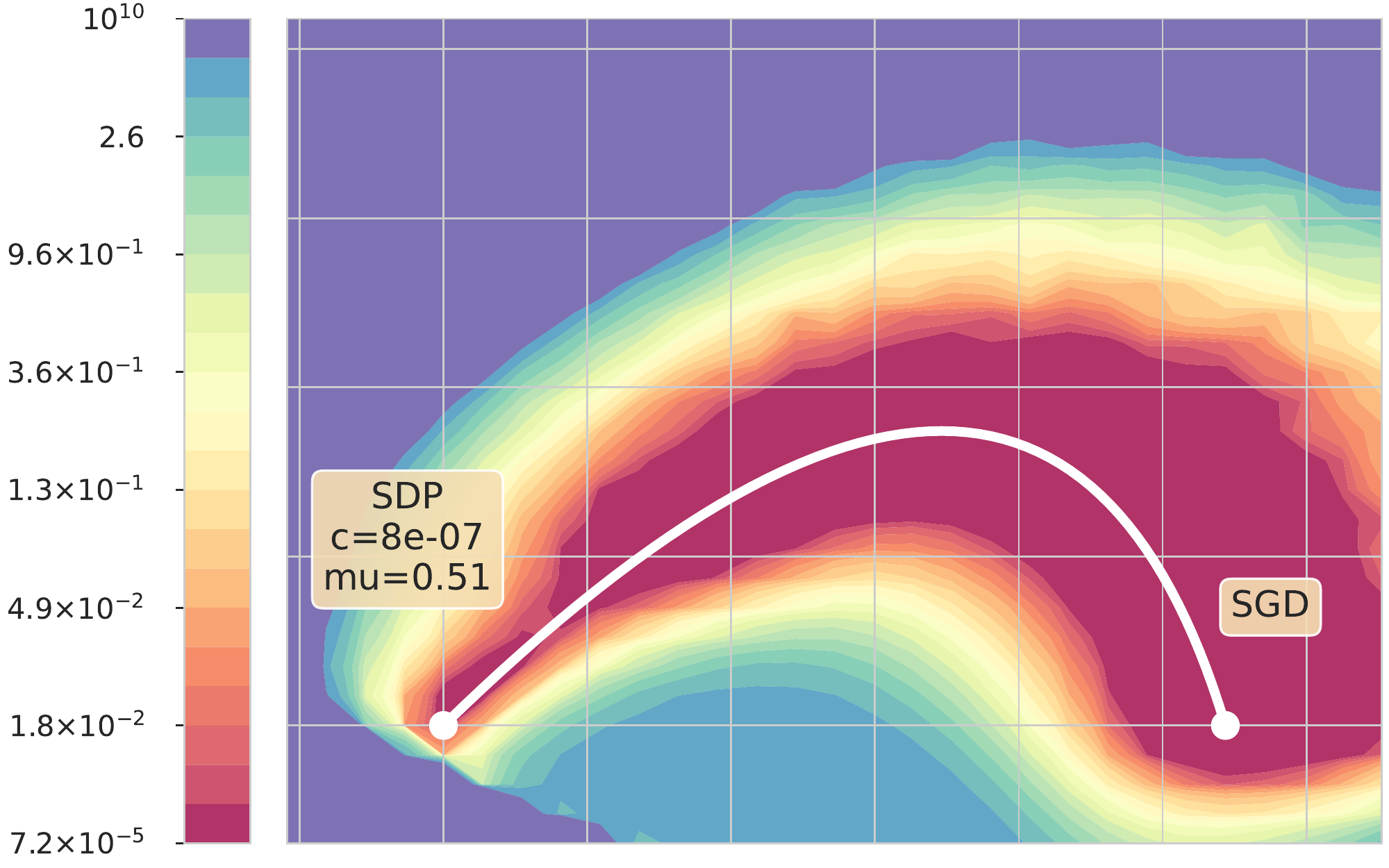}
	}
	\quad
	\subfloat{
	\includegraphics[width=2.5in]{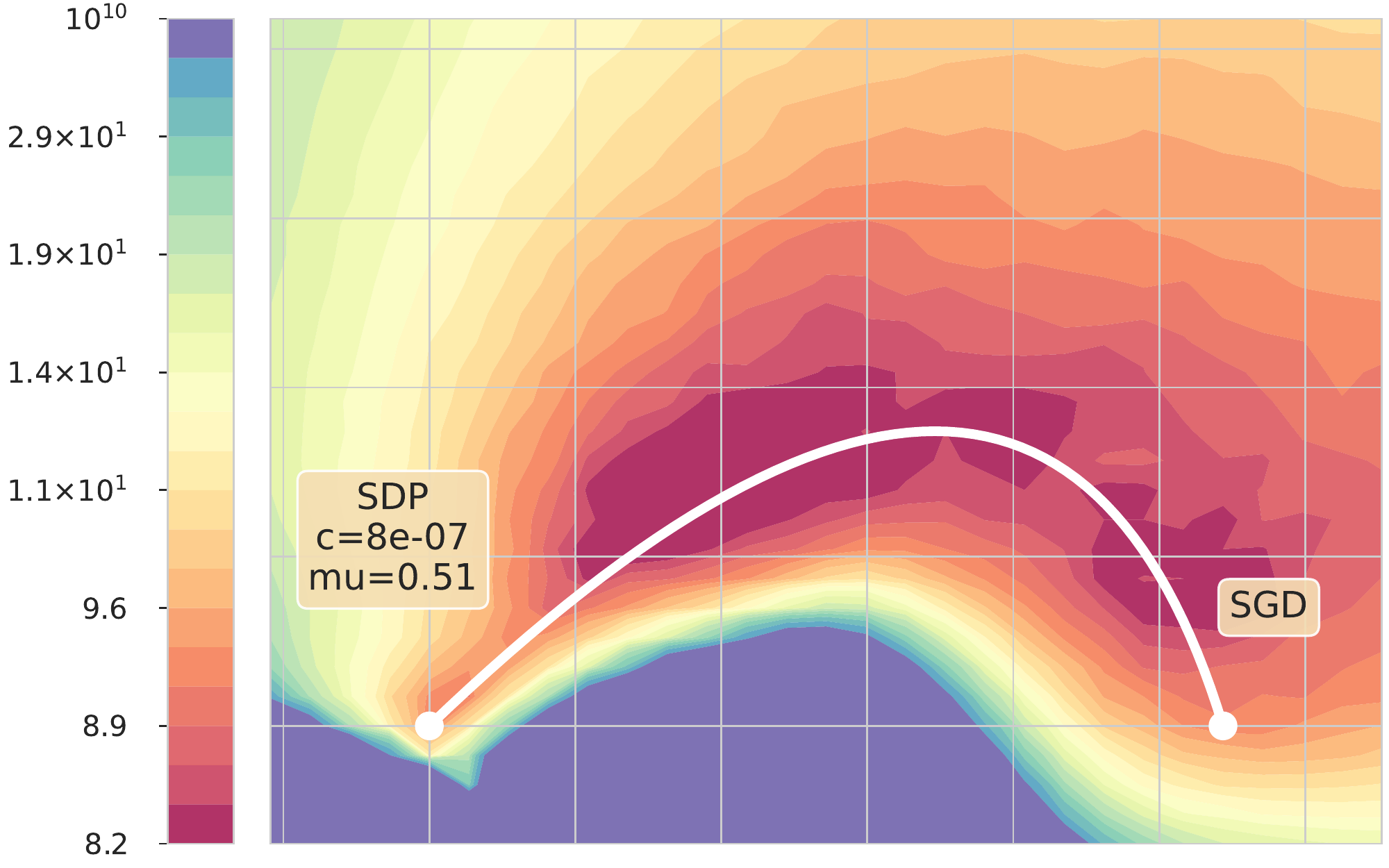}
	}
	\caption{The contour of training loss and testing error on the hyperplane for VGG-16 on CIFAR-10. Sparsity = 0.439 for {\texttt{AltSDP}} and Sparsity = 0 for SGD, where we set $c=8 \times 10^{-7}$ and $\mu = 0.51$.   The test accuracy of {\texttt{AltSDP}} is 0.9110 while that of SGD is 0.9089. When $\mu$ is slightly greater than 0.5 and $c$ increases, the model becomes sparse and  the test accuracy also improves.}
	\label{AP-contour-6}
\end{figure}

\begin{figure}[htbp]
	\centering
	\subfloat{
	\includegraphics[width=2.5in]{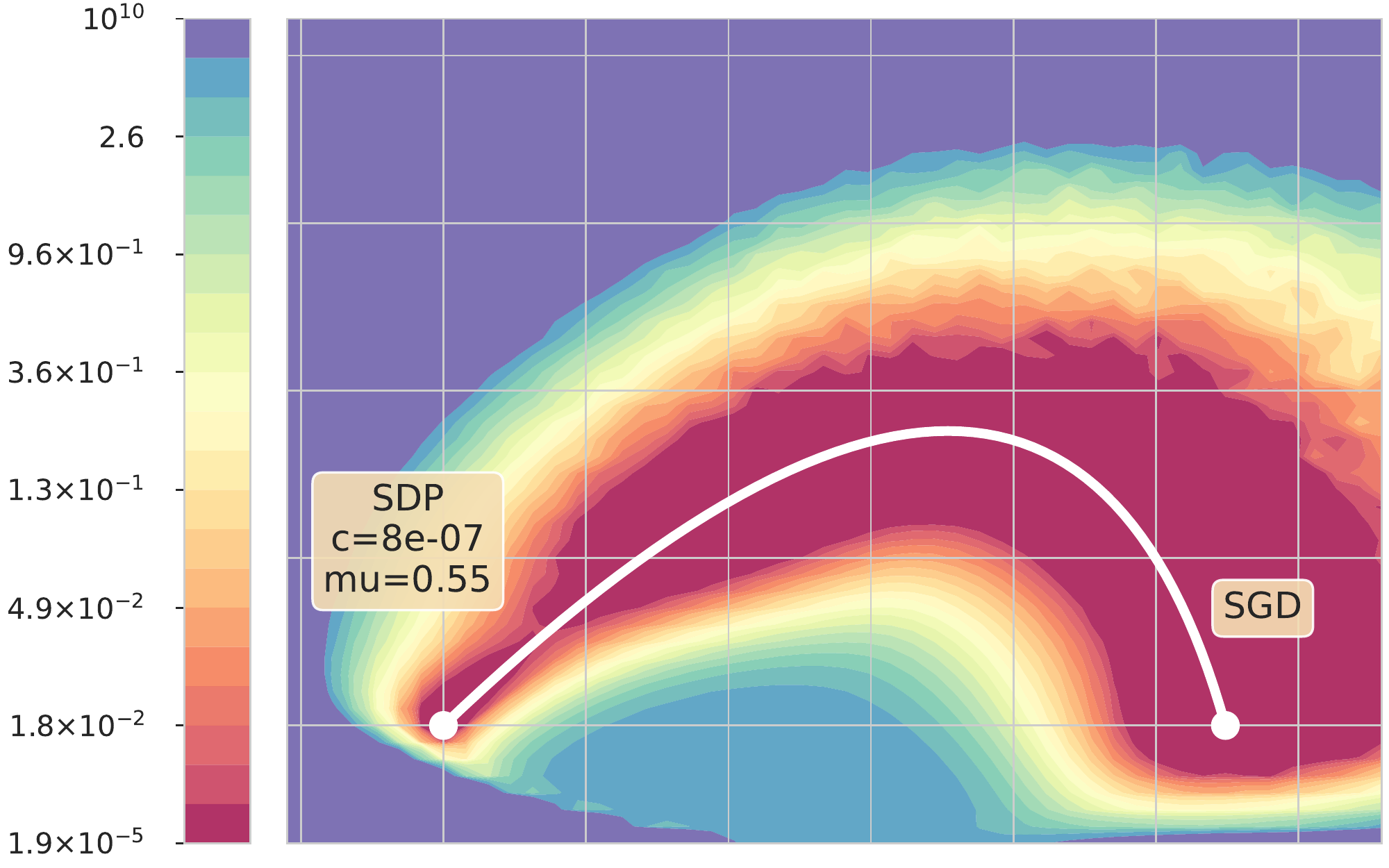}
	}
	\quad
	\subfloat{
	\includegraphics[width=2.5in]{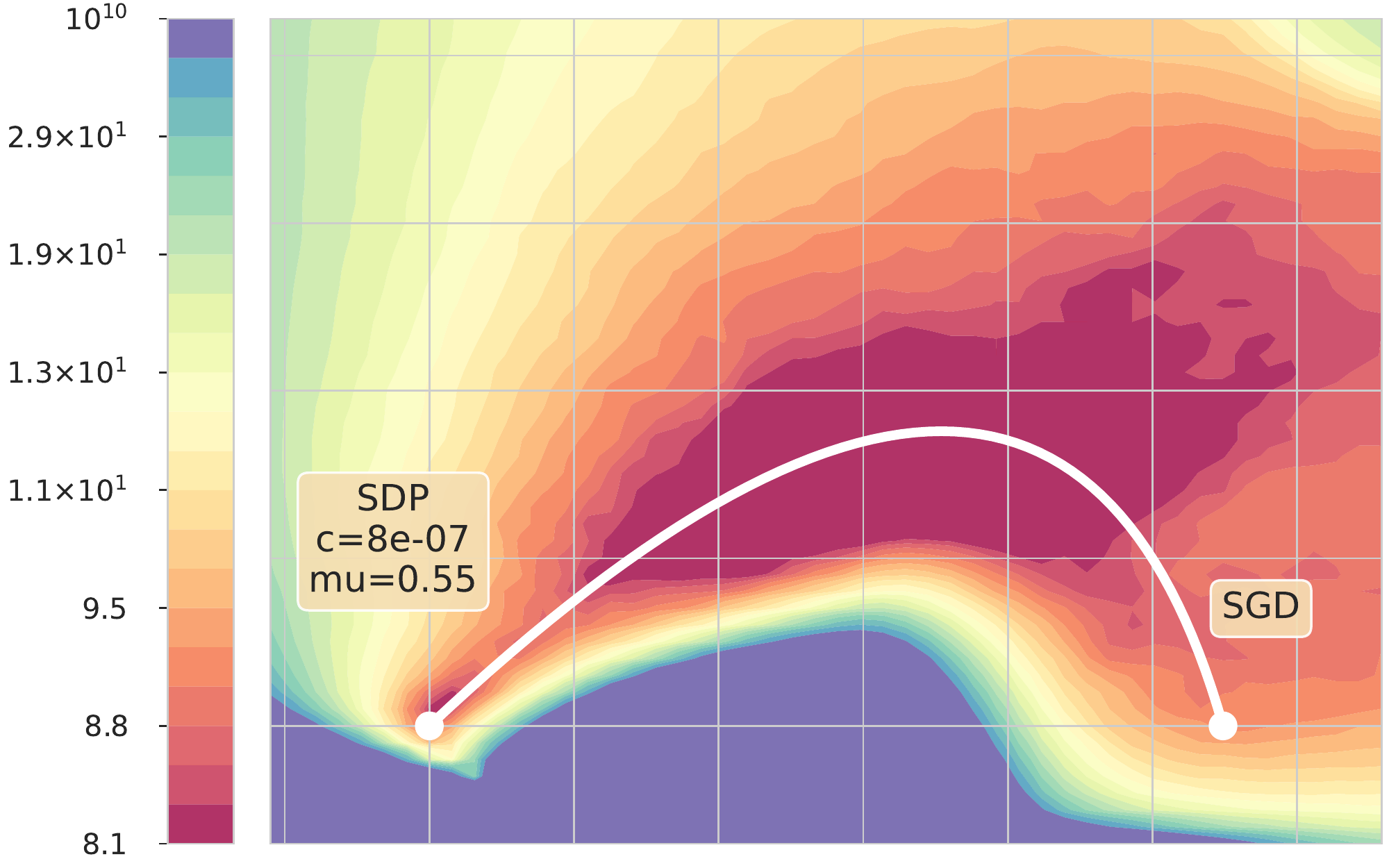}
	}
	\caption{The contour of training loss and testing error on the hyperplane for VGG-16 on CIFAR-10. Sparsity = 0.677 for {\texttt{AltSDP}} and Sparsity = 0 for SGD, where we set $c=8 \times 10^{-7}$ and $\mu = 0.55$.   The test accuracy of {\texttt{AltSDP}} is 0.9127 while that of SGD is 0.9089. When $\mu$ is slightly greater than 0.5 and $c$ increases, the model becomes sparse and  the test accuracy also improves.}
	\label{AP-contour-7}
\end{figure}

\begin{figure}[htbp]
	\centering
	
	\hspace{0pt}
	
	\includegraphics[width=5.5in]{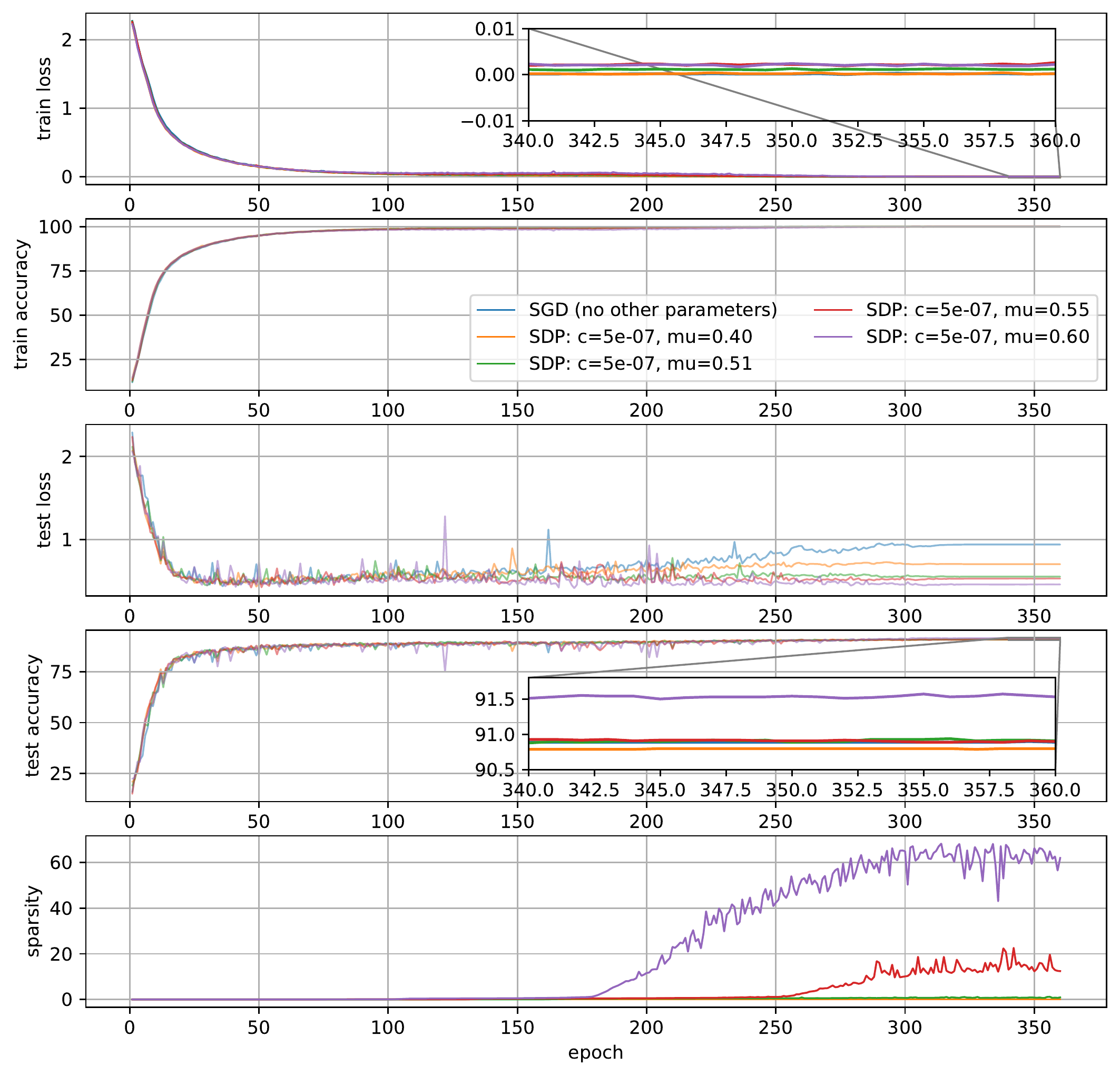}
	\caption{Learning trajectories of {\texttt{AltSDP} } and SGD for VGG-16 on CIFAR-10.}
	\label{AP-vgg16-cifar10}
\end{figure}

Here we first present the visualizing results for VGG-16 under different hyperparameters, which are used to check whether {\texttt{AltSDP}} reaches the same valley found by SGD. We train VGG16 on CIFAR-10 until nearly zero training loss using both SGD and {\texttt{AltSDP}}. We use the method of~\cite{garipov2018loss} to search for a quadratic B{\'e}zier curve of minimal training loss connecting the minima found by optimizers. In Figures~\ref{AP-contour-1}-\ref{AP-contour-7}, we respectively present the contour of training loss and testing error on the hyperplane for VGG-16 on CIFAR-10, where the hyperparameters are set according to that in Table~\ref{hyper-table}  presented in the main paper. We recall the Table~\ref{hyper-table} in Table~\ref{AP-hyper-table} here, where a more case when $c=5 \times 10^{-7}$ and $\mu = 0.6$ is added. 
We can see that {\texttt{AltSDP}} performs the structured directional pruning since the learned parameters of both SGD and {\texttt{AltSDP}} lie in the same flat minimum valley on the training loss landscape. In addition, Table~\ref{detail-result-vgg} presents the details of the learning trajectories for VGG-16 on CIFAR-10, and Figure~\ref{AP-vgg16-cifar10} presents the learning trajectories of {\texttt{AltSDP} } and SGD for VGG-16 on CIFAR-10.

\begin{figure}[htbp]
	\centering
	\subfloat{
	\includegraphics[width=2.5in]{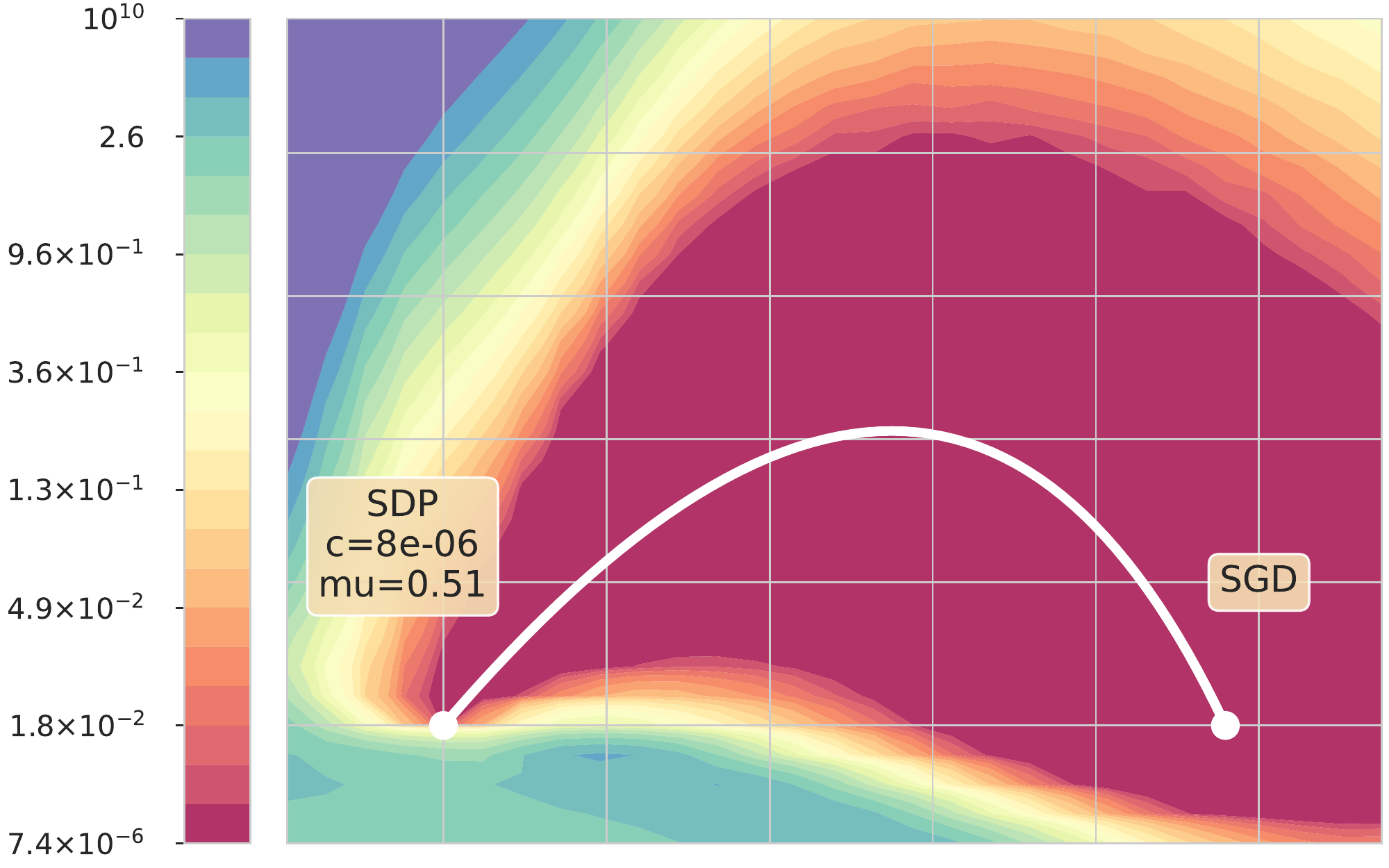}
	}
	\quad
	\subfloat{
	\includegraphics[width=2.5in]{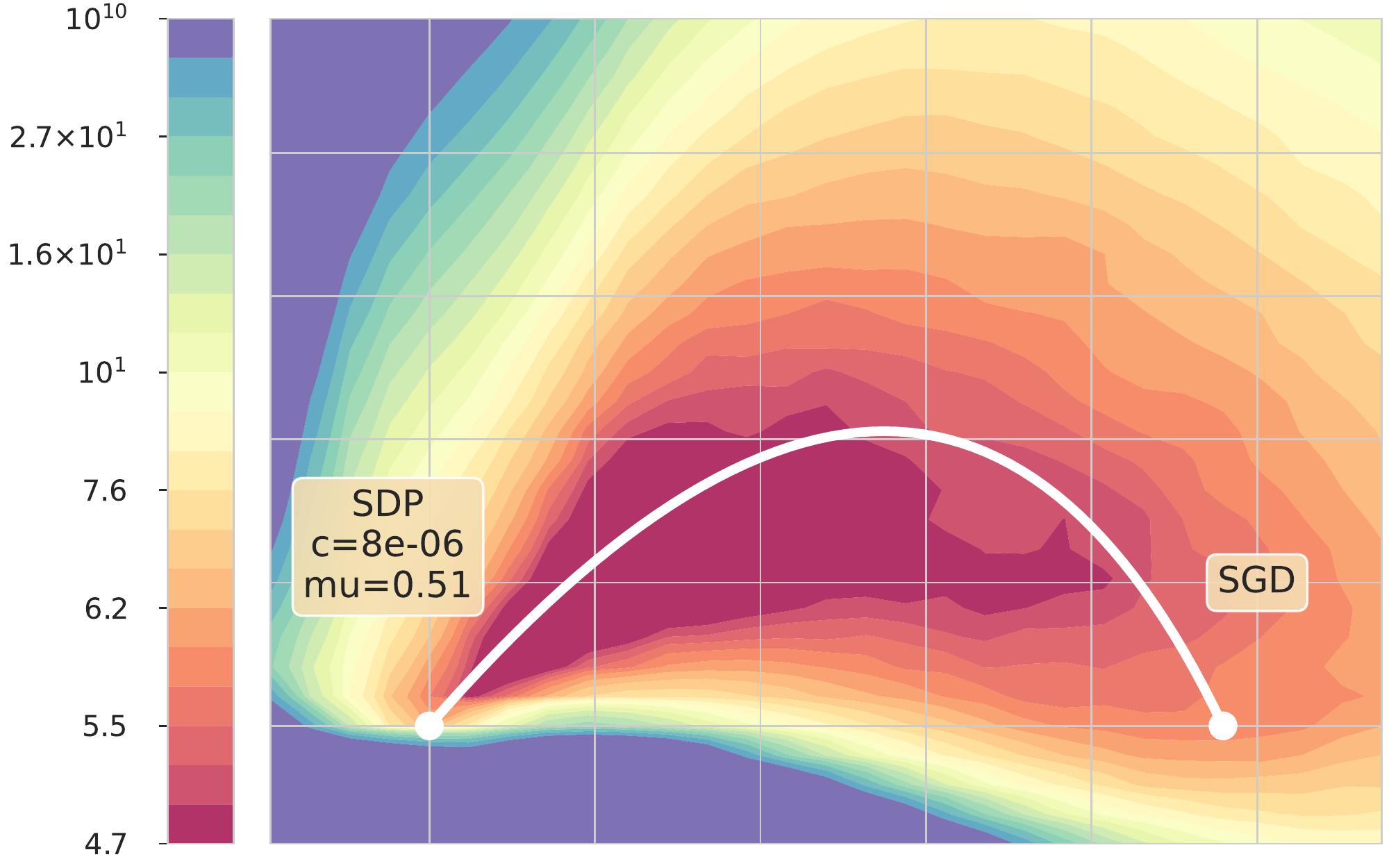}
	}
	\caption{The contour of training loss and testing error on the hyperplane for WRN28$\times$10 on CIFAR-10. Sparsity = 0.489 for {\texttt{AltSDP}} and Sparsity = 0 for SGD, where we set $c=8 \times 10^{-6}$ and $\mu = 0.51$.   The test accuracy of {\texttt{AltSDP}} is 0.9369 while that of SGD is 0.9418. When $\mu$ is slightly greater than 0.5 and $c$ further increases, the model becomes sparse, while the test accuracy is only reduced a little.}
	\label{AP-contour-8}
\end{figure}

Next, we present a visualizing example for WRN28$\times$10 to check whether {\texttt{AltSDP}} reaches the same valley found by SGD. We train WRN28$\times$10 on CIFAR-10 until nearly zero training loss using both SGD and {\texttt{AltSDP}}. In Figure~\ref{AP-contour-8}, we present the contour of training loss and testing error on the hyperplane for WRN28$\times$10 on CIFAR-10.
We can see that {\texttt{AltSDP}} performs the structured directional pruning since the learned parameters of both SGD and {\texttt{AltSDP}} lie in the same flat minimum valley on the training loss landscape. In addition, Table~\ref{detail-result-wrn} presents the details of the learning trajectories for WRN28$\times$10 on CIFAR-10.

\begin{table}[!htbp]
  \centering
  \scalebox{0.80}{
  \begin{tabular}{cccccccccccc}
    \toprule
    \multirow{2}{*}{Dataset/Model}  & Learning & \multirow{2}{*}{Decay Scheme} & Batch Size & \multirow{2}{*}{Hyper-parameters} & \multirow{2}{*}{Result}\\
     & Rate &  & Epoch &  &\\
    \midrule
        \multirow{2}{*}{CIFAR-10/ResNet-56}  & \multirow{2}{*}{0.05}  & [60, 160, 200, 220, 240] & \multirow{2}{*}{64/260} & \multirow{2}{*}{$c=10^{-5}$, $\mu=0.55$} &\multirow{8}{*}{Table~\ref{dif-alg-table}}\\
           &    &  @[0.2, 0.2, 0.2, 0.2, 0.4]  & \\
        \multirow{2}{*}{CIFAR-10/VGG16}  & \multirow{2}{*}{0.05}  & [60, 160, 200, 220, 240] & \multirow{2}{*}{64/260}  & \multirow{2}{*}{$c=10^{-5}$, $\mu=0.55$}\\
           &    &  @[0.2, 0.2, 0.2, 0.2, 0.4]  & \\
           
        \multirow{2}{*}{CIFAR-100/ResNet-56}  & \multirow{2}{*}{0.05}  & [60, 160, 200, 220, 240] & \multirow{2}{*}{64/260} & \multirow{2}{*}{$c=10^{-5}$, $\mu=0.53$}\\
           &    &  @[0.2, 0.2, 0.2, 0.2, 0.4]  & \\

        \multirow{2}{*}{CIFAR-10/VGG16}  & \multirow{2}{*}{0.05}  & [60, 160, 200, 220, 240] & \multirow{2}{*}{64/260} & \multirow{2}{*}{$c=10^{-5}$, $\mu=0.60$}\\
           &    &  @[0.2, 0.2, 0.2, 0.2, 0.4]  & \\
        \midrule
        \multirow{2}{*}{CIFAR-10/ResNet-56}  & \multirow{2}{*}{0.05}  & [100, 200, 220, 240, 260] & \multirow{2}{*}{64/280} & \multirow{2}{*}{$c=10^{-5}$, $\mu=0.58$} &\multirow{2}{*}{Table~\ref{larg-flops-table}}\\
           &    &  @[0.2, 0.2, 0.2, 0.2, 0.4]  & \\
    \bottomrule
  \end{tabular}}
  \caption{Detail hyper-parameters to obtain results in Section~4.2.}
  \label{parameters-best-result-table}
\end{table}

\begin{table}[!ht] \small
  \centering
  \scalebox{0.95}{
  \begin{tabular}{ccccccccc}
    \toprule
        &SGD   &\multicolumn{7}{c}{Structured directional pruning}\\
    \cmidrule(r){2-2}\cmidrule(r){3-9}
        & no other   & $c=$ 5e-7 & $c=$ 5e-7 & $c=$ 5e-7 & $c=$ 5e-7  & $c=$ 8e-7 &  $c=$ 8e-7 & $c=$ 8e-7\\
        & parameters & $u=0.40$  & $u=0.51$  & $u=0.55$ & $u=0.60$  & $u=0.40$  & $u=0.51$  & $u=0.55$\\
    \midrule
    Train loss  & 0.0001  & 0.0002  & 0.0012  & 0.0026 & 0.0022 & 0.0006  & 0.0023  & 0.0027 \\
    Test Acc.   & 0.9089  & 0.9080  & 0.9091  & 0.9090 & 0.9153 & 0.9077  & 0.9110  & 0.9127 \\
    Sparsity    & 0.0000  & 0.0000  & 0.0090  & 0.1242 & 0.7190 & 0.0084  & 0.4391  & 0.6767 \\
    \bottomrule
  \end{tabular}}
  \caption{The effect of hyper-parameters}
  \label{AP-hyper-table}
\end{table}

\begin{table}[htbp]
  \centering
  \scalebox{0.75}{
  \begin{tabular}{cccccccccccc}
    \toprule
    \multirow{2}{*}{Epoch}  & \multirow{2}{*}{Method} & {Training} & Training  & Testing & \multirow{2}{*}{Sparsity}\\
     &  & Loss & Accuracy(\%)  &   Accuracy(\%) &\\
    \midrule
        \multirow{8}{*}{60}  & SGD  & 0.1122   & 96.3280   & 87.2000   & 0.0000\\
        & SDP($c=5\times 10^{-7}$, $\mu=0.40$) & 0.1026    &96.5320     &88.0200    & 0.0000\\
        & SDP($c=5\times 10^{-7}$, $\mu=0.51$) & 0.1091    &96.3160     &87.4300    & 0.0000\\
        & SDP($c=5\times 10^{-7}$, $\mu=0.55$) & 0.1128    &96.1500     &87.7800    & 0.0000\\
        & SDP($c=5\times 10^{-7}$, $\mu=0.60$) & 0.1109    &96.3660     &84.7100    & 0.0000\\
        & SDP($c=8\times 10^{-7}$, $\mu=0.40$) & 0.1080    &96.4580     &88.3300    & 0.0000\\
        & SDP($c=8\times 10^{-7}$, $\mu=0.51$) & 0.1058    &96.4920     &84.9100    & 0.0000\\
        & SDP($c=8\times 10^{-7}$, $\mu=0.60$) & 0.1185    &96.0960     &88.1400    & 0.0000\\
    \midrule
        \multirow{8}{*}{120}  & SGD  & 0.0282  & 99.0680    &  88.3300   & 0.0000\\
        & SDP($c=5\times 10^{-7}$, $\mu=0.40$) & 0.0276    &99.0820     &88.9700    & 0.0000\\
        & SDP($c=5\times 10^{-7}$, $\mu=0.51$) & 0.0351    &98.8400     &89.6900    & 0.0000\\
        & SDP($c=5\times 10^{-7}$, $\mu=0.55$) & 0.0393    &98.7480     &89.1500    & 0.0000\\
        & SDP($c=5\times 10^{-7}$, $\mu=0.60$) & 0.0568    &98.2820     &88.6300    & 0.0043\\
        & SDP($c=8\times 10^{-7}$, $\mu=0.40$) & 0.0386    &98.8860     &89.1900    & 0.0000\\
        & SDP($c=8\times 10^{-7}$, $\mu=0.51$) & 0.0441    &98.6060     &88.0200    & 0.0017\\
        & SDP($c=8\times 10^{-7}$, $\mu=0.60$) & 0.0513    &98.3940     &88.0000    & 0.0042\\
    \midrule
        \multirow{8}{*}{180}  & SGD  & 0.0166   & 99.5040   & 89.7000 & 0.0000\\
        & SDP($c=5\times 10^{-7}$, $\mu=0.40$) & 0.0178    &99.4440     &89.1100    & 0.0000\\
        & SDP($c=5\times 10^{-7}$, $\mu=0.51$) & 0.0288    &99.0740     &89.1300    & 0.0036\\
        & SDP($c=5\times 10^{-7}$, $\mu=0.55$) & 0.0358    &98.8660     &88.4000    & 0.0048\\
        & SDP($c=5\times 10^{-7}$, $\mu=0.60$) & 0.0601    &98.1660     &88.0600    & 0.0150\\
        & SDP($c=8\times 10^{-7}$, $\mu=0.40$) & 0.0203    &99.3360     &89.3400    & 0.0000\\
        & SDP($c=8\times 10^{-7}$, $\mu=0.51$) & 0.0355    &98.8800     &88.7400    & 0.0055\\
        & SDP($c=8\times 10^{-7}$, $\mu=0.60$) & 0.0565    &98.2220     &89.1600    & 0.0464\\
    \midrule
        \multirow{8}{*}{240}  & SGD  &  0.0018  &  99.9480   & 90.5200 & 0.0000\\
        & SDP($c=5\times 10^{-7}$, $\mu=0.40$) & 0.0034    &99.8880     &90.1400    & 0.0000\\
        & SDP($c=5\times 10^{-7}$, $\mu=0.51$) & 0.0062    &99.8240     &90.6300    & 0.0047\\
        & SDP($c=5\times 10^{-7}$, $\mu=0.55$) & 0.0109    &99.6920     &90.3500    & 0.0109\\
        & SDP($c=5\times 10^{-7}$, $\mu=0.60$) & 0.0267    &99.1940     &89.1700    & 0.4466\\
        & SDP($c=8\times 10^{-7}$, $\mu=0.40$) & 0.0046    &99.8700     &90.2200    & 0.0029\\
        & SDP($c=8\times 10^{-7}$, $\mu=0.51$) & 0.0135    &99.6180     &90.6500    & 0.0734\\
        & SDP($c=8\times 10^{-7}$, $\mu=0.60$) & 0.0309    &99.0800     &90.5700    & 0.5221\\
    \midrule
        \multirow{8}{*}{300}  & SGD  & 0.0004  &  99.9820   &  90.7000  & 0.0000\\
        & SDP($c=5\times 10^{-7}$, $\mu=0.40$) & 0.0004    &99.9920     &90.7500    & 0.0000\\
        & SDP($c=5\times 10^{-7}$, $\mu=0.51$) & 0.0014    &99.9880     &90.8900    & 0.0081\\
        & SDP($c=5\times 10^{-7}$, $\mu=0.55$) & 0.0035    &99.9380     &90.9000    & 0.1010\\
        & SDP($c=5\times 10^{-7}$, $\mu=0.60$) & 0.0039    &99.9200     &91.2700    & 0.6490\\
        & SDP($c=8\times 10^{-7}$, $\mu=0.40$) & 0.0007    &99.9920     &90.9200    & 0.0036\\
        & SDP($c=8\times 10^{-7}$, $\mu=0.51$) & 0.0043    &99.8940     &91.1000    & 0.2136\\
        & SDP($c=8\times 10^{-7}$, $\mu=0.60$) & 0.0039    &99.9380     &91.2000    & 0.6496\\
    \midrule
        \multirow{8}{*}{360}  & SGD  &  0.0002   & 99.9960  &   90.8900 & 0.0000\\
        & SDP($c=5\times 10^{-7}$, $\mu=0.40$) & 0.0002    &99.9940     &90.8000    & 0.0000\\
        & SDP($c=5\times 10^{-7}$, $\mu=0.51$) & 0.0012    &99.9820     &90.9100    & 0.0090\\
        & SDP($c=5\times 10^{-7}$, $\mu=0.55$) & 0.0026    &99.9640     &90.9000    & 0.1242\\
        & SDP($c=5\times 10^{-7}$, $\mu=0.60$) & 0.0022    &99.9760     &91.5300    & 0.6197\\
        & SDP($c=8\times 10^{-7}$, $\mu=0.40$) & 0.0006    &99.9920     &90.7700    & 0.0084\\
        & SDP($c=8\times 10^{-7}$, $\mu=0.51$) & 0.0023    &99.9700     &91.1000    & 0.4391\\
        & SDP($c=8\times 10^{-7}$, $\mu=0.60$) & 0.0027    &99.9640     &91.2700    & 0.6767\\
    \bottomrule
  \end{tabular}}
  \caption{Details of the learning trajectories for VGG-16.}
  \label{detail-result-vgg}
\end{table}

% \clearpage

\begin{table}[htbp]
  \centering
  \scalebox{0.9}{
  \begin{tabular}{cccccccccccc}
    \toprule
    \multirow{2}{*}{Epoch}  & \multirow{2}{*}{Method} & {Training} & Training  & Testing & \multirow{2}{*}{Sparsity}\\
     &  & Loss & Accuracy(\%)  &   Accuracy(\%) &\\
    \midrule
        \multirow{2}{*}{40}  & SGD  & 0.0278    &99.0920     &90.5600    & 0.0000\\
        & SDP($c=8\times 10^{-6}$, $\mu=0.55$) & 0.3481    &87.9880     &69.4000 & 0.0000\\
    \midrule
        \multirow{2}{*}{80}  & SGD  & 0.0091    &99.7140     &91.7900    & 0.0000\\
        & SDP($c=8\times 10^{-6}$, $\mu=0.55$) & 0.3325    &88.5620     &57.8100 & 0.0678\\
    \midrule
        \multirow{2}{*}{120}  & SGD  & 0.0003    &99.9960     &93.9300  & 0.0000\\
        & SDP($c=8\times 10^{-6}$, $\mu=0.55$) & 0.2923    &90.0820     &83.4300 & 0.1945\\
    \midrule
        \multirow{2}{*}{160}  & SGD  & 0.0001    &99.9980     &94.1200   & 0.0000\\
        & SDP($c=8\times 10^{-6}$, $\mu=0.55$) & 0.1749    &94.0400     &86.5100 & 0.2438\\
    \midrule
        \multirow{2}{*}{200}  & SGD  & 0.0001   &100.0000     &94.1800  & 0.0000\\
        & SDP($c=8\times 10^{-6}$, $\mu=0.55$) & 0.0118    &99.7720     &93.6900 & 0.4890\\
    \bottomrule
  \end{tabular}}
  \caption{Details of the learning trajectories for WRN28$\times$10.}
  \label{detail-result-wrn}
\end{table}

\newpage

\subsection{Experimental Results on MNIST Dataset}
We also test {\texttt{AltSDP}} in a basic DNN model with 2 convolution layers and 2 full connection layers on the MNIST dataset. The learning rate is 0.1 at the beginning and mutiplied by 0.5 each 30 epochs. The results are listed in Table~\ref{detail-result-dnn}. We can see that when $\mu$ becomes larger, the model becomes sparser, which eventually leads to performance degradation.

\begin{table}[htbp]
  \centering
  \scalebox{0.9}{
  \begin{tabular}{cccccccccccc}
    \toprule
    \multirow{2}{*}{Epoch}  & \multirow{2}{*}{Method} & {Training} & Training  & Testing & \multirow{2}{*}{Sparsity}\\
     &  & Loss & Accuracy(\%)  &   Accuracy(\%) &\\
    \midrule
        \multirow{4}{*}{10}  & SGD  & 0.0050    &99.8533     &99.0400    & 0.0000\\
        & SDP($c=1\times 10^{-6}$, $\mu=0.51$) & 0.0063    &99.8250     &98.8600 & 0.0000\\
        & SDP($c=1\times 10^{-6}$, $\mu=0.55$) &  0.0079    &99.7533    &98.9700 & 0.0000\\
        & SDP($c=1\times 10^{-6}$, $\mu=0.60$) & 0.0074    &99.7633     & 98.9400 & 0.0000\\
    \midrule
        \multirow{4}{*}{40}  & SGD  & 0.0000   &100.0000     &99.2300    & 0.0000\\
        & SDP($c=1\times 10^{-6}$, $\mu=0.51$) & 0.0005   &100.0000     &99.2400 & 0.0000\\
        & SDP($c=1\times 10^{-6}$, $\mu=0.55$) & 0.0012   & 99.9883   & 99.2500 & 0.0000\\
        & SDP($c=1\times 10^{-6}$, $\mu=0.60$) & 0.0055    &99.8867   &  99.0400 & 0.2254\\
    \midrule
        \multirow{4}{*}{80}  & SGD  & 0.0000   &100.0000     &99.2300    & 0.0000\\
        & SDP($c=1\times 10^{-6}$, $\mu=0.51$) & 0.0031    &99.9667     &99.1600 & 0.2041\\
        & SDP($c=1\times 10^{-6}$, $\mu=0.55$) & 0.0038    &99.9483     &99.2000 & 0.3466\\
        & SDP($c=1\times 10^{-6}$, $\mu=0.60$) & 0.0045    &99.9300     & 99.1400 & 0.4748\\
    \midrule
        \multirow{4}{*}{120}  & SGD  & 0.0000   &100.0000     &99.2300  & 0.0000\\
        & SDP($c=1\times 10^{-6}$, $\mu=0.51$) & 0.0019    &99.9967     &99.2300 & 0.3245\\
        & SDP($c=1\times 10^{-6}$, $\mu=0.55$) & 0.0023    &99.9917     & 99.2700 & 0.4350\\
        & SDP($c=1\times 10^{-6}$, $\mu=0.60$) & 0.0032    &99.9850    & 99.2200 & 0.5100\\
    \midrule
        \multirow{4}{*}{160}  & SGD  & 0.0000   &100.0000     &99.2300   & 0.0000\\
        & SDP($c=1\times 10^{-6}$, $\mu=0.51$) & 0.0021   &100.0000     &99.1300 & 0.3626\\
        & SDP($c=1\times 10^{-6}$, $\mu=0.55$) & 0.0025    &99.9983     & 99.2700 & 0.4599\\
        & SDP($c=1\times 10^{-6}$, $\mu=0.60$) & 0.0034    &99.9900     & 99.1300 & 0.5184\\
    \midrule
        \multirow{4}{*}{200}  & SGD  & 0.0000   &100.0000    &99.2300  & 0.0000\\
        & SDP($c=1\times 10^{-6}$, $\mu=0.51$) & 0.0023   &100.0000     &99.2100 & 0.3667\\
        & SDP($c=1\times 10^{-6}$, $\mu=0.55$) & 0.0026   &100.0000     &99.2700    & 0.4683\\
        & SDP($c=1\times 10^{-6}$, $\mu=0.60$) & 0.0035    &99.9883     & 99.1500 & 0.5204\\
    \bottomrule
  \end{tabular}}
  \caption{Results on small dataset (MNIST)}
  \label{detail-result-dnn}
\end{table}

\end{spacing}

\end{document}